\documentclass{article}
\usepackage{iclr2027_conference,times}
\usepackage{amsmath,amssymb}
\usepackage{graphicx}
\usepackage{float}
\usepackage{caption}
\usepackage{wrapfig}
\usepackage{booktabs}
\usepackage{colortbl}
\usepackage{tabularx}
\usepackage{hyperref}
\usepackage[capitalize]{cleveref}
\crefname{section}{Sec.}{Secs.}
\Crefname{section}{Section}{Sections}
\crefname{table}{Tab.}{Tabs.}
\Crefname{table}{Table}{Tables}

\title{EgoFound3R: End-to-End Egocentric Hand \mbox{Reconstruction} in World Space with \mbox{Point-Wise} Interaction Attributes}
\author{Hongming Fu\textsuperscript{1,2}\quad Jingcheng Shi\textsuperscript{2,3}\quad Wenjia Wang\textsuperscript{4}\quad Binhua Zuo\textsuperscript{2}$^{*}$\quad Bo Zhao\textsuperscript{1}\thanks{Corresponding authors.}\\[0.45em]
\small\textsuperscript{1}School of AI, Shanghai Jiao Tong University\quad
\textsuperscript{2}QuicRobot\\[0.15em]
\small\textsuperscript{3}Rutgers University\quad
\textsuperscript{4}The University of Hong Kong\\[0.30em]
\small\texttt{\{fuhongming, bo.zhao\}@sjtu.edu.cn}\quad
\href{https://mint-sjtu.github.io/EgoFound3R.io/}{\textcolor{magenta}{\texttt{\small mint-sjtu.github.io/EgoFound3R.io}}}}

\newcommand{\method}{EgoFound3R}
\definecolor{bestcell}{RGB}{210,227,246}
\definecolor{secondcell}{RGB}{252,242,213}
\newcommand{\bestv}[1]{\cellcolor{bestcell}\textbf{#1}}
\newcommand{\secondv}[1]{\cellcolor{secondcell}\underline{#1}}
\hypersetup{pdftitle={EgoFound3R: End-to-End Egocentric Hand Reconstruction in World Space with Point-Wise Interaction Attributes},pdfauthor={Hongming Fu, Jingcheng Shi, Wenjia Wang, Binhua Zuo, Bo Zhao}}

\iclrfinalcopy
\begin{document}
\maketitle

\fancyhead{}
\renewcommand{\headrulewidth}{0pt}

\vspace{-1.4em}

\begin{abstract}
Egocentric video has become a primary source of supervision for embodied models, and its value rests on recovering hand motion in world coordinates, which camera motion and hand occlusion make difficult.
Existing reconstruction pipelines typically separate hand and scene estimation, leave interaction attributes to separate task-specific models, and invoke several models per video, so no prior reconstruction model estimates these attributes and throughput becomes a practical constraint on large-scale annotation.
We therefore introduce \method, a unified end-to-end model that estimates world-space hand geometry in a metric scale shared with the scene, and predicts point-wise interaction attributes, including visibility, contact, and distance.
The model integrates three designs: (i) structured hand prompts that transfer pretrained geometric priors to world-space hand reconstruction; (ii) an explicit hand representation that decodes hand geometry and interaction attributes; and (iii) a shared-parameter multi-rate design that lowers inference cost.
Together, these designs predict hand geometry and point-wise attributes in one pass.
On OakInk-v2, TACO, and HOI4D, \method{} reduces the mean per-joint position error (MPJPE) by 43.2\%, 22.4\%, and 11.6\% over previous methods and predicts point-wise contact and distance alongside the geometry in the same pass, while attaining approximately $6\times$ higher throughput.
\end{abstract}

\begin{figure}[H]
  \centering
  \includegraphics[width=\linewidth]{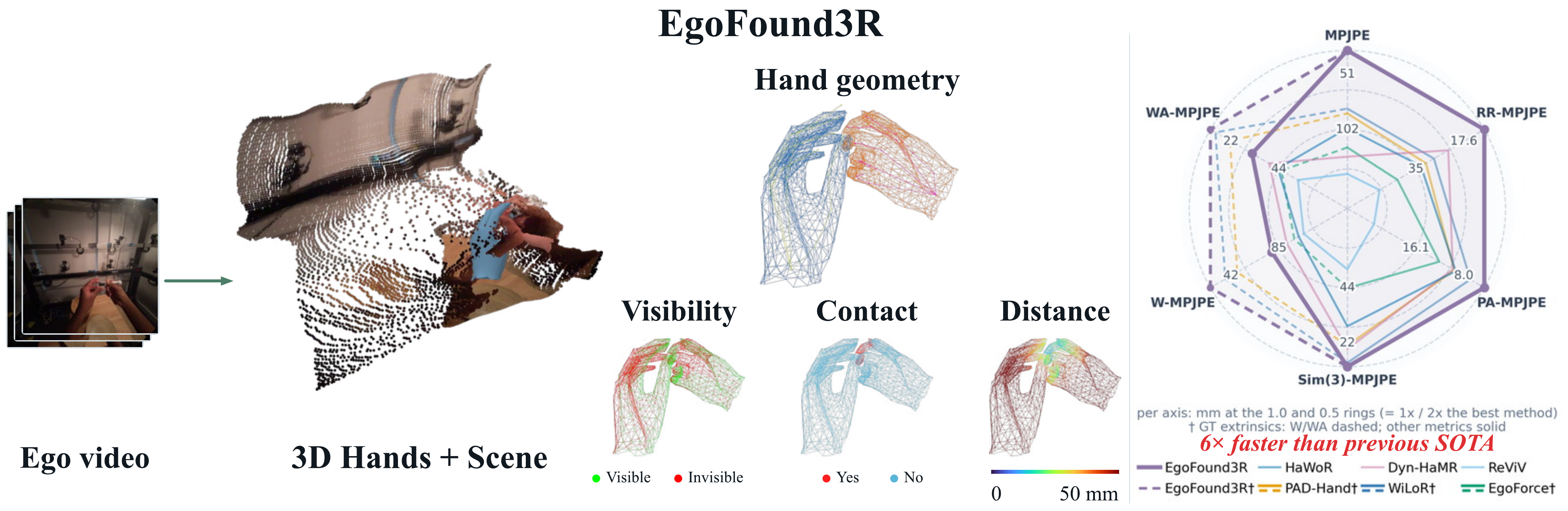}
  \caption{\textbf{\method{} at a glance.} End-to-end world-space hand reconstruction with point-wise interaction attributes; the radar panel summarizes \cref{tab:joint}.}
  \label{fig:teaser}
  \vspace{-0.8em}
\end{figure}

\section{Introduction}
\label{sec:introduction}

Egocentric video captures manipulation as it is performed, and it is the primary channel through which embodied models acquire human manipulation skill: dexterous manipulation, world modeling, and vision-language-action policies increasingly learn from first-person recordings~\citep{grauman2022ego4d,paliwal2026doasido,li2026open}, which now span $10^5$ to $10^6$ hours~\citep{zheng2026egoscalescalingdexterousmanipulation,lingbot-va2026,lingbotvla2,ye2026worldactionmodelszeroshot}. In particular, they suit these models because the hand, the objects it acts on, and the wearer's motion appear in one view, and because such data can be collected at scale without instrumenting the environment. World-space hand geometry is a primary target of that supervision: hand placement and its attributes are defined relative to the scene rather than to the camera, so they stay fixed while the wearer moves. However, recovering this geometry is difficult: camera egomotion and hand--object occlusion corrupt the monocular cues, and the observations themselves provide no absolute scale. In addition, the scale of current collections makes throughput a first-order constraint: a method that processes a fraction of a recording per day determines how much supervision the data can provide. Errors in hand placement or interaction state then propagate into the trained model, where they are difficult to detect.

\begin{wrapfigure}{r}{0.50\textwidth}
  \vspace{-1.0em}
  \centering
  \captionsetup{justification=centering,singlelinecheck=false}
  \includegraphics[width=\linewidth]{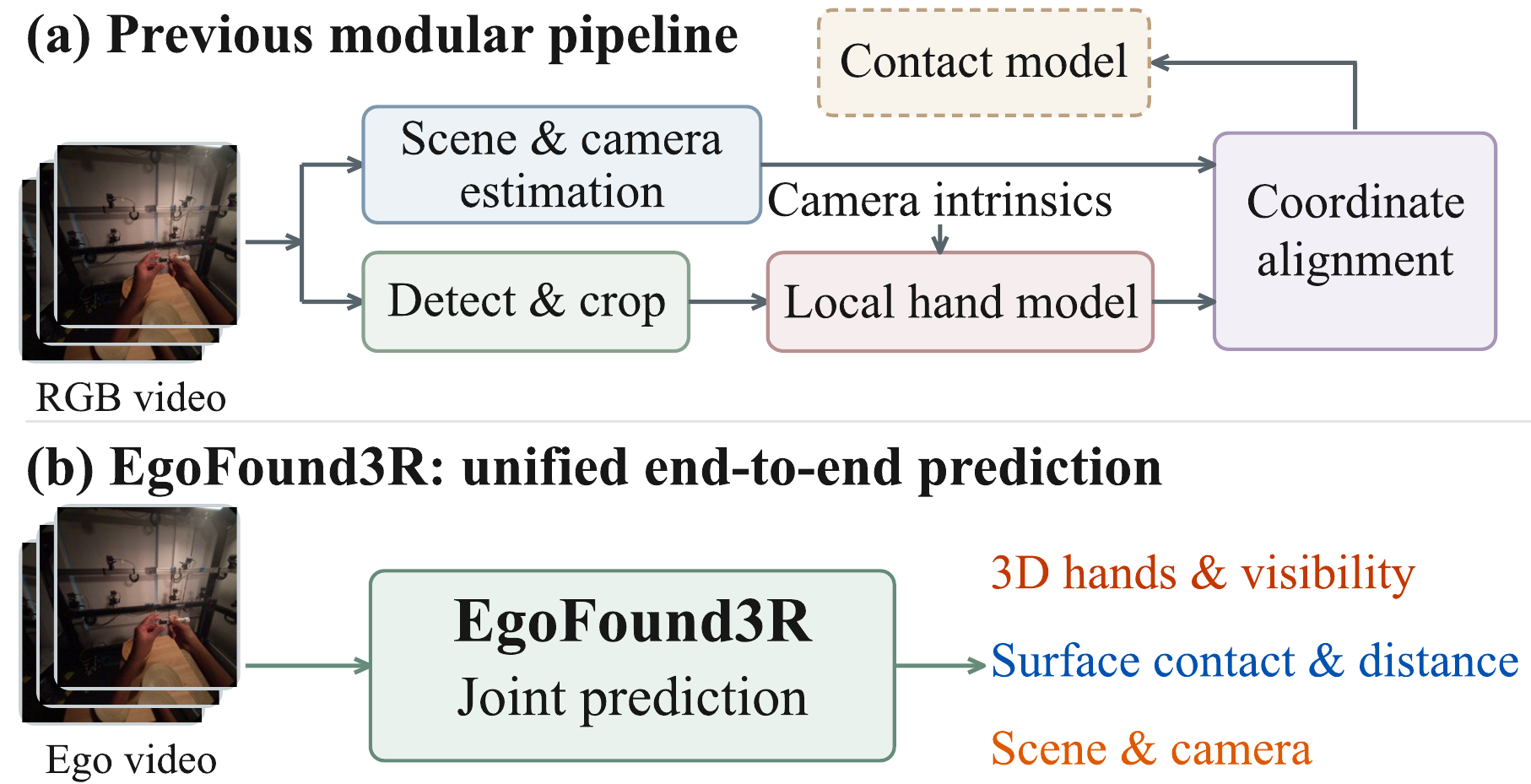}
  \caption{Pipeline vs.\ end-to-end.}
  \label{fig:task_overview}
\end{wrapfigure}

Recent estimators have improved hand pose accuracy in both camera and world space, but existing pipelines chain them in separate stages, as illustrated in \cref{fig:task_overview}(a), leaving both constraints unresolved. Specifically, camera-space estimators recover local hand geometry, world-space methods such as HaWoR~\citep{zhang2025hawor} and Dyn-HaMR~\citep{yu2025dyn} add camera tracking or motion completion, and contact is delegated to additional predictors or post-hoc optimization. The stages keep separate representations of the same evidence, so geometric priors are not shared, and errors accumulate across stages and over long sequences. Repeated visual encoding also lowers throughput, and since annotation is the bottleneck, this cost directly limits how much of a recording becomes supervision. Interaction attributes such as visibility, contact, and distance are likewise not estimated by any reconstruction model; they are left to separate predictors that take the hand as given, so errors in the hand or its placement propagate into the attributes rather than being corrected. Together, these limitations motivate a single model that places hands in world coordinates, predicts hand geometry and interaction attributes in the same pass, and does so at the throughput that annotation at scale demands.

Such a model can be built on geometry backbones pretrained on large-scale 3D data, which already encode the scene context that disambiguates an occluded hand~\citep{chen2025human3r,lyu2026reviv}. We therefore introduce \textbf{\method{}}, an end-to-end model that estimates world-space hand geometry and predicts point-wise visibility, contact, and distance in one pass. The model integrates three designs: \textbf{(i) Structured hand prompts.} An explicit hand representation is conditioned on a frozen VGGT-$\Omega$~\citep{wang2026vggtomega} backbone through structured prompts. Only the prompts, adapters, and prediction heads are trained, so inference needs only monocular video, with no depth, calibration, object template, or detector-and-crop stage. \textbf{(ii) Explicit hand representation.} Geometry and attributes are decoded from the same point features: the model regresses root-relative joints and vertices together with point-wise visibility, contact, and distance, and an observation-constrained root estimate places the hand in the scene. \textbf{(iii) Shared-parameter multi-rate design.} Hand observations need dense coverage while global scene aggregation does not, and global aggregation is the dominant cost in prior pipelines; we therefore run it at a configurable lower rate, reducing redundant computation without duplicating parameters.

We evaluate EgoFound3R on OakInk-v2, TACO, and HOI4D, where it reduces the mean per-joint position error (MPJPE) by 43.2\%, 22.4\%, and 11.6\% relative to previous methods and predicts point-wise contact and distance alongside the geometry in the same pass, while attaining approximately $6\times$ higher throughput than WiLoR~\citep{potamias2025wilor}, the previous state-of-the-art hand-reconstruction method.\par\noindent Our contributions are threefold:
\begin{itemize}
  \item We present EgoFound3R, an end-to-end model that reconstructs world-space hands and predicts point-wise visibility, contact, and distance in one pass, replacing the chained pipeline of hand estimation, camera tracking, and contact prediction with a single model.
  \item We show that interaction attributes can be predicted inside the reconstruction model: visibility, contact, and distance are decoded point-wise from the same features as the geometry, without separate task-specific models.
  \item At annotation scale, EgoFound3R runs $6\times$ faster than WiLoR while improving world-space hand accuracy on all three datasets, enabled by its shared-parameter multi-rate design.
\end{itemize}

\section{Related Work}
\label{sec:related_work}

\paragraph{Geometric priors and unified reconstruction.}
DUSt3R~\citep{Wang_2024_CVPR}, VGGT~\citep{wang2025vggt}, $\pi^3$~\citep{wang2026pi}, and Depth Anything 3~\citep{lin2025depth} establish feed-forward geometric reconstruction.
VGGT-$\Omega$~\citep{wang2026vggtomega}, CUT3R~\citep{wang2025continuous}, and LingBot-Map~\citep{chen2026geometric} extend this line toward scalable or persistent scene representations.
These priors have also been used for articulated reconstruction. Human3R~\citep{chen2025human3r} introduces human prompts and ReViV~\citep{lyu2026reviv} jointly predicts hand joints, cameras, and scene depth.
EgoFound3R adds interaction attributes to this joint reconstruction, decoding hand geometry and attributes from shared full-image features without a separate detection or crop stage.

\paragraph{Egocentric data for embodied models.}
Dexterous manipulation, world-modeling, vision-language-action, and world-action systems such as EgoScale~\citep{zheng2026egoscalescalingdexterousmanipulation}, LingBot-VA~\citep{lingbot-va2026}, LingBotVLA2~\citep{lingbotvla2}, and DreamZero~\citep{ye2026worldactionmodelszeroshot} are increasingly trained on egocentric data, where annotation throughput becomes a practical constraint. EgoFound3R addresses this constraint by producing world-space hand geometry, scene context, and point-wise visibility, contact, and distance in one model.

\paragraph{Hand geometry and motion.}
HaMeR~\citep{pavlakos2024reconstructing} and WiLoR~\citep{potamias2025wilor} estimate camera-space hand geometry; EgoForce~\citep{millerdurai2026egoforce} incorporates forearm crops and calibrated geometry for metric placement.
World-space methods combine hand and camera estimates, including HaWoR~\citep{zhang2025hawor} and Dyn-HaMR~\citep{yu2025dyn}.
Generative refinement further constrains motion: PAD-Hand~\citep{ismayilzada2026pad} applies physics-aware diffusion to WiLoR estimates, and StableHand~\citep{zeng2026stablehand} uses quality-aware flow matching with hand and camera conditioning.
Explicit representations such as CMR~\citep{chen2021camera} and MobRecon~\citep{chen2022mobrecon} regress hand-mesh coordinates directly.
EgoFound3R extends this direction by attaching point-wise interaction attributes to the same explicit representation.

\paragraph{Contact and interaction data.}
Contact estimation relies on geometric or visual evidence: ContactOpt~\citep{grady2021contactopt} and S$^2$Contact~\citep{tse2022s} classify contact from hand-object geometry, InteractVLM~\citep{dwivedi2025interactvlm} transfers image-based interaction cues, and EgoPHI~\citep{ilic2026egophi} predicts dense contact and force from additional geometry.
Beyond classification, WHOLE~\citep{ye2025whole} and EgoGrasp~\citep{fu2026egograsp} estimate world-space hand--object interaction.
For large-scale annotation, Do as I Do~\citep{paliwal2026doasido} builds on HaWoR, EgoInfinity~\citep{wang2026egoinfinity} refines WiLoR in a static-camera setting, and Open-AoE~\citep{li2026open} extracts supervision at scale.
Because these systems chain several learned models per video, cost accumulates with each added stage. EgoFound3R replaces the chain with one end-to-end model that predicts hand geometry and interaction attributes and requires no object meshes at inference.

\section{Method}
\label{sec:method}

\begin{figure}[t]
  \centering
  \includegraphics[width=\linewidth]{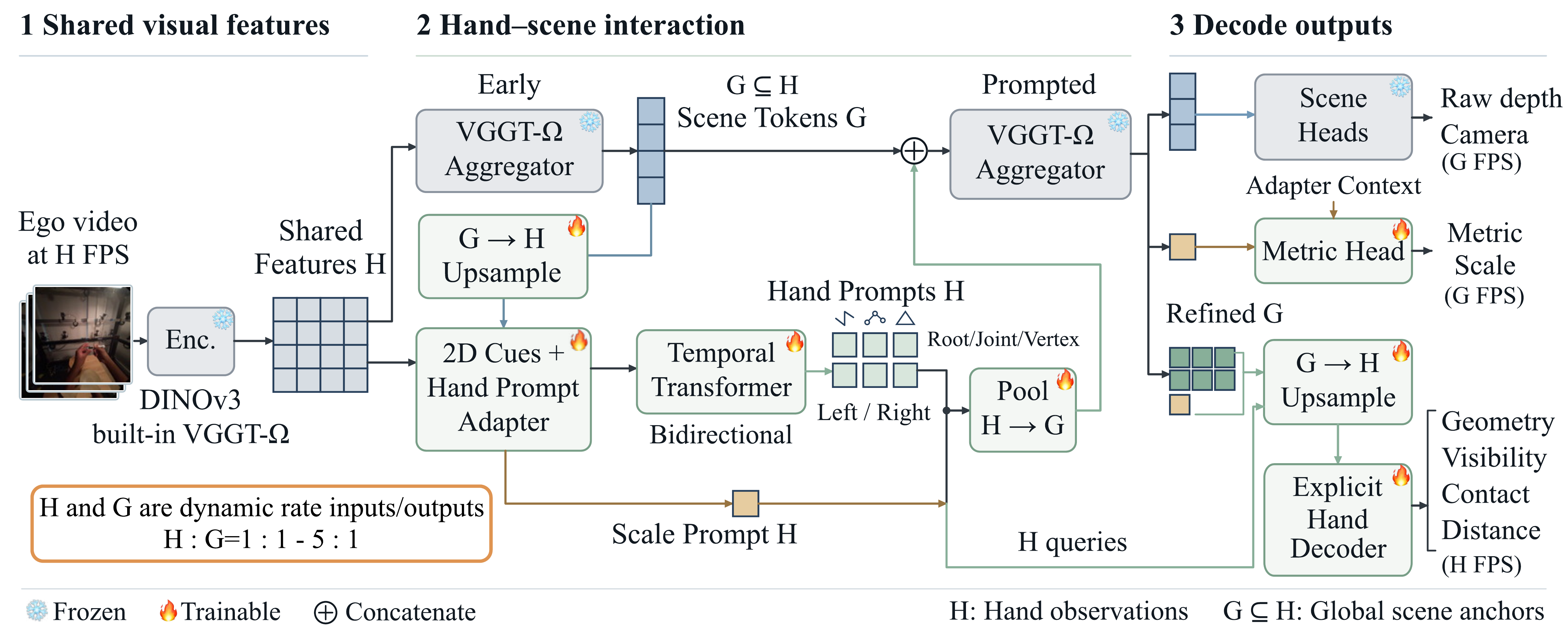}
  \caption{\textbf{Overview of \method{}.} Egocentric video input is encoded at the hand rate $H$, and the built-in DINOv3 features of the frozen VGGT-$\Omega$ backbone provide the shared visual features, from which an adapter builds structured root, joint, and surface prompts. A bidirectional temporal Transformer links the two hands across $H$, and pooling neighboring prompts yields global scene anchors $G\subseteq H$ at an adjustable $H{:}G$ ratio. The prompted VGGT-$\Omega$ aggregator drives the explicit hand head that decodes hand geometry, visibility, contact, and distance, and the scene and metric heads that place the hands in world coordinates.}
  \label{fig:framework}
\end{figure}

\subsection{Problem Formulation}
\label{sec:overview}
Given a monocular egocentric RGB video, we aim to recover hand geometry in world space together with point-wise interaction attributes, and to place the hands in the surrounding scene.
The setting is difficult in three respects: the camera moves with the wearer, so camera motion and hand motion are entangled; the hands are small and frequently occluded by the objects they hold; and a single monocular view carries no absolute scale.
The two sides, however, do not behave alike in time: hands change quickly and demand dense observation, whereas the scene evolves slowly while still providing the scale and camera motion that place them.
As illustrated in \cref{fig:framework}(1, 2), EgoFound3R therefore keeps both in a single feed-forward model and assigns them different temporal rates, in place of the detector, hand model, camera tracker, and contact predictor that prior pipelines run in sequence.
We write $H$ for the hand-observation times, $G\subseteq H$ for the sparser anchors at which the scene is updated, and $R$ for the requested output times. Hand root translation, joints, and surface points with their attributes are predicted at every $h\in H$, while the anchors $g\in G$ carry depth and camera motion that place them, and hand and camera outputs are returned at every time in $R$.

\subsection{Hand Prompts that Reuse Pretrained Geometry}
\label{sec:prompts}
The first design question is what a hand model should look at. A hand crop contains little beyond the hand, which is exactly the evidence that occlusion and camera motion corrupt, whereas a geometry backbone pretrained on large-scale 3D data already encodes the surrounding scene that disambiguates the hand.
We therefore keep such a backbone frozen---VGGT-$\Omega$~\citep{wang2026vggtomega}, whose built-in DINOv3 features~\citep{simeoni2025dinov3} provide the shared visual representation---and let the hands read from it through learned queries, as shown in \cref{fig:framework}(1), instead of regressing them from a crop in isolation.
The queries are organized by the question they ask: root queries locate the hand, joint queries describe its articulation, and surface queries describe what it touches, with a bidirectional temporal Transformer allowing the two hands to exchange information over the clip.
Detection is one more question of the same form, so the model estimates where and when each hand is visible from 2D heatmaps and in-view scores computed on the same features, and no external detector is required.
The queries are the only hand-specific inputs to the frozen backbone, whose geometric priors are reused rather than overwritten, and only the prompts, adapters, and prediction heads are trained; Appendix~\ref{app:detailed_framework} details the token layout.

\subsection{Cross-Rate Interaction between Hands and Scene}
\label{sec:crossrate}
\begin{figure}[t]
  \centering
  \includegraphics[width=\linewidth]{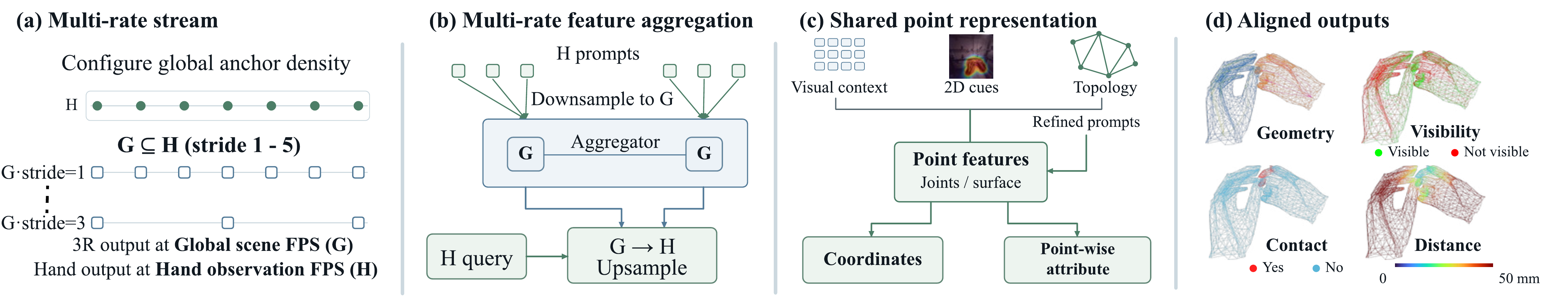}
  \vspace{-0.6em}
  \caption{\textbf{Cross-rate prompting and point decoding.} (a, b) Hand prompts run at rate $H$ while global anchors $G\subseteq H$ aggregate them at a configurable lower rate. (c, d) A shared point representation decodes geometry and point-wise attributes.}
  \label{fig:multirate}
  \vspace{-0.8em}
\end{figure}

Grounding the hands in the scene raises a second design question: how often must that scene context be refreshed? Re-aggregating the scene at the hand rate is a major source of cost in prior pipelines, while the hands still require dense observation, so we decouple the two rates and expose their ratio as a single user-facing knob, as shown in \cref{fig:multirate}(a, b).
A stride $s\in\{1,\ldots,5\}$ subsamples the hand times $H$ into the anchors $G$, and the hand prompts observed since the previous update are pooled into each anchor, so that the scene receives a compact summary of the hand motion it has missed:
\begin{equation}
 p_g=p_{a_g}+\gamma f\!\left(
 \operatorname{LN}(\bar p_g-p_{a_g}),\operatorname{LN}(p_{a_g})\right).
 \label{eq:anchor_pool}
\end{equation}

Here $\bar p_g$ is the pooled summary of the neighboring hand prompts, $p_{a_g}$ the anchor prompt before pooling, $p_g$ the updated anchor prompt, $f$ a residual projection, $\gamma$ a scalar gate, and $\operatorname{LN}$ layer normalization. The update is residual and zero-initialized, so training starts from the dense behavior and larger strides only thin the aggregation.
The anchors then re-enter the frozen aggregator, and a second bridge returns the aggregated context to the hand rate, so evidence flows in both directions.
Because every stride reuses the same adapters and phase embedding, the ratio acts as a computation knob rather than defining a new model.

\subsection{Explicit Hand Geometry with Interaction Attributes}
\label{sec:hand_decoding}
The hand features now carry scene context and 2D hand evidence, but they still have to be turned into geometry and interaction attributes. The attributes we target are defined on individual hand points---whether a point is visible, whether it touches, and how far it lies from the contact surface---so we decode point coordinates directly instead of describing the hand through pose and shape coefficients, as shown in \cref{fig:multirate}(c, d).
Root, joint, and surface queries read the same point features, which keeps placement, articulation, and contact aligned; the surface branch additionally inherits features from associated joints and mixes over incident faces, so the decoded points respect the connectivity of the hand.
Each point then receives three attributes from a small prediction head: visibility, contact, and distance to the contact surface.
Contact merges object contact and opposite-hand contact into one per-point state that never names the partner, so no object template or object pose is required. The hand stays supervised through MANO topology~\citep{MANO:SIGGRAPHASIA:2017}, yet its geometry is decoded as point coordinates rather than pose or shape parameters.

\begin{figure}[!ht]
  \centering
  \includegraphics[width=\linewidth]{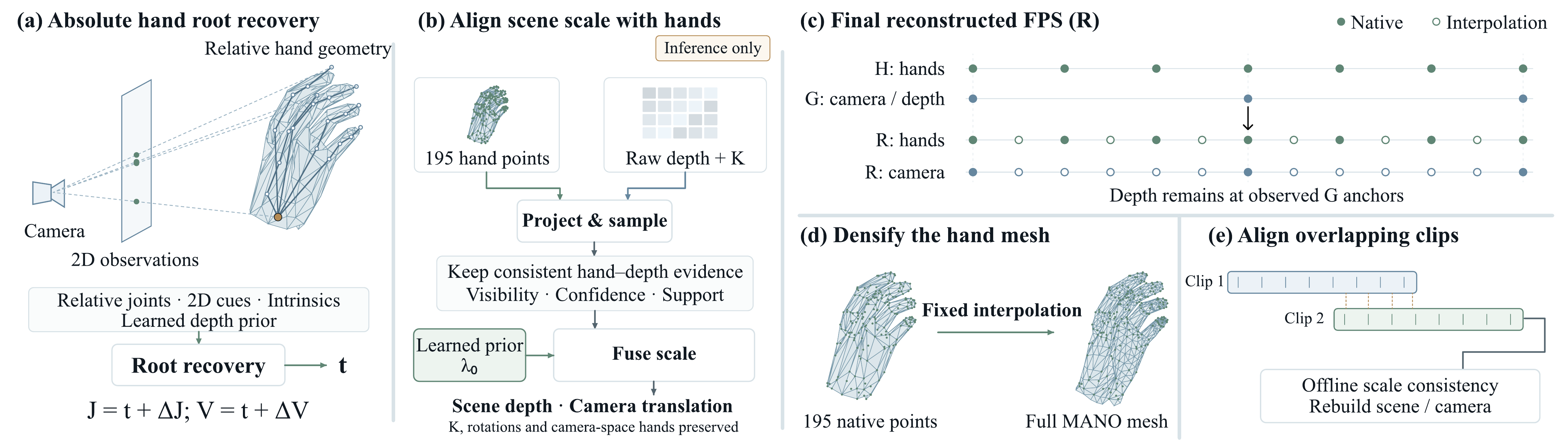}
  \vspace{-0.6em}
  \caption{\textbf{Root recovery, scale alignment, and outputs.} (a) Root recovery from 2D rays, relative joints, and a learned depth prior. (b) Inference-time scale alignment, with camera-space hands and rotations fixed. (c) Outputs at requested times $R$; (d) 195$\to$MANO densification; (e) scale-consistent clip stitching.}
  \label{fig:scale_recovery}
\end{figure}

\subsection{Metric Hand Placement}
\label{sec:metric_scene}
Decoding hand geometry does not yet place the hand in the camera, and the depth of a monocular scene is defined only up to an unknown factor. Both are questions of scale, and the hands answer both.
As shown in \cref{fig:scale_recovery}(a), the hand root is recovered from the observed 2D rays and the predicted relative joints under the wrist constraint, with a learned depth prior stabilizing the solve:
\begin{equation}
 J_j=t+\Delta J_j,\qquad V_v=t+\Delta V_v,
 \label{eq:absolute_hand}
\end{equation}
where $t$ is the root translation, $\Delta J_j$ and $\Delta V_v$ the predicted offsets of joint $j$ and vertex $v$, and $J_j$ and $V_v$ the resulting point coordinates, which the predicted cameras finally express in world space.

The same evidence sets the metric scale of the scene, as shown in \cref{fig:scale_recovery}(b): a metric head predicts it from scene summaries together with hand features, a learned prior keeps it well defined when hand evidence is weak, and at inference the agreement between the projected hand vertices and the raw depth refines it further.
Calibration rescales scene depth and camera translation only, so rotations, intrinsics, and the camera-space hands stay unchanged and the hand branch is never recomputed; Appendix~\ref{app:scene} reports the scene and camera outputs.

\subsection{Training and Outputs at Arbitrary Times}
\label{sec:learning}
The designs above form a single model that is trained end-to-end on a mixture of egocentric datasets, with the backbone frozen so that only the prompts, adapters, and prediction heads are optimized; Appendix~\ref{app:training} states the resulting objectives.
The network runs at the hand rate, but applications need outputs at their own times and at the mesh resolution their downstream use requires. We therefore treat the requested times $R$ as an interface and recover them from the network outputs by deterministic interpolation, as shown in \cref{fig:scale_recovery}(c), and the same stage densifies the native 195-vertex geometry to the full 778-vertex MANO mesh, as shown in \cref{fig:scale_recovery}(d). Dense outputs thus cost no additional network computation, and the predicted points are preserved by construction.
Long videos are processed as overlapping clips that must share one scale before stitching, which the same stage restores by rescaling their reconstructions to agree on the shared frames, as shown in \cref{fig:scale_recovery}(e).
Because interpolation uses future frames and the alignment compares whole clips, EgoFound3R is an offline method, and both operations remain outside the feed-forward network.

\section{Experiments}
\label{sec:experiments}

\subsection{Evaluation Setting}
We train on seven egocentric datasets: H2O~\citep{Kwon_2021_ICCV}, HOT3D Aria~\citep{banerjee2025hot3d}, OakInk-v2~\citep{zhan2024oakink2}, TACO~\citep{liu2024taco}, HOI4D~\citep{Liu_2022_CVPR}, ARCTIC~\citep{fan2023arctic}, and STERA-10M~\citep{palanisamy2026mobileego}. Evaluation uses 2{,}378 windows sampled at a fixed ratio per dataset, each a 2\,s clip and held out from training; the main tables report OakInk-v2, TACO, and HOI4D, where the full metric suite is available, and the remaining datasets appear in the qualitative comparisons. Our model uses the default stride $s=5$; we defer the temporal and scene studies to Appendices~\ref{app:efficiency} and~\ref{app:scene}.
To bound evaluation cost, Dyn-HaMR and InteractVLM are each evaluated on a 100-window subset paired with an \method{} control on the same windows.
Geometry is reported in millimeters as mean per joint position error (MPJPE) and mean per vertex position error (MPVPE) under six alignments: unaligned (Raw), root-relative (RR), per-frame aligned (PA), similarity-aligned ($\mathrm{Sim}(3)$), world (W), and world-aligned (WA). Contact and visibility use precision (P), recall (R), and F1, and distance uses the mean absolute error (MAE, mm) of the point-to-surface distance, reported over all valid points (MAE(All)), predicted contacts (MAE(Pred)), and ground-truth contacts (MAE(GT)). See Appendix~\ref{app:metrics} for details.

\subsection{Hand Geometry and Interaction}
\paragraph{Hand geometry.}
\cref{tab:joint} reports world-space hand reconstruction, comparing \method{} with camera-space estimators (WiLoR~\citep{potamias2025wilor} and its refinements PAD-Hand~\citep{ismayilzada2026pad} and EgoForce~\citep{millerdurai2026egoforce}), world-space pipelines (HaWoR~\citep{zhang2025hawor} and Dyn-HaMR~\citep{yu2025dyn}, the latter on matched 100-window subsets), and the joint reconstruction system ReViV~\citep{lyu2026reviv}.
\method{} attains the lowest raw and root-relative joint error on every subset, reducing MPJPE by 43.2\%, 22.4\%, and 11.6\% relative to the strongest baseline on OakInk-v2, TACO, and HOI4D; the ordering survives the removal of the root translation, and on the common windows the model also attains the lowest world and world-aligned error among the methods that predict their own camera poses.
The three benchmarks differ in capture rigs and object vocabularies, so the consistent ordering across them reflects the design rather than a single benchmark, and the baselines fail as their designs predict:
Camera-space estimators recover shape once the prediction is aligned per frame, but their raw and world errors stay high because they never estimate where the camera sits in the scene.
World-space pipelines place the hands but drift as sequences grow: on the matched subsets Dyn-HaMR's world error rises several-fold from OakInk-v2 to HOI4D, HaWoR loses articulation on OakInk-v2, where its raw error is more than twice its value on the other two datasets, and ReViV is the weakest overall, with per-frame aligned errors several times ours on every subset.
Our single-pass design targets this failure mode: hands and scene are decoded by one model, so the hand trajectory inherits one scale across the clip instead of the drift of chained stages; Appendices~\ref{app:surface} and~\ref{app:temporal} report the vertex-level and temporal results.

\paragraph{Contact prediction.}
\cref{tab:contact} reports joint-level contact prediction, comparing \method{} with InteractVLM~\citep{dwivedi2025interactvlm}, S$^2$Contact~\citep{tse2022s}, and ContactOpt~\citep{grady2021contactopt}. \method{} predicts contact together with the hand geometry, on the same points and from the same scene-aware features rather than by a separate classifier, and it is the only compared method that also predicts a per-point distance to the contact surface.
\method{} attains the highest precision, recall, and F1 on all three subsets.
The baselines fail in opposite directions: the geometric classifiers keep precision moderate but recover few contacting joints, whereas InteractVLM trades precision for recall.
Decoding contact on the points that carry the geometry, from features that already encode the surrounding scene, improves both sides at once.
The compared systems use different contact conventions and the two geometric baselines additionally receive ground-truth geometry; Appendix~\ref{app:contact} states the protocol and the sensitivity analysis.

\paragraph{Visibility.}
The same head predicts point visibility, which the compared pipelines do not report: as shown in \cref{tab:visibility}, \method{} surpasses the joint-only predictor HVD~\citep{hara2026handvisibilitydetector}, in joint-visibility F1 on all three datasets, with higher precision on two subsets and higher recall on two others, and it also predicts vertex-level visibility; Appendix~\ref{app:contact} reports the vertex-level results.

\begin{table}[!t]
\centering
\captionsetup{skip=4pt,justification=raggedright,singlelinecheck=false}
\caption{\textbf{Joint reconstruction} (MPJPE, mm); each cell lists OakInk-v2, TACO, and HOI4D left to right, with $\dagger$ marking methods placed with ground-truth extrinsics and $\ddagger$ marking \method{} with ground-truth intrinsics and predicted extrinsics. For \method{}$^{\dagger}$ the camera-space columns are identical to the default row and are left blank. $^{100\mathrm{w}}$ subsets are unranked, W/WA are ranked within each extrinsics group.}
\label{tab:joint}
\small
\renewcommand{\arraystretch}{1.05}
\setlength{\tabcolsep}{2pt}
\setlength{\aboverulesep}{1.2pt}
\setlength{\belowrulesep}{1.2pt}
\setlength{\fboxsep}{0.6pt}
\resizebox{\linewidth}{!}{%
\begin{tabular}{@{}lr@{\hspace{2.4pt}\textcolor{black!70}{\rule[-0.22ex]{0.8pt}{1.8ex}}\hspace{2.4pt}}r@{\hspace{2.4pt}\textcolor{black!70}{\rule[-0.22ex]{0.8pt}{1.8ex}}\hspace{2.4pt}}r@{\hspace{8pt}}r@{\hspace{2.4pt}\textcolor{black!70}{\rule[-0.22ex]{0.8pt}{1.8ex}}\hspace{2.4pt}}r@{\hspace{2.4pt}\textcolor{black!70}{\rule[-0.22ex]{0.8pt}{1.8ex}}\hspace{2.4pt}}r@{\hspace{8pt}}r@{\hspace{2.4pt}\textcolor{black!70}{\rule[-0.22ex]{0.8pt}{1.8ex}}\hspace{2.4pt}}r@{\hspace{2.4pt}\textcolor{black!70}{\rule[-0.22ex]{0.8pt}{1.8ex}}\hspace{2.4pt}}r@{\hspace{8pt}}r@{\hspace{2.4pt}\textcolor{black!70}{\rule[-0.22ex]{0.8pt}{1.8ex}}\hspace{2.4pt}}r@{\hspace{2.4pt}\textcolor{black!70}{\rule[-0.22ex]{0.8pt}{1.8ex}}\hspace{2.4pt}}r@{\hspace{8pt}}r@{\hspace{2.4pt}\textcolor{black!70}{\rule[-0.22ex]{0.8pt}{1.8ex}}\hspace{2.4pt}}r@{\hspace{2.4pt}\textcolor{black!70}{\rule[-0.22ex]{0.8pt}{1.8ex}}\hspace{2.4pt}}r@{\hspace{8pt}}r@{\hspace{2.4pt}\textcolor{black!70}{\rule[-0.22ex]{0.8pt}{1.8ex}}\hspace{2.4pt}}r@{\hspace{2.4pt}\textcolor{black!70}{\rule[-0.22ex]{0.8pt}{1.8ex}}\hspace{2.4pt}}r@{}}
\toprule
\multicolumn{19}{c}{\scriptsize \textit{OakInk-v2} \,$\cdot$\, \textit{TACO} \,$\cdot$\, \textit{HOI4D}} \\
\midrule
Method & \multicolumn{3}{c}{Raw$\downarrow$} & \multicolumn{3}{c}{RR$\downarrow$} & \multicolumn{3}{c}{PA$\downarrow$} & \multicolumn{3}{c}{$\mathrm{Sim}(3)\!\downarrow$} & \multicolumn{3}{c}{W$\downarrow$} & \multicolumn{3}{c}{WA$\downarrow$} \\
\midrule
WiLoR$^\dagger$ & 91.24 & 70.40 & 81.51 & 27.66 & 28.33 & 27.75 & \bestv{7.78} & \secondv{9.49} & 9.62 & \bestv{16.89} & 25.95 & 24.14 & \secondv{42.14} & \secondv{51.14} & \bestv{49.41} & \bestv{17.04} & \secondv{26.14} & \secondv{24.57} \\
PAD-Hand$^\dagger$ & 107.16 & 72.97 & 75.35 & 32.23 & 30.30 & 29.77 & 9.63 & 11.19 & 10.12 & 21.37 & 29.21 & 24.79 & 49.01 & 57.06 & \secondv{51.92} & 21.62 & 29.60 & 25.04 \\
EgoForce$^\dagger$ & 123.57 & 126.63 & 147.77 & 47.75 & 44.53 & 52.78 & 11.09 & 11.93 & 12.92 & 32.62 & 47.84 & 49.22 & 80.59 & 105.52 & 142.58 & 33.10 & 46.65 & 55.76 \\
\rowcolor{black!4}
\method{}$^\dagger$ & \multicolumn{3}{c}{--} & \multicolumn{3}{c}{--} & \multicolumn{3}{c}{--} & \multicolumn{3}{c}{--} & \bestv{27.49} & \bestv{36.68} & 63.01 & \secondv{17.89} & \bestv{24.06} & \bestv{23.30} \\
\cmidrule(lr){14-19}
Dyn-HaMR$^{100\mathrm{w}}$ & 96.39 & 127.12 & 245.06 & 17.91 & 27.47 & 26.49 & 7.84 & 11.62 & 11.89 & 13.76 & 29.27 & 30.47 & 36.16 & 77.88 & 173.62 & 17.19 & 31.46 & 65.71 \\
HaWoR & 174.73 & 74.28 & 53.46 & 45.85 & 24.19 & 28.47 & 9.49 & 10.55 & 10.49 & 32.19 & 27.71 & 27.85 & 100.46 & 86.38 & 171.09 & 37.47 & 35.34 & 55.96 \\
ReViV & 253.68 & 91.19 & 357.46 & 80.60 & 62.48 & 82.12 & 44.14 & 40.14 & 39.66 & 58.05 & 54.08 & 58.04 & 123.25 & 93.25 & 182.33 & 57.75 & 54.26 & 68.74 \\
\rowcolor{black!4}
\method{} & \secondv{51.81} & \secondv{54.64} & \secondv{47.28} & \bestv{17.77} & \secondv{18.54} & \bestv{16.63} & \secondv{8.02} & \bestv{8.13} & \bestv{7.99} & \secondv{17.87} & \bestv{24.15} & \bestv{23.29} & \bestv{30.88} & \bestv{40.03} & \secondv{159.70} & \bestv{18.77} & \bestv{25.16} & \secondv{49.84} \\
\rowcolor{black!4}
\method{}$^\ddagger$ & \bestv{39.45} & \bestv{50.44} & \bestv{43.43} & \secondv{17.84} & \bestv{18.52} & \secondv{16.67} & 8.04 & \bestv{8.13} & \secondv{8.00} & 18.01 & \secondv{24.63} & \secondv{23.39} & \secondv{31.56} & \secondv{41.78} & \bestv{159.18} & \secondv{18.94} & \secondv{25.75} & \bestv{49.77} \\
\bottomrule
\end{tabular}%
}
\end{table}

\begin{table}[!t]
\centering
\captionsetup{skip=4pt,justification=raggedright,singlelinecheck=false}
\caption{\textbf{Joint-level contact prediction}. Superscripts mark 100-window subsets. InteractVLM is shown for completeness and does not enter the ranking.}
\label{tab:contact}
\small
\renewcommand{\arraystretch}{1.0}
\setlength{\aboverulesep}{1.5pt}
\setlength{\belowrulesep}{1.5pt}
\setlength{\tabcolsep}{3pt}
\begin{tabularx}{\linewidth}{l*{3}{>{\centering\arraybackslash}X}@{\hspace{6pt}}*{3}{>{\centering\arraybackslash}X}@{\hspace{6pt}}*{3}{>{\centering\arraybackslash}X}}
\toprule
& \multicolumn{3}{c}{OakInk-v2} & \multicolumn{3}{c}{TACO} & \multicolumn{3}{c}{HOI4D} \\
\cmidrule(lr){2-4}\cmidrule(lr){5-7}\cmidrule(lr){8-10}
Method & \multicolumn{1}{c}{P$\uparrow$} & \multicolumn{1}{c}{R$\uparrow$} & \multicolumn{1}{c}{F1$\uparrow$} & \multicolumn{1}{c}{P$\uparrow$} & \multicolumn{1}{c}{R$\uparrow$} & \multicolumn{1}{c}{F1$\uparrow$} & \multicolumn{1}{c}{P$\uparrow$} & \multicolumn{1}{c}{R$\uparrow$} & \multicolumn{1}{c}{F1$\uparrow$} \\
\midrule
InteractVLM$^{100\mathrm{w}}$ & 0.180 & 0.403 & 0.202 & 0.388 & 0.705 & 0.472 & 0.313 & 0.537 & 0.281 \\
S$^2$Contact & 0.327 & 0.081 & 0.079 & 0.519 & 0.104 & 0.146 & 0.444 & 0.265 & 0.284 \\
ContactOpt & \secondv{0.334} & \secondv{0.196} & \secondv{0.154} & \secondv{0.557} & \secondv{0.197} & \secondv{0.259} & \secondv{0.518} & \secondv{0.477} & \secondv{0.435} \\
\rowcolor{black!4}
\method{} & \bestv{0.510} & \bestv{0.665} & \bestv{0.493} & \bestv{0.654} & \bestv{0.777} & \bestv{0.697} & \bestv{0.612} & \bestv{0.835} & \bestv{0.675} \\
\bottomrule
\end{tabularx}
\end{table}

\begin{table}[!t]
\centering
\captionsetup{skip=4pt,justification=raggedright,singlelinecheck=false}
\caption{\textbf{Visibility prediction} at 21 joints on OakInk-v2, TACO, and HOI4D. HVD predicts joint visibility only; vertex-level results are reported in Appendix~\ref{app:contact}.}
\label{tab:visibility}
\small
\renewcommand{\arraystretch}{1.0}
\setlength{\aboverulesep}{1.5pt}
\setlength{\belowrulesep}{1.5pt}
\setlength{\tabcolsep}{2pt}
\begin{tabularx}{\linewidth}{l*{9}{>{\centering\arraybackslash}X}}
\toprule
& \multicolumn{3}{c}{OakInk-v2} & \multicolumn{3}{c}{TACO} & \multicolumn{3}{c}{HOI4D} \\
\cmidrule(lr){2-4}\cmidrule(lr){5-7}\cmidrule(lr){8-10}
Method & P$\uparrow$ & R$\uparrow$ & F1$\uparrow$ & P$\uparrow$ & R$\uparrow$ & F1$\uparrow$ & P$\uparrow$ & R$\uparrow$ & F1$\uparrow$ \\
\midrule
HVD & 0.823 & \bestv{0.808} & 0.792 & 0.878 & 0.834 & 0.853 & \bestv{0.917} & 0.769 & 0.831 \\
\rowcolor{black!4}
\method{} & \bestv{0.919} & 0.747 & \bestv{0.802} & \bestv{0.917} & \bestv{0.850} & \bestv{0.880} & 0.892 & \bestv{0.884} & \bestv{0.885} \\
\bottomrule
\end{tabularx}
\end{table}

\subsection{World-Space Qualitative Comparison}
\begin{figure}[t]
  \centering
  \includegraphics[width=\linewidth]{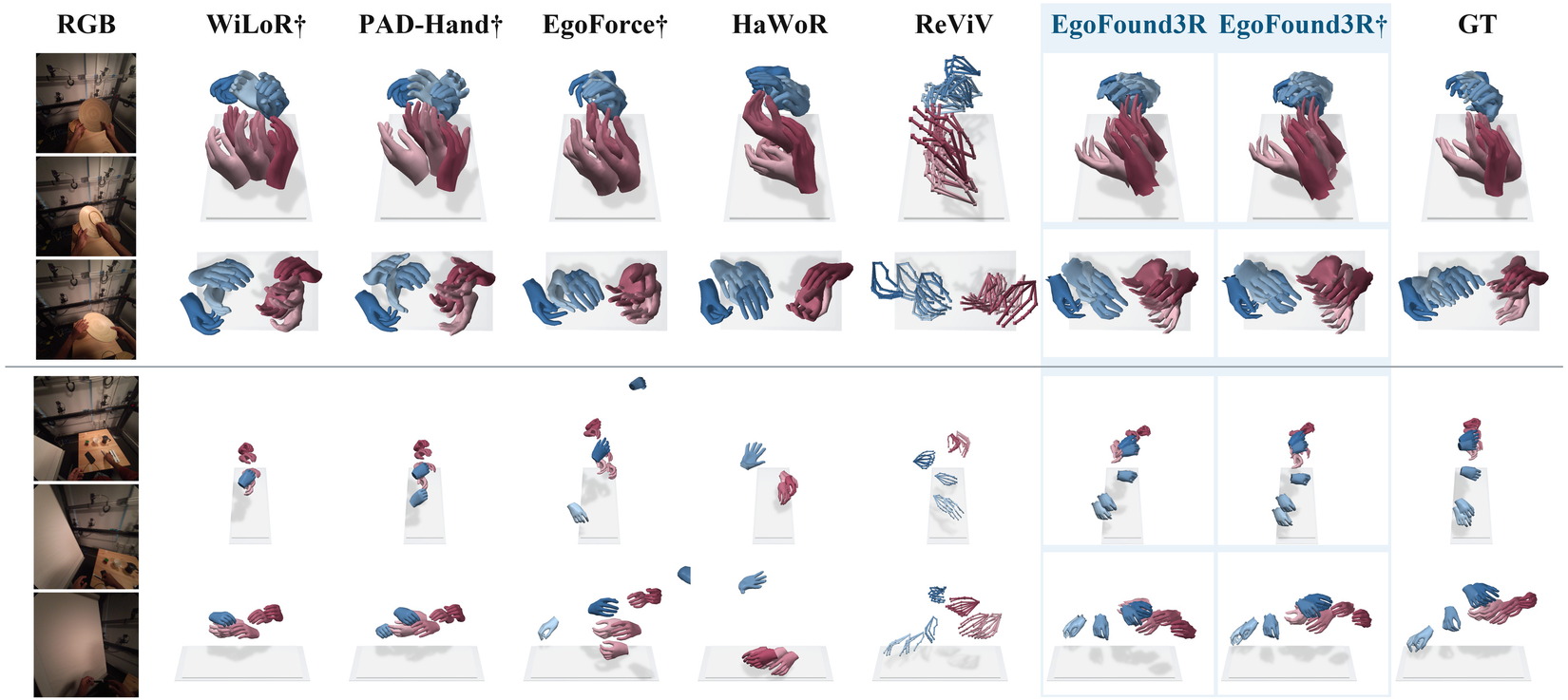}
  \caption{\textbf{World-space qualitative comparison.} Rows: two viewpoints of each clip; columns: the compared methods, rendered in the reconstructed
  world frame. Hands are pink (left) and blue (right), darker over time; $\dagger$ marks methods placed
  with ground-truth extrinsics.}
  \label{fig:world3d_main}
\end{figure}

\cref{fig:world3d_main} shows two clips rendered in the reconstructed world frame from two fixed viewpoints, so the comparison includes where the cameras place the hands rather than a single projected view.
Camera-space estimators recover the hand shape but offset it relative to the object surface, and both world-space pipelines drift away from it instead of recovering the hand motion; ReViV predicts joints only and provides no hand surface.
\method{} reproduces the ground-truth layout with its own predicted cameras; Appendix~\ref{app:world3d} shows additional results.

\subsection{Camera-Space Qualitative Comparison}
\begin{figure}[t]
  \centering
  \includegraphics[width=\linewidth]{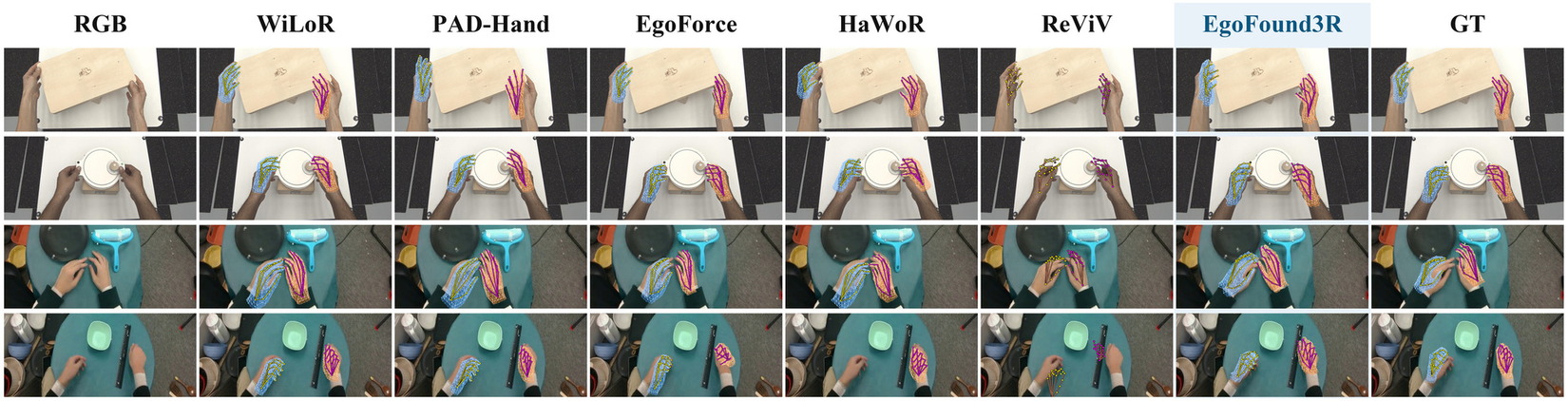}\\[3pt]
  \includegraphics[width=\linewidth]{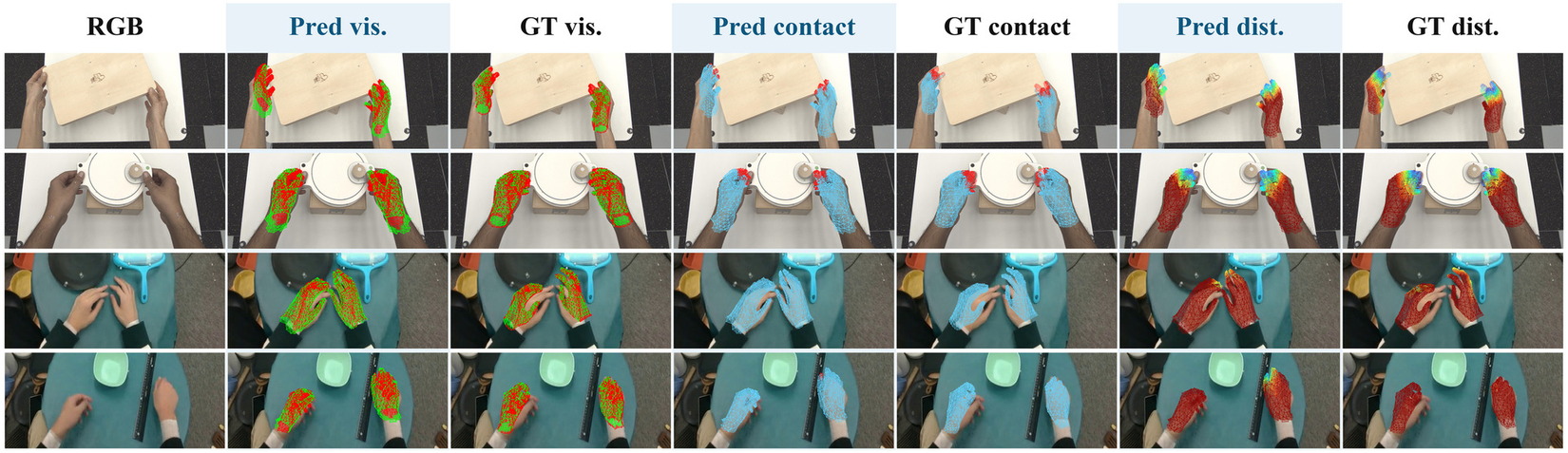}
  \caption{\textbf{Camera-space qualitative comparison.} Top: hand geometry (left/right hands in pink/blue). Bottom: visibility, contact, and distance for \method{} (\emph{Pred}) and the ground truth (\emph{GT}); green/red = visible/occluded, red/light blue = contact/no contact, and distance is a blue-to-red 0--50\,mm map.}
  \label{fig:qualitative_main}
  \vspace{-0.8em}
\end{figure}

\cref{fig:qualitative_main} shows four held-out windows, two from ARCTIC and two from TACO; both halves are 2D overlays on the observed pixels, with hand geometry on top and visibility, contact, and distance below.
On all four windows \method{} keeps the hand aligned with the object surface and the attribute maps follow the ground truth, consistent with \cref{tab:joint} and \cref{tab:contact}.
In these windows every compared method is displaced from the annotated hand and misestimates its pose and size, whereas our overlay tracks the annotation.
The attribute heads read the same points that carry the geometry; Appendix~\ref{app:qualitative} shows additional results.

\subsection{Core Design Ablation}
We ablate the design choices on TACO; Appendix~\ref{app:ablation} defines each variant and repeats the study on the three evaluation datasets. (i)~\textbf{Model capacity.} Reducing the hand module to 0.05B and 0.1B parameters raises raw error monotonically, and the reduced variants are also the weakest in the contact and visibility columns and among the weakest in distance, so the representation benefits from capacity rather than saturating early. (ii)~\textbf{Pretrained backbone.} Removing the frozen VGGT-$\Omega$ backbone (w/o $\Omega$) raises raw error by about half over the default and costs the distance columns, which supports reusing pretrained geometric priors rather than relearning them from the supervision alone. (iii)~\textbf{Hand--scene exchange.} Restricting the exchange to one direction (1-way) degrades geometry and attributes alike, and it costs more contact F1 than removing the backbone does, indicating that hand evidence must also shape the scene representation rather than only read from it. (iv)~\textbf{MANO parameters.} Replacing the explicit point representation with MANO parameter regression degrades every column relative to the default, and it is the weakest of the three structural variants in geometry, visibility, and distance, which is consistent with decoding attributes on the same points that carry the geometry. (v)~\textbf{Multi-rate default.} S5 is the default configuration of \method{}, and the stride ablation for the other rates is reported in Appendix~\ref{app:ablation}.

\begin{table}[!htbp]
\centering
\scriptsize
\captionsetup{skip=2pt,justification=raggedright,singlelinecheck=false}
\caption{\textbf{Core design ablation} on TACO; the F1 columns are measured at 21 joints, and S$k$ is \method{} at stride $k$.}
\label{tab:ablation_main}
\setlength{\tabcolsep}{1.2pt}
\renewcommand{\arraystretch}{0.80}
\begin{tabularx}{\linewidth}{l*{9}{>{\centering\arraybackslash}X}}
\toprule
Variant & Raw$\downarrow$ & RR$\downarrow$ & W$\downarrow$ & WA$\downarrow$ & Contact F1$\uparrow$ & Vis. F1$\uparrow$ & MAE(All)$\downarrow$ & MAE(Pred)$\downarrow$ & MAE(GT)$\downarrow$ \\
\midrule
0.05B & 109.71 & 36.74 & 81.37 & 42.66 & 0.542 & 0.800 & 41.47 & 45.30 & 49.54 \\
0.1B & 96.90 & 30.55 & 61.66 & 35.81 & 0.579 & 0.801 & 49.27 & 52.15 & 58.61 \\
MANO & 86.66 & 24.29 & 57.45 & 35.15 & 0.643 & 0.854 & 45.46 & 48.21 & 51.97 \\
w/o $\Omega$ & 82.59 & \secondv{23.49} & \secondv{53.38} & 32.33 & \secondv{0.661} & 0.856 & 36.19 & 40.42 & 42.52 \\
1-way & \secondv{78.82} & 23.70 & 54.16 & \secondv{32.14} & 0.628 & \secondv{0.861} & \secondv{32.79} & \secondv{35.52} & \secondv{37.29} \\
\rowcolor{black!4}S5 & \bestv{54.64} & \bestv{18.54} & \bestv{40.03} & \bestv{25.16} & \bestv{0.697} & \bestv{0.880} & \bestv{26.26} & \bestv{28.09} & \bestv{30.47} \\
\bottomrule
\end{tabularx}
  \vspace{-0.8em}
\end{table}

\subsection{Inference Efficiency}
\label{sec:efficiency}
At annotation scale, the cost of a single pass matters as much as its accuracy.

\begin{figure}[!ht]
  \centering
  \includegraphics[width=0.87\linewidth]{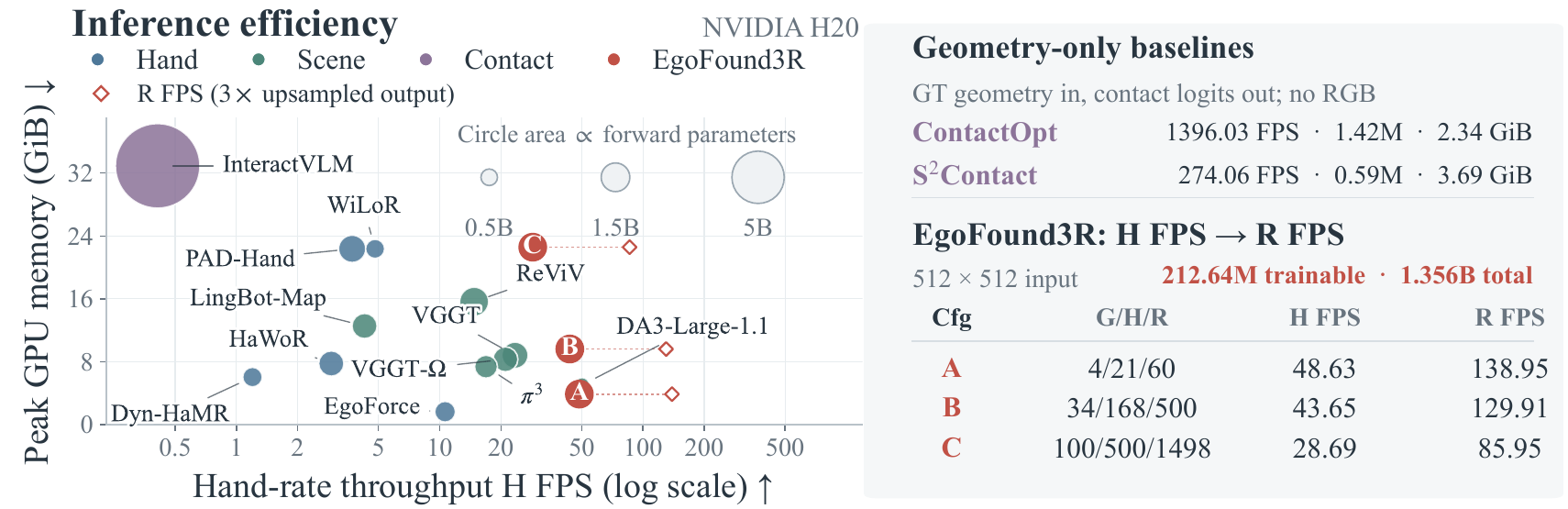}
  \captionsetup{skip=7pt}
  \caption{\textbf{Inference efficiency.}}
  \label{fig:efficiency_comparison}
\end{figure}

We report the hand-rate throughput (H FPS), the hand-observation frames processed per second and therefore the rate single-rate baselines report, together with the dense output rate (R FPS) that $3\times$ temporal upsampling yields.
As shown in \cref{fig:efficiency_comparison} and \cref{tab:inference_efficiency}, \method{} processes 28.69 H FPS at $512\times512$, $5.97\times$ the 4.81 FPS of WiLoR and faster than every compared hand pipeline.
One forward pass determines throughput, not the sum of chained stages.
See Appendix~\ref{app:efficiency} for details.

\section{Conclusion}
\label{sec:conclusion}
We presented \method{}, a unified end-to-end model that recovers hands in world space and predicts point-wise visibility, contact, and distance in a single pass.
The model combines structured prompts that carry pretrained geometric context into hand decoding, an explicit hand representation that decodes geometry and attributes, and a shared-parameter multi-rate design that lowers inference cost.
On OakInk-v2, TACO, and HOI4D, it reduces MPJPE by 43.2\%, 22.4\%, and 11.6\% over the strongest baseline and runs $6\times$ faster than the previous state of the art.

\bibliography{EgoFound3R_references}
\bibliographystyle{iclr2027_conference}

\clearpage
\appendix
\raggedbottom
\setcounter{topnumber}{3}
\setcounter{bottomnumber}{3}
\setcounter{totalnumber}{4}
\renewcommand{\topfraction}{0.92}
\renewcommand{\bottomfraction}{0.70}
\renewcommand{\textfraction}{0.06}
\renewcommand{\floatpagefraction}{0.80}
\makeatletter
\setlength{\@fptop}{0pt}
\setlength{\@fpbot}{0pt plus 1fil}
\makeatother
\section*{Appendix Contents}
\begingroup
\hypersetup{hidelinks}
\newcommand{\appendixentry}[2]{\noindent\hyperref[#1]{\ref*{#1}\quad #2}\dotfill\pageref{#1}\par}
\appendixentry{app:detailed_framework}{Detailed Architecture and Information Flow}
\appendixentry{app:training}{Training Objectives and Supervision}
\appendixentry{app:efficiency}{Inference Efficiency and Temporal Rates}
\appendixentry{app:metrics}{Evaluation Metrics and Ranking Rules}
\appendixentry{app:surface}{Additional Hand Geometry Evaluation}
\appendixentry{app:temporal}{Temporal Evaluation}
\appendixentry{app:contact}{Contact and Visibility Evaluation}
\appendixentry{app:matched_windows}{Matched-Subset Controls for 100-Window Baselines}
\appendixentry{app:ablation}{Architecture and Multi-Rate Ablations}
\appendixentry{app:scene}{Scene and Camera Evaluation}
\appendixentry{app:qualitative}{Camera-Space Qualitative Results}
\appendixentry{app:world3d}{World-Space Qualitative Results}
\endgroup
\medskip

\section{Detailed Architecture and Information Flow}
\label{app:detailed_framework}
\begin{figure}[H]
  \centering
  \includegraphics[width=\linewidth]{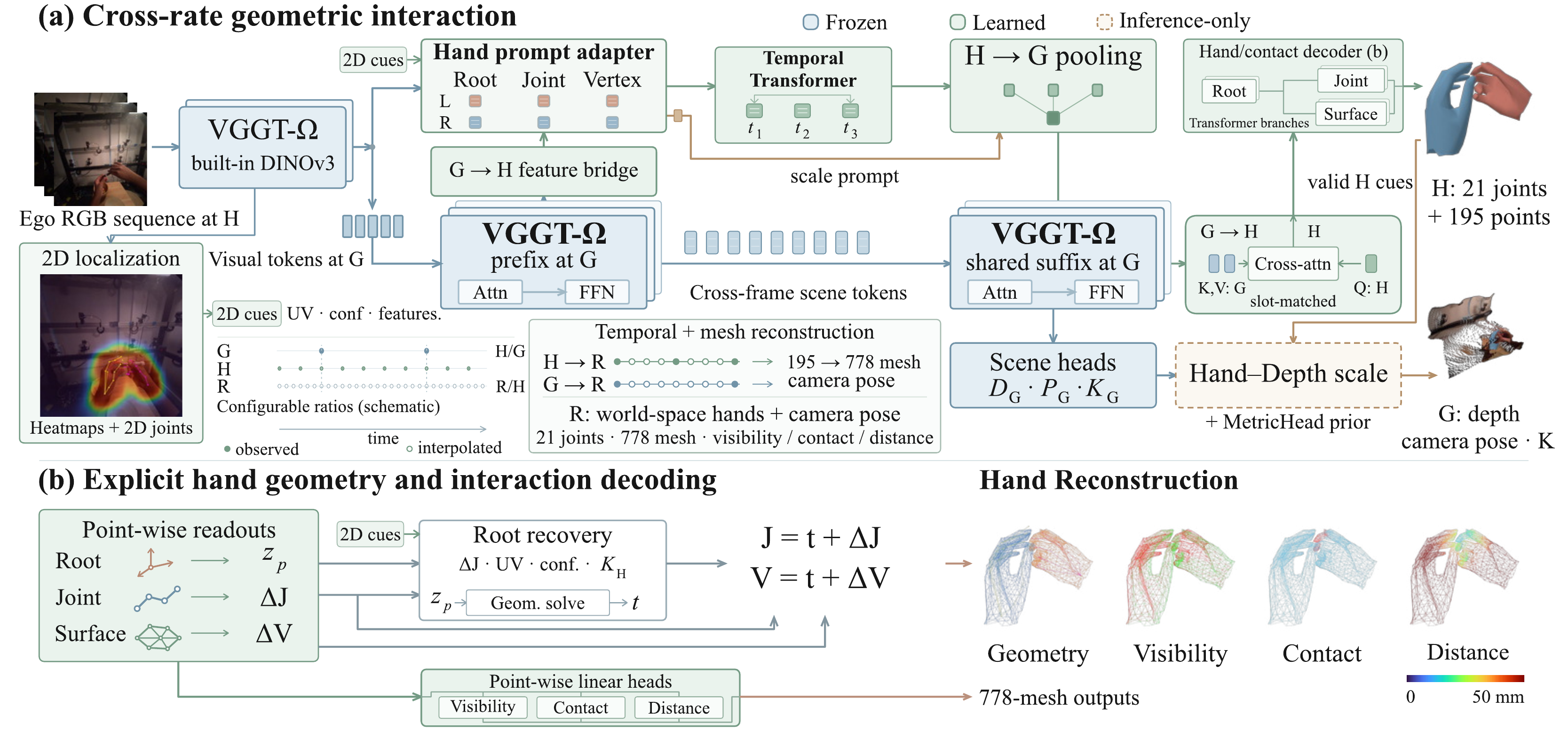}
  \caption{\textbf{Detailed technical framework.} Full token layout, geometric interaction, explicit decoding, and auxiliary branches.}
  \label{fig:detailed_framework}
\end{figure}

\cref{fig:detailed_framework} expands the design-level description of \cref{sec:method} with the token layout, the localization and decoding branches, and the auxiliary heads. The paragraphs below follow the order of the main text and record the details that it leaves out: what the prompts carry, how the two rates exchange information, and what the decoding heads read. They are the choices that turn the design-level description of the main text into a concrete model. We keep the notation of \cref{sec:method}: $H$ denotes the hand-observation times, $G\subseteq H$ the anchors, $s$ the stride, and $R$ the requested output times.

\paragraph{Hand prompts and localization.}
The frozen VGGT-$\Omega$ backbone~\citep{wang2026vggtomega} is built on a DINOv3 encoder~\citep{simeoni2025dinov3}, which embeds every hand-rate frame, and its aggregator converts the anchor frames into cross-frame scene tokens. An attention bridge copies that context into the hand features, which are organized as root, joint, and surface queries. Two small heads localize the hands inside the model: one predicts joint heatmaps together with in-frame offsets, and one predicts whether each hand is in view. Their outputs are detached from the gradient and reused as queries. A bidirectional temporal Transformer links the two hands across the clip, and a learnable scale prompt carries the metric evidence that the decoding heads consume.

\paragraph{Cross-rate interaction.}
A stride $s\in\{1,\ldots,5\}$ selects the anchor times $G\subseteq H$. Each anchor gathers the valid hand prompts of its stride block and pools them with learned attention weights that also observe the position of a prompt inside the block, and the pooled summary is added to the anchor through the gated residual update of \cref{eq:anchor_pool}. Because the residual is zero-initialized and single-element blocks return the anchor unchanged, the dense configuration is recovered exactly. The anchors re-enter the frozen aggregator, where only the adapters are trained, and a slot-matched cross-attention step with the hand prompts as queries and the aggregated global tokens as keys and values returns the context to the hand rate, as shown in \cref{fig:detailed_framework}(a); the same layout carries information back from the hands to the anchors, so H$\to$G and G$\to$H exchanges share one interface. A single phase embedding is shared across strides and read at continuous, cyclically wrapped positions, which is why the ratio changes aggregation density rather than parameter count.

\paragraph{Explicit decoding and attribute heads.}
As shown in \cref{fig:detailed_framework}(b), root, joint, and surface queries read the same point features. Joint and surface queries regress offsets relative to the root, $\Delta J_j$ and $\Delta V_v$ in \cref{eq:absolute_hand}, the surface branch inherits features from the joints it is attached to and mixes over the incident faces of the MANO mesh~\citep{MANO:SIGGRAPHASIA:2017}, and the root query additionally predicts a depth prior $z_p$. The root translation is then recovered by a geometry solve rather than regressed directly: a wrist-anchored, confidence-weighted robust fit reconciles the observed 2D cues, the predicted relative joints, the depth prior, and the hand intrinsics $K_H$. Three point-wise linear heads decode visibility, contact, and distance from the same features, with a sigmoid for the two classification targets and a non-negative decode for distance, and contact merges object contact and opposite-hand contact without identifying the partner.

\paragraph{Scene heads and the metric head.}
The scene heads predict anchor depth $D_G$, camera poses $P_G$, and intrinsics $K_G$. The metric head is a clip-level decoder: it reads scene summaries, detached hand features, in-view probabilities, and the scale prompt, and predicts the scene scale together with a raw-to-meter prior $\lambda_0=\hat\rho/\rho_{\mathrm{raw}}$, where $\rho_{\mathrm{raw}}$ is computed from the valid raw scene points and $\hat\rho$ denotes the predicted scale. Because the hand and scene heads read different grids, $K_H$ is estimated as a clip-level consensus over per-frame estimates with interpolated local residuals and is detached when the root is solved, whereas scene back-projection keeps the native anchor intrinsics $K_G$. At inference, visible hand vertices are projected into the raw depth maps to check the hand--depth agreement, and the resulting factor rescales scene depth and camera translation only, leaving rotations, intrinsics, and the camera-space hands untouched. Because world-space hand positions are composed with the camera trajectory, they follow the rescaled scene rather than being predicted again.

\paragraph{Requested output times.}
The network runs at the hand rate; a parameter-free stage produces the requested times $R$ by interpolating cameras and hands separately and by densifying the native 195 vertices to the 778-vertex MANO mesh with the fixed level-1 upsampling of Mesh Graphormer~\citep{lin2021mesh}, which reproduces the measured points exactly. Overlapping clips are stitched by rescaling them to a common scale from the raw depths at shared frames. Both operations read future frames and therefore stay outside the feed-forward network, so the deployment remains offline.

\section{Training Objectives and Supervision}
\label{app:training}
We summarize the representative loss terms used to train the unified model.
The training pipeline draws from the seven egocentric datasets listed in the main text and combines their available masked supervision over 2D observations, hand and scene geometry, cameras, scale, point attributes, and temporal motion. Contact and distance targets follow from point-to-triangle distances between the annotated hand points and the annotated object or opposite-hand surfaces. For each term, only finite targets covered by its supervision mask contribute to the loss; valid task losses are aggregated into the multi-task objective. We write $p$ for a hand query point (a joint or a hand-surface point) and $\mathcal{Q}$ for the valid points of a term, i.e.\ the points whose target is finite and whose supervision mask is valid. A hat denotes a predicted quantity, the confidence weight $\alpha$ is used by the geometry and contact terms, and each term below carries its own balancing weight.

\paragraph{Training recipe.}
During training, we sample the supported temporal strides in fixed proportions.
\begin{table}[!htbp]
\centering
\captionsetup{skip=4pt,justification=raggedright,singlelinecheck=false}
\caption{\textbf{Training configuration.} Only the hand prompts, adapters, and prediction heads are optimized; the VGGT-$\Omega$ backbone stays frozen.}
\label{tab:training_config}
\small
\setlength{\tabcolsep}{4pt}
\renewcommand{\arraystretch}{1.0}
\begin{tabularx}{\linewidth}{@{}>{\raggedright\arraybackslash}p{0.27\linewidth}X@{}}
\toprule
Setting & Value \\
\midrule
Optimizer & AdamW ($\beta=(0.9,0.999)$, $\epsilon=10^{-8}$), weight decay $0.01$ \\
Learning rate & $2.5\times10^{-4}$ \\
Schedule & linear warmup over $100$ steps, then cosine decay to $10^{-5}$ \\
Batch size & $8$ per device on $8\times$ NVIDIA H20 (effective $64$) \\
Resolution & $256\times256$ \\
Training steps & $3{,}200$ \\
Gradient clipping & $1.0$ \\
Precision & bfloat16 with DeepSpeed ZeRO-2 \\
\bottomrule
\end{tabularx}
\end{table}

\paragraph{Explicit hand geometry.}
For a query point $p\in\mathcal{Q}$, the model predicts a root-relative offset $\hat{\delta}_p$ and a log-confidence $s_p$, and $\delta_p$ is the annotated offset of the same point. Offsets are normalized by the average valid hand-point scale $a$ of the instance, with $\hat{a}$ its predicted counterpart, and $\alpha$ weights the confidence term, giving the confidence-weighted robust regression loss
\begin{equation}
\mathcal{L}_{\mathrm{geom}}
=\frac{1}{|\mathcal{Q}|}
\sum_{p\in\mathcal{Q}}
\left[
\exp(s_p)\,
\left\|
\frac{\hat{\delta}_p}{\hat{a}}-
\frac{\delta_p}{a}
\right\|_1
-\alpha s_p
\right].
\end{equation}
The same form is applied to joints and the native surface points, while the root is recovered from the metric scene and image observations.

\paragraph{Visibility and contact.}
Visibility is supervised with a masked binary cross-entropy, where $\hat{v}_p$ is the predicted visibility logit and $v_p\in\{0,1\}$ the annotated visibility of point $p$:
\begin{equation}
\mathcal{L}_{\mathrm{vis}}
=\frac{1}{|\mathcal{Q}|}
\sum_{p\in\mathcal{Q}}
\operatorname{BCEWithLogits}(\hat{v}_p,v_p).
\end{equation}
For contact, the classification error is coupled with a learned point confidence:
\begin{equation}
\mathcal{L}_{\mathrm{con}}
=\frac{1}{|\mathcal{Q}|}
\sum_{p\in\mathcal{Q}}
\left[
\exp(c_p)\,
\operatorname{BCEWithLogits}(\hat{z}_p,y_p)
-\alpha c_p
\right],
\end{equation}
where $\hat{z}_p$ is the predicted contact logit, $y_p\in\{0,1\}$ the annotated contact of the same point, $c_p$ the predicted log-confidence, and $\alpha$ the confidence weight introduced above. These losses are evaluated at both joints and surface points when the corresponding labels are available.

\paragraph{Distance.}
The distance head predicts a log-distance $\hat{\ell}$. To preserve resolution near contact while limiting the influence of distant points, we regress
\begin{equation}
\mathcal{L}_{\mathrm{dist}}
=\frac{1}{|\mathcal{Q}|}
\sum_{p\in\mathcal{Q}}
\operatorname{SmoothL1}\!\left(
\hat{\ell}_p,\log(1+1000d_p)
\right),
\end{equation}
where $d_p$ is the annotated point-to-surface distance of point $p$ in meters and the factor $1000$ expresses the target in millimeters.

\paragraph{2D cues and scene supervision.}
The full-image localization branch uses a weighted heatmap BCE $\mathcal{L}_{\mathrm{heatmap}}$ and a Smooth-$L_1$ offset loss $\mathcal{L}_{\mathrm{offset}}$; the predicted 3D points $\hat{X}$ are additionally constrained by the observed 2D coordinates $u$ through the projection $\Pi$, and the three terms are weighted by $\lambda_h$, $\lambda_o$, and $\lambda_r$:
\begin{equation}
\mathcal{L}_{\mathrm{2D}}
=\lambda_h\mathcal{L}_{\mathrm{heatmap}}
\,+\,\lambda_o\mathcal{L}_{\mathrm{offset}}
\,+\,\lambda_r
\left\|\Pi(\hat{X})-u\right\|_1 .
\end{equation}
The scene and camera branches provide masked depth, camera-intrinsics, and camera-motion supervision: the depth term uses a confidence-weighted robust error, camera translation is compared after scene-scale normalization, and camera rotation uses a geodesic rotation loss. We write $\mathcal{L}_{\mathrm{3R}}$ for this scene-and-camera objective; the geometry term keeps unit weight, so the full training loss is
\begin{equation}
\mathcal{L}
=\mathcal{L}_{\mathrm{geom}}
\,+\,\lambda_{\mathrm{vis}}\mathcal{L}_{\mathrm{vis}}
\,+\,\lambda_{\mathrm{con}}\mathcal{L}_{\mathrm{con}}
\,+\,\lambda_{\mathrm{dist}}\mathcal{L}_{\mathrm{dist}}
\,+\,\lambda_{\mathrm{2D}}\mathcal{L}_{\mathrm{2D}}
\,+\,\lambda_{\mathrm{3R}}\mathcal{L}_{\mathrm{3R}} .
\end{equation}

\section{Inference Efficiency and Temporal Rates}
\label{app:efficiency}
\begin{table}[!htbp]
\centering
\small
\setlength{\tabcolsep}{3pt}
\caption{\textbf{Inference efficiency on one NVIDIA H20.} The $G/H/R$ column lists the anchor, hand, and output rates; H FPS and R FPS are in frames per second, VRAM is in GiB, and parameter counts use B/M; timings follow the released configuration of each system.}
\label{tab:inference_efficiency}
\begin{tabularx}{\linewidth}{@{}>{\raggedright\arraybackslash}X c r r r r@{}}
\toprule
Method & $G/H/R$ & H FPS$\uparrow$ & R FPS$\uparrow$ & VRAM (GiB)$\downarrow$ & Total params. (B/M) \\
\midrule
\multicolumn{6}{@{}l}{\textit{EgoFound3R configurations}} \\
\midrule
EgoFound3R-A ($512^2$) & 4/21/60 & 48.63 & 138.95 & 3.87 & 1.356B \\
EgoFound3R-B ($512^2$) & 34/168/500 & 43.65 & 129.91 & 9.61 & 1.356B \\
EgoFound3R-C ($512^2$) & 100/500/1498 & 28.69 & 85.95 & 22.57 & 1.356B \\
\midrule
\multicolumn{6}{@{}l}{\textit{Contact methods: GT hand + object mesh + object 6DoF}} \\
\midrule
ContactOpt & -- & 1396.03 & -- & 2.34 & 1.42M \\
S$^2$Contact & -- & 274.06 & -- & 3.69 & 0.59M \\
\midrule
\multicolumn{6}{@{}l}{\textit{Multimodal contact method}} \\
\midrule
InteractVLM & -- & 0.41 & -- & 32.92 & 13.05B \\
\midrule
\multicolumn{6}{@{}l}{\textit{Camera-space hand methods}} \\
\midrule
PAD-Hand & -- & 3.71 & -- & 22.36 & 1.36B \\
WiLoR & -- & 4.81 & -- & 22.36 & 0.67B \\
EgoForce & -- & 10.63 & -- & 1.63 & 0.79B \\
\midrule
\multicolumn{6}{@{}l}{\textit{World-space hand methods}} \\
\midrule
Dyn-HaMR & -- & 1.20 & -- & 6.03 & 0.70B \\
HaWoR & -- & 2.93 & -- & 7.76 & 1.17B \\
ReViV & -- & 14.77 & -- & 15.63 & 1.54B \\
\midrule
\multicolumn{6}{@{}l}{\textit{Scene methods}} \\
\midrule
DA3-Large-1.1 & -- & 50.15 & -- & 5.02 & 0.41B \\
VGGT & -- & 23.46 & -- & 8.81 & 1.26B \\
VGGT-$\Omega$ & -- & 21.05 & -- & 8.30 & 1.14B \\
$\pi^3$ & -- & 16.90 & -- & 7.36 & 0.96B \\
LingBot-Map & -- & 4.27 & -- & 12.54 & 1.16B \\
\bottomrule
\end{tabularx}
\end{table}

\begin{figure}[H]
  \centering
  \includegraphics[width=\linewidth]{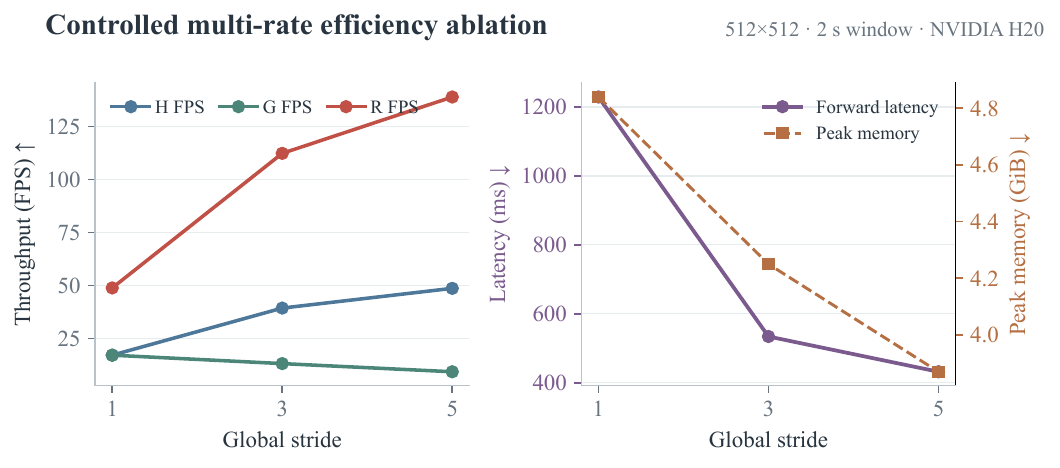}
  \caption{\textbf{Hand-rate efficiency across global strides.} Only the global stride $s\in\{1,3,5\}$ changes, with parameters, $512\times512$ input, and $2$\,s windows held fixed, so the panels isolate its effect on hand-rate throughput (H FPS), the anchor and output rates, forward latency, and peak memory.}
  \label{fig:stride_ablation_appendix}
\end{figure}

\cref{tab:inference_efficiency} lists the configurations underlying the main-text efficiency plot; its $G/H/R$ column reports the anchor, hand, and output rates of each configuration. \cref{fig:stride_ablation_appendix} ablates the global stride at fixed parameters, resolution, and window length, reporting hand-rate throughput together with latency and peak memory; reconstruction accuracy across strides is reported in Appendix~\ref{app:ablation}.
Timings use one NVIDIA H20 with 500-frame measurements and follow the original measurement reports; PAD-Hand includes its WiLoR prerequisite. ContactOpt and S$^2$Contact consume ground-truth hand geometry, object meshes, and object 6DoF poses, so they are reported separately from the throughput comparison.

\section{Evaluation Metrics and Ranking Rules}
\label{app:metrics}
This section expands the metric summary given in the experimental setup: we state how each geometry variant aligns prediction and ground truth before measuring the error, and how contact and visibility scores are computed.
Let $\hat X$ and $X$ denote predicted and ground-truth points of a window, respectively, and let $N$ be the number of valid points.
Geometry metrics are reported in millimeters for joints (mean per joint position error (MPJPE)) and vertices (mean per vertex position error (MPVPE)), and each variant fixes the alignment applied before the error is measured.

\paragraph{Evaluation protocol.}
The common evaluation uses 2{,}378 randomly drawn windows, each a 2\,s (60-frame) clip whose frames are kept at the native frame rate of its dataset. Windows are sampled at a fixed ratio per dataset from six datasets (H2O, HOT3D Aria, ARCTIC, OakInk-v2, TACO, and HOI4D) and are kept disjoint from the training split; clips are non-overlapping within a sequence, and the sampling ratio bounds evaluation cost.
To bound evaluation cost, Dyn-HaMR and InteractVLM are each evaluated on a separate, randomly sampled subset of 100 windows and paired with an \method{} control computed on exactly the same windows; these pairs are ranked only within the pair, as reported in Appendix~\ref{app:matched_windows}.

\begin{table}[!htbp]
\centering
\captionsetup{skip=4pt,justification=raggedright,singlelinecheck=false}
\caption{\textbf{Dataset splits.} Train counts source sequences in the training partition; eval counts the $2$\,s windows sampled from the test partition under the common protocol. STERA-10M contributes training data only.}
\label{tab:dataset_splits}
\small
\setlength{\tabcolsep}{4pt}
\renewcommand{\arraystretch}{1.0}
\begin{tabular}{@{}l@{\hspace{1.5em}}rr@{}}
\toprule
Dataset & Train sequences & Eval clips \\
\midrule
OakInk-v2 & 565 & 400 \\
TACO & 839 & 400 \\
HOI4D & 1{,}431 & 461 \\
H2O & 138 & 283 \\
HOT3D Aria & 136 & 400 \\
ARCTIC & 226 & 434 \\
STERA-10M & all & -- \\
\bottomrule
\end{tabular}
\end{table}

\paragraph{Raw and root-relative error.}
Raw is the unaligned error, computed after expressing both prediction and ground truth in the coordinate system of the window without any similarity alignment.
RR (root-relative) subtracts the root translation, here the wrist, from both prediction and ground truth before evaluation, so that only the relative articulation of joints or vertices is measured.

\paragraph{Aligned error.}
PA (Procrustes-aligned) computes one similarity transform per frame by Procrustes analysis and reports the residual after alignment, which removes global rotation, translation, and scale from each frame independently.
$\mathrm{Sim}(3)$ instead estimates a single similarity transform per window and applies it to all frames of that window, which preserves within-window motion while removing window-level scale and rotation.

\paragraph{World-coordinate error.}
W (world-coordinate) evaluates hand points in the world frame, after the predictions are mapped there with the camera poses: a method that predicts its own trajectory uses its predicted cameras, whereas a camera-space method without one falls back to the ground-truth extrinsics and is marked $\dagger$ in the tables.
Both metrics then fit one Sim(3) from the predicted to the ground-truth hand points: W on the first two valid frames, applying that transform to the whole window, and WA on all valid hand points of the window, which removes global drift of the reconstructed hand trajectory while retaining local motion.
For W and WA, methods marked with $\dagger$ in the tables use ground-truth extrinsics and are ranked separately from methods that use predicted extrinsics; $\ddagger$ marks our model when it receives ground-truth camera intrinsics while still predicting extrinsics; superscripts identify the 100-window subsets, and local metrics are ranked jointly.

\paragraph{Contact and visibility.}
Contact is evaluated per hand point, that is, per joint or per vertex, before any aggregation over points.
Precision is $\mathrm{TP}/(\mathrm{TP}+\mathrm{FP})$, recall is $\mathrm{TP}/(\mathrm{TP}+\mathrm{FN})$, and F1 is their harmonic mean $2PR/(P+R)$; all three are unitless scores in $[0,1]$ and are computed with the same labels for joints, native vertices, and interpolated vertices. Scores are computed per window and then averaged over windows, so the reported F1 is the mean of the per-window F1 values rather than the harmonic mean of the reported precision and recall. The threshold sweep of \cref{tab:contact_threshold_sweep} is the one exception: its AP and best-F1 pool the points of a dataset, while its F1@0.5 column repeats the window-averaged value.
Visibility uses the same three definitions on the binary per-point visibility label.
Distance is the point-to-surface distance from a hand point to the annotated object or opposite-hand surface, and the tables report its mean absolute error (MAE, mm) over all valid query points (MAE(All)), over points with predicted contact (MAE(Pred)), and over points with ground-truth contact (MAE(GT)).
The best and second-best values at the displayed precision are marked in bold and with an underline, shaded light blue and amber, and ties share the same marking. A dash denotes an undefined metric.
Rows evaluated on the 100-window subsets are ranked only within their matched groups, as described in Appendix~\ref{app:matched_windows}. The tables use these symbols as column headers: Raw, RR, PA, $\mathrm{Sim}(3)$, W, and WA for joints and vertices, and P, R, and F1 for contact and visibility.

\section{Additional Hand Geometry Evaluation}
\label{app:surface}
\cref{tab:vertex} reports hand geometry at 21 joints (MPJPE) together with native 195-vertex and interpolated 778-vertex surfaces (MPVPE) on OakInk-v2, TACO, and HOI4D, complementing the main-text joint comparison. Across resolutions the ranking follows the joint comparison: \method{} leads the unaligned, root-relative, and world metrics.
The 195 points are the fixed level-1 MANO sampling released with Mesh Graphormer~\citep{lin2021mesh}, a subset of the 778 MANO vertices selected by a one-hot matrix; our 778-vertex results are recovered from them with the matching upsampling matrix from the same release.
\begin{table}[!htbp]
\centering
\captionsetup{skip=4pt,justification=raggedright,singlelinecheck=false}
\caption{\textbf{Hand geometry} on OakInk-v2, TACO, and HOI4D in mm at 21 joints and 195/778 vertices; marks follow Table~\ref{tab:joint}.}
\label{tab:vertex}
\scriptsize
\renewcommand{\arraystretch}{1.0}
\setlength{\aboverulesep}{1.5pt}
\setlength{\belowrulesep}{1.5pt}
\setlength{\tabcolsep}{1.0pt}
\resizebox{\linewidth}{!}{%
\begin{tabular}{l*{18}{c}}
\toprule
& \multicolumn{6}{c}{21 joints: MPJPE} & \multicolumn{6}{c}{195 vertices: MPVPE} & \multicolumn{6}{c}{778 vertices: MPVPE} \\
\cmidrule(lr){2-7}\cmidrule(lr){8-13}\cmidrule(lr){14-19}
Method & \multicolumn{1}{c}{Raw$\downarrow$} & \multicolumn{1}{c}{RR$\downarrow$} & \multicolumn{1}{c}{PA$\downarrow$} & \multicolumn{1}{c}{$\mathrm{Sim}(3)\!\downarrow$} & \multicolumn{1}{c}{W$\downarrow$} & \multicolumn{1}{c}{WA$\downarrow$} & \multicolumn{1}{c}{Raw$\downarrow$} & \multicolumn{1}{c}{RR$\downarrow$} & \multicolumn{1}{c}{PA$\downarrow$} & \multicolumn{1}{c}{$\mathrm{Sim}(3)\!\downarrow$} & \multicolumn{1}{c}{W$\downarrow$} & \multicolumn{1}{c}{WA$\downarrow$} & \multicolumn{1}{c}{Raw$\downarrow$} & \multicolumn{1}{c}{RR$\downarrow$} & \multicolumn{1}{c}{PA$\downarrow$} & \multicolumn{1}{c}{$\mathrm{Sim}(3)\!\downarrow$} & \multicolumn{1}{c}{W$\downarrow$} & \multicolumn{1}{c}{WA$\downarrow$} \\
\midrule
\multicolumn{19}{l}{\textit{OakInk-v2}}\\[1.5pt]
WiLoR$^{\dagger}$ & 91.24 & 27.66 & \bestv{7.78} & \bestv{16.89} & \secondv{42.14} & \bestv{17.04} & 91.16 & 26.00 & \bestv{7.55} & \bestv{16.90} & \secondv{41.84} & \bestv{17.06} & 91.46 & 26.78 & \bestv{7.44} & \bestv{16.82} & \secondv{42.02} & \bestv{16.97} \\
PAD-Hand$^{\dagger}$ & 107.16 & 32.23 & 9.63 & 21.37 & 49.01 & 21.62 & 106.47 & 30.03 & 9.32 & 21.15 & 47.23 & 21.40 & 106.98 & 30.79 & 9.12 & 21.02 & 47.41 & 21.26 \\
EgoForce$^{\dagger}$ & 123.57 & 47.75 & 11.09 & 32.62 & 80.59 & 33.10 & 123.63 & 45.07 & 10.95 & 33.19 & 81.00 & 33.65 & 123.95 & 45.82 & 10.64 & 32.93 & 81.19 & 33.40 \\
\rowcolor{black!4}
\method{}$^{\dagger}$ & -- & -- & -- & -- & \bestv{27.49} & \secondv{17.89} & -- & -- & -- & -- & \bestv{27.79} & \secondv{18.07} & -- & -- & -- & -- & \bestv{28.53} & \secondv{18.76} \\
\cmidrule(lr){6-7}\cmidrule(lr){12-13}\cmidrule(lr){18-19}
Dyn-HaMR$^{100\mathrm{w}}$ & 96.39 & 17.91 & 7.84 & 13.76 & 36.16 & 17.19 & 95.78 & 16.87 & 7.67 & 13.73 & 35.84 & 17.20 & 95.94 & 17.22 & 7.56 & 13.62 & 35.90 & 17.06 \\
HaWoR & 174.73 & 45.85 & 9.49 & 32.19 & 100.46 & 37.47 & 172.35 & 42.56 & 9.54 & 32.32 & 100.62 & 37.78 & 173.68 & 44.08 & 9.24 & 32.19 & 100.83 & 37.59 \\
\rowcolor{black!4}
\method{} & \secondv{51.81} & \bestv{17.77} & \secondv{8.02} & \secondv{17.87} & \bestv{30.88} & \bestv{18.77} & \secondv{51.40} & \bestv{17.11} & \secondv{7.96} & \secondv{18.05} & \bestv{31.32} & \bestv{18.98} & \secondv{51.64} & \bestv{18.37} & \secondv{8.98} & \secondv{18.75} & \bestv{31.99} & \bestv{19.63} \\
\rowcolor{black!4}
\method{}$^{\ddagger}$ & \bestv{39.45} & \secondv{17.84} & 8.04 & 18.01 & \secondv{31.56} & \secondv{18.94} & \bestv{39.28} & \secondv{17.18} & 7.98 & 18.21 & \secondv{32.03} & \secondv{19.17} & \bestv{39.53} & \secondv{18.44} & 9.01 & 18.88 & \secondv{32.58} & \secondv{19.80} \\
\midrule
\multicolumn{19}{l}{\textit{TACO}}\\[1.5pt]
WiLoR$^{\dagger}$ & 70.40 & 28.33 & \secondv{9.49} & 25.95 & \secondv{51.14} & \secondv{26.14} & 69.91 & 26.94 & \secondv{9.87} & 26.11 & \secondv{50.71} & \secondv{26.29} & 70.13 & 27.30 & \bestv{9.72} & 26.02 & \secondv{50.52} & \secondv{26.20} \\
PAD-Hand$^{\dagger}$ & 72.97 & 30.30 & 11.19 & 29.21 & 57.06 & 29.60 & 72.45 & 29.03 & 10.84 & 28.74 & 54.03 & 29.15 & 72.73 & 29.32 & 10.64 & 28.64 & 54.50 & 29.06 \\
EgoForce$^{\dagger}$ & 126.63 & 44.53 & 11.93 & 47.84 & 105.52 & 46.65 & 126.58 & 41.99 & 12.18 & 48.41 & 106.20 & 47.28 & 126.83 & 42.56 & 11.89 & 48.06 & 106.11 & 46.92 \\
\rowcolor{black!4}
\method{}$^{\dagger}$ & -- & -- & -- & -- & \bestv{36.68} & \bestv{24.06} & -- & -- & -- & -- & \bestv{37.47} & \bestv{24.45} & -- & -- & -- & -- & \bestv{38.18} & \bestv{25.37} \\
\cmidrule(lr){6-7}\cmidrule(lr){12-13}\cmidrule(lr){18-19}
Dyn-HaMR$^{100\mathrm{w}}$ & 127.12 & 27.47 & 11.62 & 29.27 & 77.88 & 31.46 & 127.79 & 26.04 & 11.55 & 29.49 & 78.23 & 31.82 & 127.94 & 26.33 & 11.46 & 29.45 & 78.09 & 31.73 \\
HaWoR & 74.28 & 24.19 & 10.55 & 27.71 & 86.38 & 35.34 & 73.75 & \secondv{23.02} & 10.99 & 27.91 & 86.78 & 35.81 & 74.05 & 23.26 & 10.71 & 27.74 & 86.63 & 35.53 \\
\rowcolor{black!4}
\method{} & \secondv{54.64} & \secondv{18.54} & \bestv{8.13} & \bestv{24.15} & \bestv{40.03} & \bestv{25.16} & \secondv{53.76} & \bestv{18.36} & \bestv{8.67} & \bestv{24.49} & \bestv{40.69} & \bestv{25.55} & \secondv{54.03} & \bestv{19.91} & \secondv{10.19} & \bestv{25.39} & \bestv{41.36} & \bestv{26.42} \\
\rowcolor{black!4}
\method{}$^{\ddagger}$ & \bestv{50.44} & \bestv{18.52} & \bestv{8.13} & \secondv{24.63} & \secondv{41.78} & \secondv{25.75} & \bestv{50.61} & \bestv{18.36} & \bestv{8.67} & \secondv{25.03} & \secondv{42.68} & \secondv{26.20} & \bestv{51.54} & \secondv{19.99} & 10.31 & \secondv{25.98} & \secondv{43.22} & \secondv{27.11} \\
\midrule
\multicolumn{19}{l}{\textit{HOI4D}}\\[1.5pt]
WiLoR$^{\dagger}$ & 81.51 & 27.75 & 9.62 & 24.14 & \bestv{49.41} & \secondv{24.57} & 81.34 & 26.40 & 9.61 & 24.25 & \bestv{48.78} & \secondv{24.64} & 82.08 & 26.69 & \bestv{9.51} & \bestv{24.14} & \bestv{48.80} & \bestv{24.56} \\
PAD-Hand$^{\dagger}$ & 75.35 & 29.77 & 10.12 & 24.79 & \secondv{51.92} & 25.04 & 74.91 & 28.68 & 9.83 & 24.65 & \secondv{50.73} & 24.93 & 75.80 & 28.92 & \secondv{9.71} & \secondv{24.54} & \secondv{50.68} & 24.83 \\
EgoForce$^{\dagger}$ & 147.77 & 52.78 & 12.92 & 49.22 & 142.58 & 55.76 & 147.13 & 50.29 & 12.93 & 50.26 & 143.30 & 56.49 & 147.64 & 50.75 & 12.56 & 49.81 & 142.96 & 56.10 \\
\rowcolor{black!4}
\method{}$^{\dagger}$ & -- & -- & -- & -- & 63.01 & \bestv{23.30} & -- & -- & -- & -- & 63.02 & \bestv{23.56} & -- & -- & -- & -- & 63.62 & \secondv{24.72} \\
\cmidrule(lr){6-7}\cmidrule(lr){12-13}\cmidrule(lr){18-19}
Dyn-HaMR$^{100\mathrm{w}}$ & 245.06 & 26.49 & 11.89 & 30.47 & 173.62 & 65.71 & 245.09 & 25.53 & 12.28 & 30.99 & 173.69 & 66.73 & 245.58 & 25.77 & 12.01 & 30.74 & 173.51 & 66.03 \\
HaWoR & 53.46 & 28.47 & 10.49 & 27.85 & 171.09 & 55.96 & 53.11 & 26.53 & 10.38 & 27.73 & 170.80 & 56.69 & 53.30 & 27.33 & 10.19 & 27.67 & 170.58 & 56.31 \\
\rowcolor{black!4}
\method{} & \secondv{47.28} & \bestv{16.63} & \bestv{7.99} & \bestv{23.29} & \secondv{159.70} & \secondv{49.84} & \secondv{48.46} & \bestv{16.20} & \bestv{8.29} & \bestv{23.60} & \secondv{160.03} & \secondv{50.68} & \secondv{49.72} & \bestv{18.06} & 10.16 & 24.75 & \secondv{160.98} & \secondv{50.85} \\
\rowcolor{black!4}
\method{}$^{\ddagger}$ & \bestv{43.43} & \secondv{16.67} & \secondv{8.00} & \secondv{23.39} & \bestv{159.18} & \bestv{49.77} & \bestv{44.20} & \secondv{16.24} & \secondv{8.31} & \secondv{23.78} & \bestv{159.74} & \bestv{50.65} & \bestv{45.50} & \secondv{18.09} & 10.17 & 24.91 & \bestv{160.62} & \bestv{50.80} \\
\bottomrule
\end{tabular}}
\end{table}

\section{Temporal Evaluation}
\label{app:temporal}
Velocity (MPJVE/MPMVE/MPVVE, in mm/s) and acceleration (MPJAE/MPMAE/MPVAE, in m/s$^2$) errors summarize how faithfully the recovered trajectories follow the ground-truth motion at 21 joints, 195 vertices, and the full 778-vertex MANO mesh. \cref{tab:temporal_baselines} compares \method{} with the hand baselines on the three evaluated subsets, and \cref{tab:temporal_matched} repeats the comparison on the matched 100-window subsets shared with Dyn-HaMR. ReViV predicts joints only, so its 195- and 778-vertex entries are unavailable, and our temporal values come from the default-inference rerun of the stride-5 configuration.
\begin{table}[!htbp]
\centering
\captionsetup{skip=4pt,justification=raggedright,singlelinecheck=false}
\caption{\textbf{Temporal consistency} on OakInk-v2, TACO, and HOI4D. Velocity (MPJVE/MPMVE/MPVVE) and acceleration (MPJAE/MPMAE/MPVAE) errors at 21 joints, 195 vertices, and 778 vertices are in mm/s and m/s$^2$; marks follow Table~\ref{tab:joint}.}
\label{tab:temporal_baselines}
\small
\renewcommand{\arraystretch}{1.0}
\setlength{\tabcolsep}{3pt}
\setlength{\aboverulesep}{1.5pt}
\setlength{\belowrulesep}{1.5pt}
\begin{tabularx}{\linewidth}{l*{6}{>{\centering\arraybackslash}X}}
\toprule
& \multicolumn{2}{c}{21 joints} & \multicolumn{2}{c}{195 vertices} & \multicolumn{2}{c}{778 vertices} \\
\cmidrule(lr){2-3}\cmidrule(lr){4-5}\cmidrule(lr){6-7}
Method & MPJVE$\downarrow$ & MPJAE$\downarrow$ & MPMVE$\downarrow$ & MPMAE$\downarrow$ & MPVVE$\downarrow$ & MPVAE$\downarrow$ \\
\midrule
\multicolumn{7}{l}{\textit{OakInk-v2}}\\[1.5pt]
WiLoR$^{\dagger}$ & 116.81 & 4.66 & 110.07 & 4.38 & 112.47 & 4.49 \\
PAD-Hand$^{\dagger}$ & 124.78 & 4.29 & 118.80 & 4.08 & 120.99 & 4.16 \\
EgoForce$^{\dagger}$ & 258.44 & 12.11 & 242.73 & 11.35 & 247.57 & 11.58 \\
HaWoR & \secondv{59.59} & \secondv{2.07} & \secondv{56.20} & \secondv{1.96} & \secondv{57.23} & \secondv{1.99} \\
ReViV & 165.72 & 6.43 & -- & -- & -- & -- \\
\rowcolor{black!4}\method{} & \bestv{56.55} & \bestv{1.17} & \bestv{54.09} & \bestv{1.12} & \bestv{54.29} & \bestv{1.12} \\
\midrule
\multicolumn{7}{l}{\textit{TACO}}\\[1.5pt]
WiLoR$^{\dagger}$ & 138.28 & 5.44 & 130.00 & 5.11 & 132.79 & 5.23 \\
PAD-Hand$^{\dagger}$ & 136.35 & 4.56 & 129.16 & 4.33 & 131.18 & 4.41 \\
EgoForce$^{\dagger}$ & 453.68 & 21.99 & 424.57 & 20.56 & 433.80 & 21.02 \\
HaWoR & \secondv{93.13} & \secondv{2.67} & \secondv{87.76} & \secondv{2.52} & \secondv{89.10} & \secondv{2.56} \\
ReViV & 125.87 & 3.02 & -- & -- & -- & -- \\
\rowcolor{black!4}\method{} & \bestv{92.00} & \bestv{1.87} & \bestv{87.10} & \bestv{1.76} & \bestv{87.53} & \bestv{1.77} \\
\midrule
\multicolumn{7}{l}{\textit{HOI4D}}\\[1.5pt]
WiLoR$^{\dagger}$ & 158.49 & 5.99 & 148.44 & 5.61 & 151.50 & 5.73 \\
PAD-Hand$^{\dagger}$ & \secondv{145.21} & 4.97 & \secondv{137.02} & 4.70 & \secondv{139.47} & 4.79 \\
EgoForce$^{\dagger}$ & 544.59 & 26.48 & 510.65 & 24.80 & 520.77 & 25.31 \\
HaWoR & \bestv{140.11} & \secondv{4.73} & \bestv{131.19} & \secondv{4.42} & \bestv{133.78} & \secondv{4.52} \\
ReViV & 448.65 & 18.25 & -- & -- & -- & -- \\
\rowcolor{black!4}\method{} & 152.69 & \bestv{4.34} & 144.58 & \bestv{4.08} & 147.84 & \bestv{4.20} \\
\bottomrule
\end{tabularx}
\end{table}

\begin{table}[!htbp]
\centering
\captionsetup{skip=4pt,justification=raggedright,singlelinecheck=false}
\caption{\textbf{Matched 100-window temporal consistency.} Dyn-HaMR and \method{} are evaluated on the same windows; errors are in mm/s and m/s$^2$ and marks follow Table~\ref{tab:joint}.}
\label{tab:temporal_matched}
\small
\renewcommand{\arraystretch}{1.0}
\setlength{\tabcolsep}{3pt}
\setlength{\aboverulesep}{1.5pt}
\setlength{\belowrulesep}{1.5pt}
\begin{tabularx}{\linewidth}{l*{6}{>{\centering\arraybackslash}X}}
\toprule
& \multicolumn{2}{c}{21 joints} & \multicolumn{2}{c}{195 vertices} & \multicolumn{2}{c}{778 vertices} \\
\cmidrule(lr){2-3}\cmidrule(lr){4-5}\cmidrule(lr){6-7}
Method & MPJVE$\downarrow$ & MPJAE$\downarrow$ & MPMVE$\downarrow$ & MPMAE$\downarrow$ & MPVVE$\downarrow$ & MPVAE$\downarrow$ \\
\midrule
\multicolumn{7}{l}{\textit{OakInk-v2}}\\[1.5pt]
Dyn-HaMR$^{100\mathrm{w}}$ & 132.68 & 6.04 & 130.90 & 5.95 & 131.84 & 6.00 \\
\rowcolor{black!4}\method{}$^{100\mathrm{w},\mathrm{Dyn}}$ & \bestv{63.85} & \bestv{1.32} & \bestv{61.40} & \bestv{1.27} & \bestv{61.69} & \bestv{1.27} \\
\midrule
\multicolumn{7}{l}{\textit{TACO}}\\[1.5pt]
Dyn-HaMR$^{100\mathrm{w}}$ & 339.05 & 13.55 & 334.51 & 13.36 & 336.74 & 13.47 \\
\rowcolor{black!4}\method{}$^{100\mathrm{w},\mathrm{Dyn}}$ & \bestv{99.81} & \bestv{2.00} & \bestv{93.68} & \bestv{1.88} & \bestv{94.24} & \bestv{1.89} \\
\midrule
\multicolumn{7}{l}{\textit{HOI4D}}\\[1.5pt]
Dyn-HaMR$^{100\mathrm{w}}$ & 439.65 & 18.90 & 436.66 & 18.79 & 438.56 & 18.87 \\
\rowcolor{black!4}\method{}$^{100\mathrm{w},\mathrm{Dyn}}$ & \bestv{204.38} & \bestv{6.78} & \bestv{191.63} & \bestv{6.29} & \bestv{196.89} & \bestv{6.48} \\
\bottomrule
\end{tabularx}
\end{table}

\section{Contact and Visibility Evaluation}
\label{app:contact}
\paragraph{Contact at three point sets.}
\cref{tab:app_contact_surface} reports contact classification at 21-joint, 195-vertex, and 778-vertex resolutions. InteractVLM uses its original 100-window subsets; the corresponding matched EgoFound3R controls are reported separately in Appendix~\ref{app:matched_windows}. Our ground truth marks a point as in contact when its distance to the object or the opposite-hand surface is below 18\,mm at joints and 14\,mm at the 195 vertices of the hand surface, the fingertips included. The three compared systems define contact differently, so none of them shares this criterion: ContactOpt and S$^2$Contact treat a point as covered at a 2\,mm surface tolerance, which is a narrower band than ours, while InteractVLM inherits annotated vertex labels. The two geometric baselines are additionally given ground-truth hand and object geometry.
\begin{table}[!htbp]
\centering
\captionsetup{skip=4pt,justification=raggedright,singlelinecheck=false}
\caption{\textbf{Contact prediction} on OakInk-v2, TACO, and HOI4D at 21 joints, 195 vertices, and 778 vertices. Superscripts mark 100-window subsets, and marks follow Table~\ref{tab:joint}.}
\label{tab:app_contact_surface}
\small
\renewcommand{\arraystretch}{1.0}
\setlength{\aboverulesep}{1.5pt}
\setlength{\belowrulesep}{1.5pt}
\setlength{\tabcolsep}{1.6pt}
\begin{tabularx}{\linewidth}{l*{9}{>{\centering\arraybackslash}X}}
\toprule
& \multicolumn{3}{c}{21 joints} & \multicolumn{3}{c}{195 vertices} & \multicolumn{3}{c}{778 vertices} \\
\cmidrule(lr){2-4}\cmidrule(lr){5-7}\cmidrule(lr){8-10}
Method & \multicolumn{1}{c}{P$\uparrow$} & \multicolumn{1}{c}{R$\uparrow$} & \multicolumn{1}{c}{F1$\uparrow$} & \multicolumn{1}{c}{P$\uparrow$} & \multicolumn{1}{c}{R$\uparrow$} & \multicolumn{1}{c}{F1$\uparrow$} & \multicolumn{1}{c}{P$\uparrow$} & \multicolumn{1}{c}{R$\uparrow$} & \multicolumn{1}{c}{F1$\uparrow$} \\
\midrule
\multicolumn{10}{l}{\textit{OakInk-v2}}\\[1.5pt]
InteractVLM$^{100\mathrm{w}}$ & 0.180 & 0.403 & 0.202 & 0.173 & 0.408 & 0.205 & 0.168 & 0.412 & 0.201 \\
S$^2$Contact & 0.327 & 0.081 & 0.079 & \secondv{0.401} & 0.180 & 0.158 & \secondv{0.397} & 0.198 & 0.173 \\
ContactOpt & \secondv{0.334} & \secondv{0.196} & \secondv{0.154} & 0.334 & \secondv{0.234} & \secondv{0.178} & 0.338 & \secondv{0.248} & \secondv{0.185} \\
\rowcolor{black!4}
\rowcolor{black!4}
\method{} & \bestv{0.510} & \bestv{0.665} & \bestv{0.493} & \bestv{0.525} & \bestv{0.611} & \bestv{0.479} & \bestv{0.516} & \bestv{0.329} & \bestv{0.320} \\
\midrule
\multicolumn{10}{l}{\textit{TACO}}\\[1.5pt]
InteractVLM$^{100\mathrm{w}}$ & 0.388 & 0.705 & 0.472 & 0.348 & 0.743 & 0.451 & 0.356 & 0.750 & 0.458 \\
S$^2$Contact & 0.519 & 0.104 & 0.146 & \secondv{0.590} & 0.210 & 0.285 & \bestv{0.592} & 0.234 & \secondv{0.308} \\
ContactOpt & \secondv{0.557} & \secondv{0.197} & \secondv{0.259} & 0.571 & \secondv{0.249} & \secondv{0.311} & \secondv{0.573} & \secondv{0.241} & 0.303 \\
\rowcolor{black!4}
\rowcolor{black!4}
\method{} & \bestv{0.654} & \bestv{0.777} & \bestv{0.697} & \bestv{0.636} & \bestv{0.719} & \bestv{0.661} & 0.516 & \bestv{0.265} & \bestv{0.320} \\
\midrule
\multicolumn{10}{l}{\textit{HOI4D}}\\[1.5pt]
InteractVLM$^{100\mathrm{w}}$ & 0.313 & 0.537 & 0.281 & 0.256 & 0.556 & 0.261 & 0.249 & 0.553 & 0.261 \\
S$^2$Contact & 0.444 & 0.265 & 0.284 & 0.437 & 0.340 & 0.333 & 0.426 & 0.344 & 0.330 \\
ContactOpt & \secondv{0.518} & \secondv{0.477} & \secondv{0.435} & \secondv{0.506} & \secondv{0.548} & \secondv{0.463} & \secondv{0.496} & \bestv{0.538} & \bestv{0.449} \\
\rowcolor{black!4}
\rowcolor{black!4}
\method{} & \bestv{0.612} & \bestv{0.835} & \bestv{0.675} & \bestv{0.608} & \bestv{0.807} & \bestv{0.656} & \bestv{0.531} & \secondv{0.466} & \secondv{0.445} \\
\bottomrule
\end{tabularx}
\end{table}

\paragraph{Vertex-level visibility.}
\cref{tab:visibility_vertices} reports visibility at the native 195 vertices and the interpolated 778-vertex mesh, which the compared pipelines do not provide; HVD predicts joint visibility only, so no ranking is shown.
\begin{table}[!htbp]
\centering
\captionsetup{skip=4pt,justification=raggedright,singlelinecheck=false}
\caption{\textbf{Vertex-level visibility} at the native 195 vertices and the interpolated 778-vertex mesh.}
\label{tab:visibility_vertices}
\small
\renewcommand{\arraystretch}{1.0}
\setlength{\aboverulesep}{1.5pt}
\setlength{\belowrulesep}{1.5pt}
\setlength{\tabcolsep}{2pt}
\begin{tabularx}{\linewidth}{ll*{6}{>{\centering\arraybackslash}X}}
\toprule
Dataset & Method & \multicolumn{3}{c}{195 vertices} & \multicolumn{3}{c}{778 vertices} \\
\cmidrule(lr){3-5}\cmidrule(lr){6-8}
& & \multicolumn{1}{c}{P$\uparrow$} & \multicolumn{1}{c}{R$\uparrow$} & \multicolumn{1}{c}{F1$\uparrow$} & \multicolumn{1}{c}{P$\uparrow$} & \multicolumn{1}{c}{R$\uparrow$} & \multicolumn{1}{c}{F1$\uparrow$} \\
\midrule
OakInk-v2 & \method{} & 0.556 & 0.705 & 0.595 & 0.555 & 0.690 & 0.588 \\
TACO & \method{} & 0.437 & 0.818 & 0.567 & 0.439 & 0.805 & 0.565 \\
HOI4D & \method{} & 0.572 & 0.789 & 0.660 & 0.581 & 0.781 & 0.663 \\
\bottomrule
\end{tabularx}
\end{table}

\paragraph{Threshold and definition sensitivity.}
\cref{tab:app_contact_surface} reports one operating point under one ground-truth definition. Because the compared systems were trained under different contact conventions, we isolate the two effects that can be separated without retraining: the decision threshold applied to each method's scores, and the distance tolerance that defines the ground truth. No model or prediction is modified, and every number below is post-processing of the same cached per-point scores and distances.
\begin{table}[!htbp]
\centering
\captionsetup{skip=4pt,justification=raggedright,singlelinecheck=false}
\caption{\textbf{Decision-threshold sweep for contact prediction} on the 195 vertices. AP and best-F1 pool the points of a dataset over 99 thresholds, with the best-F1 threshold in parentheses; marks follow Table~\ref{tab:joint}.}
\label{tab:contact_threshold_sweep}
\small
\setlength{\tabcolsep}{4pt}
\renewcommand{\arraystretch}{1.0}
\begin{tabularx}{\linewidth}{@{}l l *{3}{>{\centering\arraybackslash}X}}
\toprule
Dataset & Method & AP$\uparrow$ & best-F1$\uparrow$ & F1@0.5$\uparrow$ \\
\midrule
\textit{OakInk-v2} & ContactOpt & 0.308 & 0.411 (0.21) & \secondv{0.178} \\
 & S$^2$Contact & \secondv{0.395} & \secondv{0.449} (0.09) & 0.158 \\
\rowcolor{black!4}\cellcolor{white} & \method{} & \bestv{0.603} & \bestv{0.617} (0.39) & \bestv{0.479} \\
\midrule
\textit{TACO} & ContactOpt & 0.475 & 0.542 (0.15) & \secondv{0.311} \\
 & S$^2$Contact & \secondv{0.512} & \secondv{0.558} (0.09) & 0.285 \\
\rowcolor{black!4}\cellcolor{white} & \method{} & \bestv{0.714} & \bestv{0.691} (0.43) & \bestv{0.661} \\
\midrule
\textit{HOI4D} & ContactOpt & \secondv{0.482} & \secondv{0.589} (0.35) & \secondv{0.463} \\
 & S$^2$Contact & 0.399 & 0.515 (0.26) & 0.333 \\
\rowcolor{black!4}\cellcolor{white} & \method{} & \bestv{0.735} & \bestv{0.714} (0.51) & \bestv{0.656} \\
\midrule
\multicolumn{5}{@{}l}{\textit{100-window paired subsets}} \\
\textit{OakInk-v2} & InteractVLM$^{100\mathrm{w}}$ & 0.184 & 0.275 (0.71) & 0.205 \\
\rowcolor{black!4}\cellcolor{white} & \method{}$^{100\mathrm{w}}$ & \bestv{0.559} & \bestv{0.594} (0.42) & \bestv{0.442} \\
\midrule
\textit{TACO} & InteractVLM$^{100\mathrm{w}}$ & 0.397 & 0.494 (0.90) & 0.451 \\
\rowcolor{black!4}\cellcolor{white} & \method{}$^{100\mathrm{w}}$ & \bestv{0.725} & \bestv{0.712} (0.42) & \bestv{0.691} \\
\midrule
\textit{HOI4D} & InteractVLM$^{100\mathrm{w}}$ & 0.249 & 0.371 (0.82) & 0.261 \\
\rowcolor{black!4}\cellcolor{white} & \method{}$^{100\mathrm{w}}$ & \bestv{0.700} & \bestv{0.693} (0.48) & \bestv{0.670} \\
\bottomrule
\end{tabularx}
\end{table}

\paragraph{Sweep and reading.}
\cref{tab:contact_threshold_sweep} sweeps each method's scores from 0.01 to 0.99. AP summarizes ranking quality independently of the operating point, and best-F1 reports the best attainable point together with the threshold that attains it. Under this sweep the proposed model attains the highest AP and best-F1 on the three evaluation datasets, and its best-F1 thresholds (0.39--0.51) lie close to the conventional 0.5 decision point, whereas the two geometric baselines peak at thresholds as low as 0.09.
\begin{table}[!htbp]
\centering
\captionsetup{skip=4pt,justification=raggedright,singlelinecheck=false}
\caption{\textbf{Narrow-band contact comparison} on the 195 vertices at the 0.5 operating point. The two column groups differ only in the labeling tolerance; the 2\,mm band is defined on the 195 vertices only.}
\label{tab:contact_narrow_band}
\small
\setlength{\tabcolsep}{4pt}
\renewcommand{\arraystretch}{1.0}
\begin{tabularx}{\linewidth}{@{}l l *{3}{>{\centering\arraybackslash}X}@{\hspace{5pt}}*{3}{>{\centering\arraybackslash}X}}
\toprule
& & \multicolumn{3}{c}{our GT} & \multicolumn{3}{c}{2\,mm band} \\
\cmidrule(lr){3-5}\cmidrule(lr){6-8}
Dataset & Method & P$\uparrow$ & R$\uparrow$ & F1$\uparrow$ & P$\uparrow$ & R$\uparrow$ & F1$\uparrow$ \\
\midrule
\textit{OakInk-v2} & ContactOpt & 0.334 & \secondv{0.234} & \secondv{0.178} & 0.080 & \secondv{0.335} & 0.096 \\
 & S$^2$Contact & \secondv{0.401} & 0.180 & 0.158 & \bestv{0.117} & 0.331 & \secondv{0.115} \\
\rowcolor{black!4}\cellcolor{white} & \method{} & \bestv{0.525} & \bestv{0.611} & \bestv{0.479} & \secondv{0.103} & \bestv{0.741} & \bestv{0.164} \\
\midrule
\textit{TACO} & ContactOpt & 0.571 & \secondv{0.249} & \secondv{0.311} & 0.101 & 0.358 & 0.140 \\
 & S$^2$Contact & \secondv{0.590} & 0.210 & 0.285 & \bestv{0.127} & \secondv{0.376} & \secondv{0.171} \\
\rowcolor{black!4}\cellcolor{white} & \method{} & \bestv{0.636} & \bestv{0.719} & \bestv{0.661} & \secondv{0.108} & \bestv{0.836} & \bestv{0.188} \\
\midrule
\textit{HOI4D} & ContactOpt & \secondv{0.506} & \secondv{0.548} & \secondv{0.463} & \secondv{0.100} & \secondv{0.657} & \secondv{0.157} \\
 & S$^2$Contact & 0.437 & 0.340 & 0.333 & 0.086 & 0.390 & 0.125 \\
\rowcolor{black!4}\cellcolor{white} & \method{} & \bestv{0.608} & \bestv{0.807} & \bestv{0.656} & \bestv{0.101} & \bestv{0.865} & \bestv{0.174} \\
\midrule
\multicolumn{8}{@{}l}{\textit{100-window paired subsets}} \\
\textit{OakInk-v2} & InteractVLM$^{100\mathrm{w}}$ & 0.173 & 0.408 & 0.205 & 0.031 & 0.426 & 0.055 \\
\rowcolor{black!4}\cellcolor{white} & \method{}$^{100\mathrm{w}}$ & \bestv{0.441} & \bestv{0.638} & \bestv{0.442} & \bestv{0.090} & \bestv{0.774} & \bestv{0.151} \\
\midrule
\textit{TACO} & InteractVLM$^{100\mathrm{w}}$ & 0.348 & 0.743 & 0.451 & 0.052 & 0.758 & 0.094 \\
\rowcolor{black!4}\cellcolor{white} & \method{}$^{100\mathrm{w}}$ & \bestv{0.654} & \bestv{0.755} & \bestv{0.691} & \bestv{0.111} & \bestv{0.877} & \bestv{0.196} \\
\midrule
\textit{HOI4D} & InteractVLM$^{100\mathrm{w}}$ & 0.256 & 0.556 & 0.261 & 0.035 & 0.589 & 0.055 \\
\rowcolor{black!4}\cellcolor{white} & \method{}$^{100\mathrm{w}}$ & \bestv{0.631} & \bestv{0.794} & \bestv{0.670} & \bestv{0.095} & \bestv{0.874} & \bestv{0.168} \\
\bottomrule
\end{tabularx}
\end{table}

\begin{figure*}[t]
  \centering
  \includegraphics[width=\linewidth]{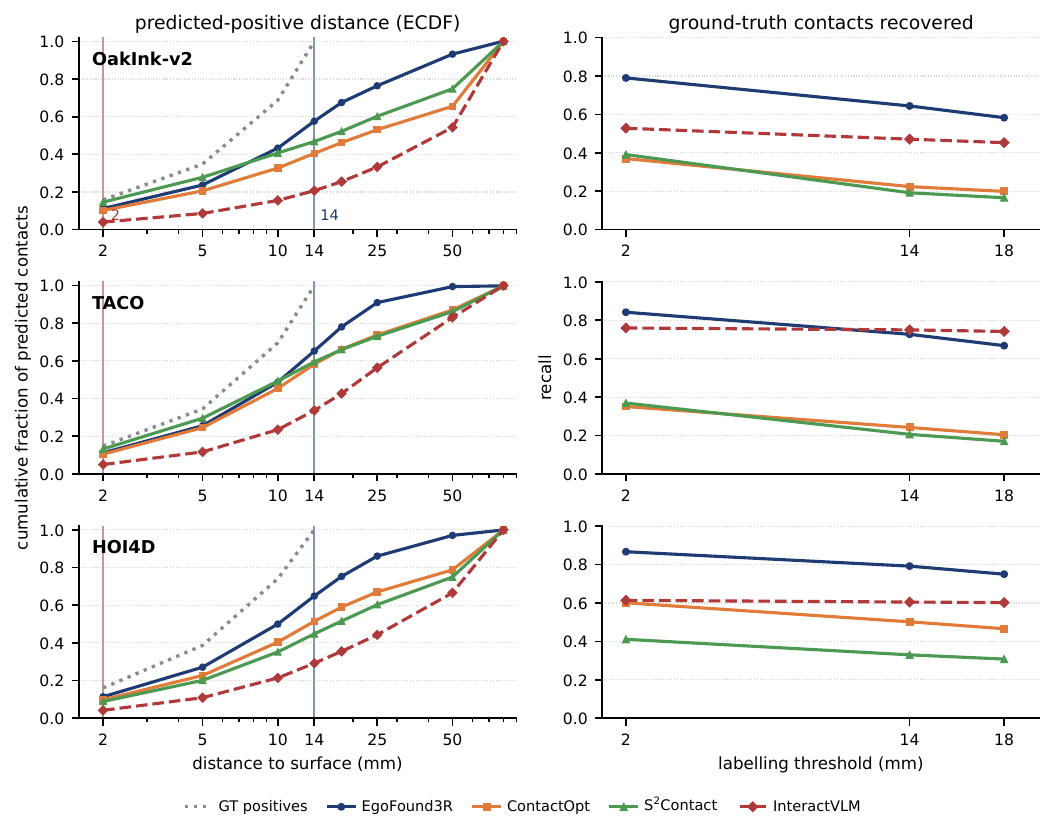}
  \caption{\textbf{Where each method's predicted contacts lie.} Predicted-positive distances on the 195 vertices, one row per dataset. Left: cumulative fraction of a method's positives within a given distance to the surface, ground truth in gray, with vertical lines marking the 2\,mm coverage band of the two geometric baselines and our 14\,mm contact band. Right: recall of ground-truth contacts as the labelling threshold varies, with reference lines at 2, 14, and 18\,mm.}
  \label{fig:contact_distance_hist}
\end{figure*}

\paragraph{Agreement on the predicted positives.}
\cref{tab:contact_narrow_band} narrows the ground truth to a 2\,mm surface band, which matches the coverage criterion of the two geometric baselines. That band is computed on the 195 vertices. The ordering is preserved on the three evaluation datasets. As shown in \cref{fig:contact_distance_hist}, the positives predicted by the two geometric baselines are not confined to the narrow band that defines their coverage: their distances spread over tens of millimeters, with a median of 10--21\,mm across the three evaluation datasets and a quarter to nearly half of the points beyond 25\,mm. The proposed model keeps a tighter distribution, with a median of 10--12\,mm and 57--65\% of its positives inside the 14\,mm contact band, which is consistent with the distance regression used to supervise its attribute head.

\section{Matched-Subset Controls for 100-Window Baselines}
\label{app:matched_windows}
\cref{tab:matched_dyn} and \cref{tab:matched_interact} use the original Dyn-HaMR hand subsets and InteractVLM contact subsets, respectively. Each baseline and its EgoFound3R controls are evaluated on the same windows. Results are ranked within each matched group, separately from the common-window evaluation.
\clearpage
\begin{table}[!t]
\centering
\scriptsize
\captionsetup{skip=4pt,justification=raggedright,singlelinecheck=false}
\caption{\textbf{Matched 100-window hand geometry.} Dyn-HaMR and \method{} are evaluated on the same windows; errors are in mm.}
\label{tab:matched_dyn}
\setlength{\tabcolsep}{0.5pt}
\renewcommand{\arraystretch}{0.94}
\resizebox{\linewidth}{!}{%
\begin{tabular}{l*{18}{c}}
\toprule
& \multicolumn{6}{c}{21 joints: MPJPE} & \multicolumn{6}{c}{195 vertices: MPVPE} & \multicolumn{6}{c}{778 vertices: MPVPE} \\
\cmidrule(lr){2-7}\cmidrule(lr){8-13}\cmidrule(lr){14-19}
Method & \multicolumn{1}{c}{Raw$\downarrow$} & \multicolumn{1}{c}{RR$\downarrow$} & \multicolumn{1}{c}{PA$\downarrow$} & \multicolumn{1}{c}{$\mathrm{Sim}(3)\!\downarrow$} & \multicolumn{1}{c}{W$\downarrow$} & \multicolumn{1}{c}{WA$\downarrow$} & \multicolumn{1}{c}{Raw$\downarrow$} & \multicolumn{1}{c}{RR$\downarrow$} & \multicolumn{1}{c}{PA$\downarrow$} & \multicolumn{1}{c}{$\mathrm{Sim}(3)\!\downarrow$} & \multicolumn{1}{c}{W$\downarrow$} & \multicolumn{1}{c}{WA$\downarrow$} & \multicolumn{1}{c}{Raw$\downarrow$} & \multicolumn{1}{c}{RR$\downarrow$} & \multicolumn{1}{c}{PA$\downarrow$} & \multicolumn{1}{c}{$\mathrm{Sim}(3)\!\downarrow$} & \multicolumn{1}{c}{W$\downarrow$} & \multicolumn{1}{c}{WA$\downarrow$} \\
\midrule
\multicolumn{19}{l}{\textit{OakInk-v2}}\\[1.5pt]
Dyn-HaMR$^{100\mathrm{w}}$ & 96.39 & \bestv{17.91} & \bestv{7.84} & \bestv{13.76} & 36.16 & \bestv{17.19} & 95.78 & \bestv{16.87} & \bestv{7.67} & \bestv{13.73} & 35.84 & \bestv{17.20} & 95.94 & \bestv{17.22} & \bestv{7.56} & \bestv{13.62} & 35.90 & \bestv{17.06} \\
\rowcolor{black!4}
\method{}$^{100\mathrm{w},\mathrm{Dyn}}$ & \secondv{50.88} & \secondv{21.17} & \secondv{8.50} & \secondv{20.30} & \bestv{30.20} & \secondv{20.88} & \secondv{49.07} & \secondv{20.97} & \secondv{8.54} & \secondv{20.49} & \bestv{30.21} & \secondv{21.02} & \secondv{49.17} & \secondv{22.25} & \secondv{9.53} & \secondv{21.11} & \bestv{30.99} & \secondv{21.63} \\
\rowcolor{black!4}
\method{}$^{\ddagger,100\mathrm{w},\mathrm{Dyn}}$ & \bestv{43.29} & 21.18 & 8.51 & 21.15 & \secondv{32.68} & 21.88 & \bestv{42.73} & \secondv{20.97} & \secondv{8.54} & 21.33 & \secondv{32.71} & 22.06 & \bestv{42.94} & 22.28 & 9.55 & 21.93 & \secondv{33.40} & 22.63 \\
\midrule
\multicolumn{19}{l}{\textit{TACO}}\\[1.5pt]
Dyn-HaMR$^{100\mathrm{w}}$ & 127.12 & \secondv{27.47} & \secondv{11.62} & 29.27 & 77.88 & 31.46 & 127.79 & \secondv{26.04} & 11.55 & 29.49 & 78.23 & 31.82 & 127.94 & 26.33 & 11.46 & 29.45 & 78.09 & 31.73 \\
\rowcolor{black!4}
\method{}$^{100\mathrm{w},\mathrm{Dyn}}$ & \bestv{56.15} & \bestv{20.80} & \bestv{8.74} & \bestv{24.53} & \bestv{40.70} & \bestv{25.49} & \bestv{55.18} & \bestv{20.37} & \secondv{9.02} & \bestv{24.70} & \bestv{41.43} & \bestv{25.82} & \bestv{55.59} & \bestv{21.96} & \bestv{10.47} & \bestv{25.58} & \bestv{42.09} & \bestv{26.62} \\
\rowcolor{black!4}
\method{}$^{\ddagger,100\mathrm{w},\mathrm{Dyn}}$ & \secondv{59.47} & \bestv{20.80} & \bestv{8.74} & \secondv{24.90} & \secondv{42.10} & \secondv{25.93} & \secondv{59.76} & \bestv{20.37} & \bestv{9.01} & \secondv{25.13} & \secondv{43.20} & \secondv{26.34} & \secondv{60.91} & \secondv{22.04} & \secondv{10.58} & \secondv{26.06} & \secondv{43.64} & \secondv{27.19} \\
\midrule
\multicolumn{19}{l}{\textit{HOI4D}}\\[1.5pt]
Dyn-HaMR$^{100\mathrm{w}}$ & 245.06 & 26.49 & \secondv{11.89} & 30.47 & \bestv{173.62} & 65.71 & 245.09 & 25.53 & \secondv{12.28} & 30.99 & \bestv{173.69} & \secondv{66.73} & 245.58 & 25.77 & 12.01 & 30.74 & \bestv{173.51} & \bestv{66.03} \\
\rowcolor{black!4}
\method{}$^{100\mathrm{w},\mathrm{Dyn}}$ & \secondv{50.92} & \bestv{17.73} & \bestv{8.71} & \secondv{26.25} & \secondv{205.48} & \secondv{65.61} & \secondv{52.06} & \secondv{16.98} & \bestv{8.65} & \secondv{26.27} & \secondv{208.31} & 66.75 & \secondv{53.15} & \secondv{18.81} & \secondv{10.49} & \secondv{27.38} & \secondv{209.55} & 66.45 \\
\rowcolor{black!4}
\method{}$^{\ddagger,100\mathrm{w},\mathrm{Dyn}}$ & \bestv{40.88} & \secondv{17.74} & \bestv{8.71} & \bestv{26.16} & 206.30 & \bestv{65.53} & \bestv{41.73} & \bestv{16.97} & \bestv{8.65} & \bestv{26.23} & 209.41 & \bestv{66.65} & \bestv{42.88} & \bestv{18.79} & \bestv{10.48} & \bestv{27.37} & 210.35 & \secondv{66.37} \\
\bottomrule
\end{tabular}}
\end{table}

\begin{table}[!t]
\centering
\small
\captionsetup{skip=4pt,justification=raggedright,singlelinecheck=false}
\caption{\textbf{Matched 100-window contact comparison.} InteractVLM and \method{} are evaluated on exactly the same windows, and the best value in each column is marked in bold.}
\label{tab:matched_interact}
\setlength{\tabcolsep}{2pt}
\renewcommand{\arraystretch}{1.0}
\begin{tabularx}{\linewidth}{l*{9}{>{\centering\arraybackslash}X}}
\toprule
Method & \multicolumn{3}{c}{21 joints} & \multicolumn{3}{c}{195 vertices} & \multicolumn{3}{c}{778 vertices} \\
\cmidrule(lr){2-4}\cmidrule(lr){5-7}\cmidrule(lr){8-10}
 & \multicolumn{1}{c}{P$\uparrow$} & \multicolumn{1}{c}{R$\uparrow$} & \multicolumn{1}{c}{F1$\uparrow$} & \multicolumn{1}{c}{P$\uparrow$} & \multicolumn{1}{c}{R$\uparrow$} & \multicolumn{1}{c}{F1$\uparrow$} & \multicolumn{1}{c}{P$\uparrow$} & \multicolumn{1}{c}{R$\uparrow$} & \multicolumn{1}{c}{F1$\uparrow$} \\
\midrule
\multicolumn{10}{l}{\textit{OakInk-v2}}\\[1.5pt]
InteractVLM$^{100\mathrm{w}}$ & 0.180 & 0.403 & 0.202 & 0.173 & 0.408 & 0.205 & 0.168 & \bestv{0.412} & 0.201 \\
\rowcolor{black!4}
\method{}$^{100\mathrm{w},\mathrm{InteractVLM}}$ & \bestv{0.381} & \bestv{0.762} & \bestv{0.431} & \bestv{0.441} & \bestv{0.638} & \bestv{0.442} & \bestv{0.457} & 0.295 & \bestv{0.298} \\
\multicolumn{10}{l}{\textit{TACO}}\\[1.5pt]
InteractVLM$^{100\mathrm{w}}$ & 0.388 & 0.705 & 0.472 & 0.348 & 0.743 & 0.451 & 0.356 & \bestv{0.750} & \bestv{0.458} \\
\rowcolor{black!4}
\method{}$^{100\mathrm{w},\mathrm{InteractVLM}}$ & \bestv{0.670} & \bestv{0.821} & \bestv{0.730} & \bestv{0.654} & \bestv{0.755} & \bestv{0.691} & \bestv{0.582} & 0.320 & 0.388 \\
\multicolumn{10}{l}{\textit{HOI4D}}\\[1.5pt]
InteractVLM$^{100\mathrm{w}}$ & 0.313 & 0.537 & 0.281 & 0.256 & 0.556 & 0.261 & 0.249 & \bestv{0.553} & 0.261 \\
\rowcolor{black!4}
\method{}$^{100\mathrm{w},\mathrm{InteractVLM}}$ & \bestv{0.653} & \bestv{0.796} & \bestv{0.691} & \bestv{0.631} & \bestv{0.794} & \bestv{0.670} & \bestv{0.458} & 0.410 & \bestv{0.418} \\
\bottomrule
\end{tabularx}
\end{table}

\clearpage

\section{Architecture and Multi-Rate Ablations}
\label{app:ablation}
We report local and world-space hand geometry, contact, visibility, and distance on OakInk-v2, TACO, and HOI4D: \cref{tab:ablation_geometry} for hand geometry, \cref{tab:ablation_contact} for contact and visibility, and \cref{tab:ablation_distance} for distance.
\paragraph{0.05B and 0.1B (reduced capacity).}
The hand module is reduced to about 0.05B and 0.1B parameters while the rest of the design is left unchanged, so the two variants show how much of the accuracy depends on the capacity of the hand decoder.

\paragraph{MANO (parameter regression).}
Instead of predicting point coordinates, the decoder regresses MANO rotation parameters and obtains the mesh through the MANO model, which separates the value of the explicit point representation from that of the underlying mesh topology.

\paragraph{w/o $\Omega$ (no pretrained backbone).}
The frozen VGGT-$\Omega$ backbone is removed together with the features it provides, so the hand and scene heads are trained from the supervision alone; the variant measures what the pretrained geometric priors contribute.

\paragraph{1-way (one-way exchange).}
The cross-rate interface is restricted to one direction: backbone features still reach the hands, but the hand prompts no longer update the anchors, which tests whether hand evidence must also shape the scene representation.

\paragraph{S$k$ (temporal stride).}
The full model runs at temporal stride $k$, that is, with one global anchor per $k$ hand frames; S1 is the densest and S5 the default, shaded in the tables, so the variants vary aggregation density at a fixed parameter count.

\begin{table}[!htbp]
\centering
\captionsetup{skip=4pt,justification=raggedright,singlelinecheck=false}
\caption{\textbf{Hand-geometry ablation} on OakInk-v2, TACO, and HOI4D. Errors are in mm; S$k$ is \method{} at stride $k$.}
\label{tab:ablation_geometry}
\scriptsize
\setlength{\tabcolsep}{1.0pt}
\renewcommand{\arraystretch}{0.96}
\resizebox{\linewidth}{!}{%
\begin{tabular}{l*{18}{c}}
\toprule
& \multicolumn{6}{c}{21 joints: MPJPE} & \multicolumn{6}{c}{195 vertices: MPVPE} & \multicolumn{6}{c}{778 vertices: MPVPE} \\
\cmidrule(lr){2-7}\cmidrule(lr){8-13}\cmidrule(lr){14-19}
Variant & Raw$\downarrow$ & RR$\downarrow$ & PA$\downarrow$ & $\mathrm{Sim}(3)\!\downarrow$ & W$\downarrow$ & WA$\downarrow$ & Raw$\downarrow$ & RR$\downarrow$ & PA$\downarrow$ & $\mathrm{Sim}(3)\!\downarrow$ & W$\downarrow$ & WA$\downarrow$ & Raw$\downarrow$ & RR$\downarrow$ & PA$\downarrow$ & $\mathrm{Sim}(3)\!\downarrow$ & W$\downarrow$ & WA$\downarrow$ \\
\midrule
\multicolumn{19}{l}{\textit{OakInk-v2}}\\[1.5pt]
0.05B & 112.47 & 35.90 & 16.76 & 33.35 & 65.09 & 35.51 & 110.83 & 35.50 & 16.15 & 33.45 & 65.83 & 35.66 & 111.13 & 36.33 & 16.20 & 33.46 & 66.14 & 35.63 \\
0.1B & 98.34 & 31.74 & 15.24 & 30.74 & 61.31 & 32.39 & 98.33 & 30.92 & 15.27 & 30.67 & 59.29 & 32.32 & 98.69 & 31.64 & 15.36 & 30.73 & 59.56 & 32.32 \\
MANO & 73.39 & 21.02 & 8.74 & 24.13 & 45.73 & 25.71 & 73.27 & 19.87 & 8.55 & 24.30 & 45.46 & 25.92 & 73.12 & 21.01 & 9.36 & 24.70 & 46.05 & 26.27 \\
w/o $\Omega$ & 86.13 & 22.06 & 9.92 & 23.63 & 44.03 & 24.88 & 85.25 & 20.59 & 9.54 & 23.69 & 43.87 & 24.95 & 86.07 & 21.67 & 10.32 & 24.17 & 44.39 & 25.40 \\
1-way & 79.62 & 22.01 & 9.96 & 22.94 & 41.28 & 24.30 & 80.46 & 21.44 & 9.98 & 23.00 & 40.84 & 24.43 & 80.88 & 22.47 & 10.65 & 23.41 & 41.40 & 24.81 \\
S1 & 53.15 & \bestv{17.68} & \bestv{7.97} & \bestv{17.50} & \bestv{28.13} & \bestv{17.91} & 52.36 & 17.11 & \secondv{7.97} & \bestv{17.71} & \bestv{28.45} & \bestv{18.13} & 52.46 & \bestv{18.33} & \bestv{8.96} & \bestv{18.39} & \bestv{29.17} & \bestv{18.79} \\
S2 & 53.14 & 17.82 & \secondv{8.02} & 18.01 & \secondv{30.37} & 18.81 & 52.72 & \bestv{17.08} & \bestv{7.96} & 18.18 & \secondv{30.75} & 18.99 & 52.96 & \bestv{18.33} & \secondv{8.98} & 18.88 & \secondv{31.41} & 19.65 \\
S3 & 51.97 & 17.78 & \secondv{8.02} & 17.99 & 30.70 & 18.85 & 51.64 & \secondv{17.10} & \bestv{7.96} & 18.18 & 31.09 & 19.05 & 51.90 & \secondv{18.35} & \secondv{8.98} & 18.88 & 31.77 & 19.71 \\
S4 & \bestv{51.72} & 17.78 & \secondv{8.02} & 17.91 & 30.68 & 18.80 & \bestv{51.38} & \secondv{17.10} & \bestv{7.96} & 18.10 & 31.09 & 19.00 & \bestv{51.63} & \secondv{18.35} & \secondv{8.98} & 18.80 & 31.76 & 19.65 \\
\rowcolor{black!4}
S5 & \secondv{51.81} & \secondv{17.77} & \secondv{8.02} & \secondv{17.87} & 30.88 & \secondv{18.77} & \secondv{51.40} & 17.11 & \bestv{7.96} & \secondv{18.05} & 31.32 & \secondv{18.98} & \secondv{51.64} & 18.37 & \secondv{8.98} & \secondv{18.75} & 31.99 & \secondv{19.63} \\
\midrule
\multicolumn{19}{l}{\textit{TACO}}\\[1.5pt]
0.05B & 109.71 & 36.74 & 15.15 & 42.08 & 81.37 & 42.66 & 110.05 & 33.75 & 14.38 & 42.10 & 83.13 & 42.84 & 110.63 & 35.05 & 15.30 & 42.42 & 82.54 & 43.09 \\
0.1B & 96.90 & 30.55 & 13.14 & 35.07 & 61.66 & 35.81 & 93.16 & 29.92 & 13.53 & 34.95 & 60.85 & 35.78 & 93.04 & 30.89 & 14.17 & 35.34 & 61.53 & 36.14 \\
MANO & 86.66 & 24.29 & 10.68 & 34.41 & 57.45 & 35.15 & 86.25 & 23.33 & 10.80 & 34.49 & 57.54 & 35.29 & 86.18 & 24.47 & 11.96 & 35.06 & 57.95 & 35.78 \\
w/o $\Omega$ & 82.59 & 23.49 & 10.11 & 31.38 & 53.38 & 32.33 & 81.91 & 22.06 & 10.50 & 31.76 & 54.42 & 32.76 & 82.17 & 23.33 & 11.71 & 32.38 & 54.53 & 33.32 \\
1-way & 78.82 & 23.70 & 10.07 & 31.21 & 54.16 & 32.14 & 78.33 & 22.25 & 10.31 & 31.20 & 54.16 & 32.19 & 78.69 & 23.61 & 11.48 & 31.86 & 54.46 & 32.77 \\
S1 & 67.42 & 22.55 & 8.51 & 27.90 & 47.10 & 28.54 & 68.98 & 19.25 & 8.75 & 27.94 & 46.22 & 28.59 & 69.70 & 21.01 & 10.50 & 28.79 & 46.54 & 29.44 \\
S2 & \secondv{53.45} & \bestv{18.53} & \bestv{8.12} & 24.25 & \secondv{40.23} & \secondv{25.18} & \secondv{52.76} & \bestv{18.35} & \bestv{8.66} & 24.59 & \secondv{40.88} & 25.59 & \secondv{53.17} & \secondv{19.92} & 10.22 & 25.51 & \secondv{41.56} & 26.48 \\
S3 & \bestv{53.34} & \bestv{18.53} & \bestv{8.12} & 24.23 & 40.39 & \secondv{25.18} & \bestv{52.61} & \secondv{18.36} & \bestv{8.66} & 24.56 & 40.98 & \secondv{25.58} & \bestv{53.01} & 19.93 & \secondv{10.21} & 25.49 & 41.66 & 26.47 \\
S4 & 53.86 & \secondv{18.54} & \bestv{8.12} & \secondv{24.20} & 40.28 & \secondv{25.18} & 53.08 & \secondv{18.36} & \secondv{8.67} & \secondv{24.54} & 40.92 & \secondv{25.58} & 53.42 & 19.93 & \secondv{10.21} & \secondv{25.45} & 41.60 & \secondv{26.46} \\
\rowcolor{black!4}
S5 & 54.64 & \secondv{18.54} & \secondv{8.13} & \bestv{24.15} & \bestv{40.03} & \bestv{25.16} & 53.76 & \secondv{18.36} & \secondv{8.67} & \bestv{24.49} & \bestv{40.69} & \bestv{25.55} & 54.03 & \bestv{19.91} & \bestv{10.19} & \bestv{25.39} & \bestv{41.36} & \bestv{26.42} \\
\midrule
\multicolumn{19}{l}{\textit{HOI4D}}\\[1.5pt]
0.05B & 157.49 & 32.62 & 14.54 & 41.65 & 155.50 & 58.89 & 152.87 & 31.79 & 15.20 & 42.94 & 159.34 & 59.95 & 152.03 & 33.17 & 15.88 & 43.22 & \secondv{158.96} & 59.79 \\
0.1B & 140.38 & 31.89 & 13.76 & 40.07 & 177.85 & 55.65 & 136.61 & 30.75 & 14.88 & 40.69 & 176.65 & 56.59 & 135.81 & 32.01 & 15.47 & 40.99 & 176.79 & 56.39 \\
MANO & 97.37 & 25.36 & 10.71 & 40.45 & 169.67 & 57.57 & 97.08 & 24.25 & 10.74 & 40.99 & 170.05 & 58.42 & 96.93 & 25.72 & 11.97 & 41.49 & 171.27 & 58.39 \\
w/o $\Omega$ & 92.76 & 26.09 & 10.37 & 37.26 & \secondv{153.96} & 56.10 & 92.65 & 23.57 & 10.75 & 38.08 & 158.38 & 56.92 & 93.81 & 25.36 & 12.29 & 38.75 & 159.89 & 57.02 \\
1-way & 99.21 & 23.90 & 10.52 & 39.40 & 158.57 & 56.71 & 98.46 & 23.07 & 10.99 & 40.03 & \secondv{158.08} & 57.51 & 98.47 & 24.80 & 12.39 & 40.60 & 159.69 & 57.60 \\
S1 & 52.05 & 17.04 & \secondv{8.03} & 23.49 & \bestv{151.04} & \bestv{49.58} & 53.91 & 16.43 & \secondv{8.29} & \secondv{23.72} & \bestv{151.48} & \bestv{50.33} & 55.12 & 18.33 & 10.19 & \secondv{24.89} & \bestv{152.16} & \bestv{50.55} \\
S2 & 48.90 & \secondv{16.63} & \bestv{7.99} & 23.49 & 158.35 & 50.17 & 50.14 & \bestv{16.19} & \bestv{8.28} & 23.81 & 158.46 & 50.99 & 51.44 & \bestv{18.06} & \secondv{10.17} & 24.96 & 159.47 & 51.18 \\
S3 & 49.60 & 16.64 & \bestv{7.99} & 23.47 & 158.96 & 49.95 & 50.89 & \secondv{16.20} & \bestv{8.28} & 23.79 & 159.48 & 50.78 & 52.23 & \secondv{18.08} & 10.18 & 24.95 & 160.39 & 50.97 \\
S4 & \secondv{48.40} & \bestv{16.61} & \bestv{7.99} & \secondv{23.44} & 158.51 & 50.16 & \secondv{49.63} & \bestv{16.19} & \secondv{8.29} & 23.75 & 158.72 & 50.99 & \secondv{50.92} & \bestv{18.06} & \secondv{10.17} & 24.91 & 159.71 & 51.17 \\
\rowcolor{black!4}
S5 & \bestv{47.28} & \secondv{16.63} & \bestv{7.99} & \bestv{23.29} & 159.70 & \secondv{49.84} & \bestv{48.46} & \secondv{16.20} & \secondv{8.29} & \bestv{23.60} & 160.03 & \secondv{50.68} & \bestv{49.72} & \bestv{18.06} & \bestv{10.16} & \bestv{24.75} & 160.98 & \secondv{50.85} \\
\bottomrule
\end{tabular}}
\end{table}

\begin{table}[!htbp]
\centering
\captionsetup{skip=4pt,justification=raggedright,singlelinecheck=false}
\caption{\textbf{Contact and visibility ablation} on OakInk-v2, TACO, and HOI4D. P, R, and F1 denote contact precision, recall, and F1, and VP, VR, and VF1 the corresponding visibility scores.}
\label{tab:ablation_contact}
\scriptsize
\setlength{\tabcolsep}{1.0pt}
\renewcommand{\arraystretch}{0.96}
\resizebox{\linewidth}{!}{%
\begin{tabular}{l*{18}{c}}
\toprule
& \multicolumn{6}{c}{21 joints} & \multicolumn{6}{c}{195 vertices} & \multicolumn{6}{c}{778 vertices} \\
\cmidrule(lr){2-7}\cmidrule(lr){8-13}\cmidrule(lr){14-19}
Variant & P$\uparrow$ & R$\uparrow$ & F1$\uparrow$ & VP$\uparrow$ & VR$\uparrow$ & VF1$\uparrow$ & P$\uparrow$ & R$\uparrow$ & F1$\uparrow$ & VP$\uparrow$ & VR$\uparrow$ & VF1$\uparrow$ & P$\uparrow$ & R$\uparrow$ & F1$\uparrow$ & VP$\uparrow$ & VR$\uparrow$ & VF1$\uparrow$ \\
\midrule
\multicolumn{19}{l}{\textit{OakInk-v2}}\\[1.5pt]
0.05B & 0.541 & 0.283 & 0.298 & 0.820 & 0.736 & 0.760 & \secondv{0.600} & 0.251 & 0.277 & \secondv{0.678} & 0.170 & 0.244 & 0.373 & 0.165 & 0.175 & \secondv{0.695} & 0.161 & 0.235 \\
0.1B & \secondv{0.549} & 0.358 & 0.345 & 0.843 & 0.708 & 0.752 & \bestv{0.620} & 0.170 & 0.205 & \bestv{0.720} & 0.192 & 0.272 & 0.325 & 0.194 & 0.185 & \bestv{0.733} & 0.186 & 0.267 \\
MANO & 0.512 & 0.595 & 0.474 & \bestv{0.927} & \bestv{0.807} & \bestv{0.856} & 0.537 & 0.582 & \bestv{0.485} & 0.566 & \bestv{0.775} & \bestv{0.642} & \secondv{0.530} & 0.175 & 0.206 & 0.567 & \bestv{0.760} & \bestv{0.636} \\
w/o $\Omega$ & 0.535 & 0.606 & 0.480 & \secondv{0.925} & 0.796 & 0.844 & 0.552 & 0.557 & 0.463 & 0.554 & 0.770 & 0.630 & 0.276 & 0.250 & 0.203 & 0.553 & \secondv{0.757} & 0.625 \\
1-way & \bestv{0.595} & 0.489 & 0.451 & 0.923 & \secondv{0.802} & \secondv{0.848} & 0.592 & 0.492 & 0.449 & 0.561 & \secondv{0.772} & \secondv{0.637} & 0.311 & 0.254 & 0.212 & 0.561 & \secondv{0.757} & \secondv{0.632} \\
S1 & 0.511 & \bestv{0.673} & \bestv{0.498} & 0.921 & 0.744 & 0.800 & 0.524 & 0.602 & 0.473 & 0.552 & 0.706 & 0.592 & \bestv{0.552} & \bestv{0.331} & \bestv{0.329} & 0.550 & 0.692 & 0.585 \\
S2 & 0.509 & \secondv{0.670} & 0.494 & 0.920 & 0.747 & 0.802 & 0.518 & \bestv{0.613} & 0.475 & 0.557 & 0.703 & 0.594 & 0.528 & 0.319 & 0.313 & 0.555 & 0.689 & 0.587 \\
S3 & 0.513 & 0.667 & \secondv{0.497} & 0.919 & 0.747 & 0.802 & 0.522 & \secondv{0.612} & 0.478 & 0.556 & 0.704 & 0.595 & 0.513 & 0.326 & 0.318 & 0.555 & 0.690 & 0.588 \\
S4 & 0.511 & 0.665 & 0.495 & 0.919 & 0.747 & 0.802 & 0.521 & 0.611 & 0.477 & 0.557 & 0.704 & 0.595 & 0.521 & \secondv{0.329} & \secondv{0.320} & 0.555 & 0.690 & 0.588 \\
\rowcolor{black!4}
S5 & 0.510 & 0.665 & 0.493 & 0.919 & 0.747 & 0.802 & 0.525 & 0.611 & \secondv{0.479} & 0.556 & 0.705 & 0.595 & 0.516 & \secondv{0.329} & \secondv{0.320} & 0.555 & 0.690 & 0.588 \\
\midrule
\multicolumn{19}{l}{\textit{TACO}}\\[1.5pt]
0.05B & 0.610 & 0.523 & 0.542 & 0.852 & 0.761 & 0.800 & 0.609 & 0.447 & 0.493 & \bestv{0.602} & 0.260 & 0.350 & 0.333 & 0.162 & 0.188 & \bestv{0.612} & 0.249 & 0.341 \\
0.1B & 0.590 & 0.611 & 0.579 & 0.871 & 0.750 & 0.801 & 0.600 & 0.476 & 0.506 & \secondv{0.591} & 0.298 & 0.381 & 0.357 & 0.147 & 0.177 & \secondv{0.597} & 0.288 & 0.373 \\
MANO & \secondv{0.668} & 0.665 & 0.643 & \bestv{0.921} & 0.801 & 0.854 & \secondv{0.642} & 0.586 & 0.592 & 0.458 & 0.741 & 0.562 & 0.462 & 0.163 & 0.218 & 0.463 & 0.719 & 0.559 \\
w/o $\Omega$ & 0.665 & 0.695 & 0.661 & 0.913 & 0.813 & 0.856 & 0.640 & 0.608 & 0.599 & 0.447 & 0.758 & 0.557 & 0.347 & 0.163 & 0.196 & 0.451 & 0.741 & 0.554 \\
1-way & \bestv{0.705} & 0.600 & 0.628 & 0.907 & 0.825 & 0.861 & \bestv{0.645} & 0.613 & 0.609 & 0.459 & 0.749 & \secondv{0.564} & 0.370 & 0.202 & 0.236 & 0.462 & 0.732 & \secondv{0.562} \\
S1 & 0.639 & \bestv{0.800} & \bestv{0.698} & 0.912 & \bestv{0.857} & \bestv{0.881} & 0.598 & \bestv{0.782} & \bestv{0.665} & 0.432 & \bestv{0.821} & 0.563 & 0.364 & 0.258 & 0.274 & 0.434 & \bestv{0.808} & \secondv{0.562} \\
S2 & 0.655 & 0.776 & 0.696 & \secondv{0.917} & \secondv{0.850} & 0.879 & 0.636 & \secondv{0.719} & \secondv{0.661} & 0.436 & 0.818 & \bestv{0.567} & \bestv{0.516} & \bestv{0.277} & \bestv{0.330} & 0.439 & 0.805 & \bestv{0.565} \\
S3 & 0.655 & 0.776 & \secondv{0.697} & \secondv{0.917} & \secondv{0.850} & 0.879 & 0.635 & \secondv{0.719} & \secondv{0.661} & 0.436 & \secondv{0.819} & \bestv{0.567} & \secondv{0.514} & \bestv{0.277} & \bestv{0.330} & 0.439 & \secondv{0.806} & \bestv{0.565} \\
S4 & 0.655 & \secondv{0.777} & \secondv{0.697} & \secondv{0.917} & \secondv{0.850} & \secondv{0.880} & 0.636 & \secondv{0.719} & \secondv{0.661} & 0.436 & \secondv{0.819} & \bestv{0.567} & 0.513 & \secondv{0.272} & \secondv{0.326} & 0.439 & 0.805 & \bestv{0.565} \\
\rowcolor{black!4}
S5 & 0.654 & \secondv{0.777} & \secondv{0.697} & \secondv{0.917} & \secondv{0.850} & \secondv{0.880} & 0.636 & \secondv{0.719} & \secondv{0.661} & 0.437 & 0.818 & \bestv{0.567} & \bestv{0.516} & 0.265 & 0.320 & 0.439 & 0.805 & \bestv{0.565} \\
\midrule
\multicolumn{19}{l}{\textit{HOI4D}}\\[1.5pt]
0.05B & 0.612 & 0.443 & 0.448 & 0.825 & 0.762 & 0.783 & 0.612 & 0.503 & 0.475 & 0.606 & 0.332 & 0.411 & \bestv{0.561} & 0.034 & 0.051 & \secondv{0.644} & 0.328 & 0.417 \\
0.1B & 0.546 & 0.636 & 0.526 & 0.833 & 0.744 & 0.777 & \bestv{0.639} & 0.426 & 0.422 & \bestv{0.626} & 0.348 & 0.428 & 0.478 & 0.063 & 0.081 & \bestv{0.666} & 0.345 & 0.438 \\
MANO & 0.600 & 0.676 & 0.593 & 0.892 & 0.812 & 0.844 & \secondv{0.638} & 0.550 & 0.528 & \secondv{0.613} & 0.685 & 0.641 & 0.507 & 0.163 & 0.195 & 0.624 & 0.672 & 0.640 \\
w/o $\Omega$ & 0.606 & 0.699 & 0.597 & \secondv{0.902} & 0.781 & 0.833 & 0.624 & 0.678 & 0.594 & 0.589 & 0.697 & 0.631 & 0.350 & 0.263 & 0.244 & 0.601 & 0.683 & 0.631 \\
1-way & \bestv{0.630} & 0.645 & 0.582 & \bestv{0.914} & 0.787 & 0.841 & 0.605 & 0.696 & 0.592 & 0.605 & 0.693 & 0.641 & 0.443 & 0.148 & 0.169 & 0.620 & 0.680 & 0.643 \\
S1 & 0.606 & \bestv{0.845} & \secondv{0.675} & 0.892 & \bestv{0.885} & \bestv{0.886} & 0.597 & \bestv{0.820} & \secondv{0.654} & 0.570 & \bestv{0.793} & \secondv{0.660} & 0.493 & 0.442 & 0.404 & 0.578 & \bestv{0.784} & \secondv{0.662} \\
S2 & 0.612 & 0.834 & \secondv{0.675} & 0.891 & \secondv{0.884} & \secondv{0.885} & 0.606 & \secondv{0.809} & \bestv{0.656} & 0.573 & \secondv{0.789} & \bestv{0.661} & 0.515 & 0.458 & 0.433 & 0.582 & \secondv{0.781} & \bestv{0.663} \\
S3 & \secondv{0.613} & \secondv{0.835} & \bestv{0.676} & 0.891 & \secondv{0.884} & \secondv{0.885} & 0.606 & \secondv{0.809} & \bestv{0.656} & 0.573 & \secondv{0.789} & \bestv{0.661} & 0.511 & \bestv{0.468} & 0.439 & 0.581 & \secondv{0.781} & \bestv{0.663} \\
S4 & \secondv{0.613} & \secondv{0.835} & \bestv{0.676} & 0.892 & \secondv{0.884} & \secondv{0.885} & 0.607 & 0.807 & \bestv{0.656} & 0.573 & \secondv{0.789} & \bestv{0.661} & 0.521 & \secondv{0.467} & \secondv{0.441} & 0.581 & \secondv{0.781} & \bestv{0.663} \\
\rowcolor{black!4}
S5 & 0.612 & \secondv{0.835} & \secondv{0.675} & 0.892 & \secondv{0.884} & \secondv{0.885} & 0.608 & 0.807 & \bestv{0.656} & 0.572 & \secondv{0.789} & \secondv{0.660} & \secondv{0.531} & 0.466 & \bestv{0.445} & 0.581 & \secondv{0.781} & \bestv{0.663} \\
\bottomrule
\end{tabular}}
\end{table}

\begin{table}[!htbp]
\centering
\captionsetup{skip=4pt,justification=raggedright,singlelinecheck=false}
\caption{\textbf{Distance MAE ablation} on OakInk-v2, TACO, and HOI4D at 21 joints, 195 vertices, and 778 vertices. Errors are in mm; MAE(All/Pred/GT) use all valid, predicted-contact, and ground-truth-contact points.}
\label{tab:ablation_distance}
\scriptsize
\setlength{\tabcolsep}{1.0pt}
\renewcommand{\arraystretch}{0.96}
\resizebox{\linewidth}{!}{%
\begin{tabular}{l*{9}{c}}
\toprule
& \multicolumn{3}{c}{21 joints} & \multicolumn{3}{c}{195 vertices} & \multicolumn{3}{c}{778 vertices} \\
\cmidrule(lr){2-4}\cmidrule(lr){5-7}\cmidrule(lr){8-10}
Variant & MAE(All)$\downarrow$ & MAE(Pred)$\downarrow$ & MAE(GT)$\downarrow$ & MAE(All)$\downarrow$ & MAE(Pred)$\downarrow$ & MAE(GT)$\downarrow$ & MAE(All)$\downarrow$ & MAE(Pred)$\downarrow$ & MAE(GT)$\downarrow$ \\
\midrule
\multicolumn{10}{l}{\textit{OakInk-v2}}\\[1.5pt]
0.05B & 45.01 & 47.86 & 55.45 & 44.29 & 46.07 & 55.38 & 43.88 & 25.49 & 54.74 \\
0.1B & 39.91 & 41.60 & 45.88 & 39.61 & 41.08 & 47.73 & 39.35 & 25.20 & 46.76 \\
MANO & 34.73 & 36.28 & 39.00 & 35.32 & 37.76 & 41.37 & 34.66 & 17.35 & 41.13 \\
w/o $\Omega$ & 32.79 & 30.73 & 31.78 & 32.24 & 31.80 & 33.21 & 32.28 & 28.37 & 32.47 \\
1-way & 31.17 & 28.30 & 29.59 & 31.10 & 28.64 & 31.03 & 30.86 & 23.54 & 30.00 \\
S1 & 21.96 & 22.96 & 24.10 & 21.65 & 22.73 & 24.15 & 21.27 & \bestv{12.00} & 23.91 \\
S2 & 21.83 & 22.66 & 24.01 & 21.57 & 22.14 & 24.10 & 21.29 & \secondv{12.86} & 23.93 \\
S3 & 21.31 & \secondv{21.90} & 23.27 & 21.10 & \secondv{21.68} & 23.34 & 20.85 & 13.06 & 23.20 \\
S4 & \bestv{21.20} & \bestv{21.82} & \bestv{23.01} & \bestv{20.99} & \bestv{21.65} & \bestv{23.10} & \bestv{20.74} & 12.95 & \bestv{22.97} \\
\rowcolor{black!4}
S5 & \secondv{21.29} & 22.04 & \secondv{23.21} & \secondv{21.06} & 21.78 & \secondv{23.32} & \secondv{20.79} & 12.94 & \secondv{23.18} \\
\midrule
\multicolumn{10}{l}{\textit{TACO}}\\[1.5pt]
0.05B & 41.47 & 45.30 & 49.54 & 40.81 & 44.41 & 49.73 & 40.87 & 26.78 & 49.45 \\
0.1B & 49.27 & 52.15 & 58.61 & 46.88 & 50.39 & 57.48 & 46.00 & 17.83 & 56.79 \\
MANO & 45.46 & 48.21 & 51.97 & 46.16 & 51.21 & 54.07 & 45.31 & 12.78 & 53.51 \\
w/o $\Omega$ & 36.19 & 40.42 & 42.52 & 36.31 & 41.14 & 43.77 & 35.96 & 19.43 & 43.33 \\
1-way & 32.79 & 35.52 & 37.29 & 32.80 & 36.89 & 38.57 & 32.65 & 18.41 & 38.26 \\
S1 & 27.34 & 28.13 & 30.06 & 27.78 & 29.21 & 31.54 & 27.65 & 18.57 & 30.96 \\
S2 & \secondv{25.13} & \secondv{27.16} & \secondv{29.37} & \secondv{25.01} & \secondv{27.60} & \secondv{30.19} & \secondv{24.63} & 10.99 & \secondv{29.81} \\
S3 & \bestv{25.09} & \bestv{27.07} & \bestv{29.29} & \bestv{24.96} & \bestv{27.53} & \bestv{30.12} & \bestv{24.56} & 11.01 & \bestv{29.72} \\
S4 & 25.52 & 27.48 & 29.73 & 25.40 & 27.94 & 30.60 & 24.94 & \secondv{10.94} & 30.17 \\
\rowcolor{black!4}
S5 & 26.26 & 28.09 & 30.47 & 26.12 & 28.51 & 31.38 & 25.57 & \bestv{10.78} & 30.91 \\
\midrule
\multicolumn{10}{l}{\textit{HOI4D}}\\[1.5pt]
0.05B & 120.58 & 119.53 & 129.15 & 117.58 & 119.51 & 128.37 & 115.66 & \bestv{12.06} & 127.52 \\
0.1B & 104.80 & 111.40 & 120.17 & 103.23 & 115.80 & 119.63 & 101.29 & 19.07 & 118.74 \\
MANO & 52.05 & 52.19 & 57.18 & 52.96 & 52.95 & 57.71 & 52.09 & 16.03 & 57.45 \\
w/o $\Omega$ & 38.01 & 38.09 & 41.15 & 38.30 & 38.24 & 41.19 & 38.24 & 25.03 & 41.14 \\
1-way & 56.20 & 58.68 & 64.31 & 55.95 & 57.95 & 64.38 & 55.01 & 20.74 & 63.90 \\
S1 & 20.84 & 18.50 & 18.71 & 22.08 & 19.22 & 19.43 & 22.22 & 15.93 & 19.39 \\
S2 & 18.59 & 16.82 & 16.37 & 19.60 & 17.15 & 16.88 & 19.76 & 14.62 & 16.91 \\
S3 & 18.53 & 16.69 & 16.14 & 19.54 & 17.02 & 16.58 & 19.71 & 14.87 & 16.61 \\
S4 & \secondv{18.20} & \secondv{16.34} & \bestv{15.84} & \secondv{19.16} & \secondv{16.53} & \bestv{16.23} & \secondv{19.33} & 14.33 & \secondv{16.26} \\
\rowcolor{black!4}
S5 & \bestv{18.07} & \bestv{16.27} & \secondv{15.85} & \bestv{19.04} & \bestv{16.43} & \secondv{16.24} & \bestv{19.17} & \secondv{13.99} & \bestv{16.25} \\
\bottomrule
\end{tabular}}
\end{table}

\clearpage

\section{Scene and Camera Evaluation}
\label{app:scene}

\paragraph{Camera metrics.}
ATE (absolute trajectory error, m) is measured after a window-level Sim(3) alignment, and rotation error (degrees) is measured relative to the first valid frame.

\paragraph{Depth metrics and scale protocol.}
Depth uses AbsRel (absolute relative error), RMSE (root mean square error, m), and $\delta_1$, the fraction of valid pixels satisfying $\max(\hat d/d,d/\hat d)<1.25$. A model that returns depth up to scale receives one per-window ratio $\operatorname{median}(\mathrm{GT}/\mathrm{pred})$ before the metrics are computed, whereas a metric-scale model keeps scale one; the last column of \cref{tab:scene} reports the metric gauge $s^d$, that ratio aggregated in log space, so that $1$ means metric scale.

\paragraph{Comparison.}
\cref{tab:scene} is restricted to egocentric methods that reconstruct hands together with the surrounding scene, namely HaWoR, Dyn-HaMR, ReViV, and \method{}. Dedicated geometry models predict cameras and dense shape without hands or interaction attributes, so they are not part of this comparison. \method{} attains the lowest camera error of the four methods and, among the two that return depth, also the lowest AbsRel and RMSE and the highest $\delta_1$; its depth is evaluated at its own scale, whereas ReViV requires a per-window scale fit. HaWoR and Dyn-HaMR return no depth, so their depth and gauge entries are not applicable.
\begin{table}[!htbp]
\centering
\captionsetup{skip=4pt,justification=raggedright,singlelinecheck=false}
\caption{\textbf{Scene estimation on TACO} for hand-centric world-space methods. Depth follows the scale-aligned protocol defined above; $^{100\mathrm{w}}$ rows are unranked.}
\label{tab:scene}
\small
\renewcommand{\arraystretch}{1.0}
\setlength{\aboverulesep}{1.5pt}
\setlength{\belowrulesep}{1.5pt}
\setlength{\tabcolsep}{3pt}
\begin{tabularx}{\linewidth}{l*{6}{>{\centering\arraybackslash}X}}
\toprule
Method & \multicolumn{1}{c}{ATE (m)$\downarrow$} & \multicolumn{1}{c}{Rot ($^\circ$)$\downarrow$} & \multicolumn{1}{c}{AbsRel$\downarrow$} & \multicolumn{1}{c}{RMSE (m)$\downarrow$} & \multicolumn{1}{c}{$\delta_1\uparrow$} & \multicolumn{1}{c}{$s^d$} \\
\midrule
HaWoR & \secondv{0.0064} & \secondv{1.22} & -- & -- & -- & -- \\
Dyn-HaMR$^{100\mathrm{w}}$ & 0.0064 & 1.26 & -- & -- & -- & -- \\
ReViV & 0.0068 & 1.65 & \secondv{0.2649} & \secondv{0.3577} & \secondv{0.5829} & \secondv{0.6401} \\
\rowcolor{black!4}
\method{} & \bestv{0.0055} & \bestv{0.99} & \bestv{0.1318} & \bestv{0.2015} & \bestv{0.8469} & \bestv{1.1364} \\
\bottomrule
\end{tabularx}
\end{table}

\paragraph{Ablation.}
\cref{tab:ablation_scene} reports the TACO scene ablation with the variants defined in Appendix~\ref{app:ablation}. The w/o $\Omega$ and 1-way rows report the same uncoupled scene export, so their scene and camera columns are identical and are not independent scene predictions; their hand, contact, and distance metrics are evaluated separately. Dense aggregation attains the lowest camera error and the default configuration the second lowest, so the multi-rate design costs little on the scene side.
\begin{table}[!htbp]
\centering
\captionsetup{skip=4pt,justification=raggedright,singlelinecheck=false}
\caption{\textbf{Scene ablation on TACO.}}
\label{tab:ablation_scene}
\small
\renewcommand{\arraystretch}{1.0}
\setlength{\aboverulesep}{1.5pt}
\setlength{\belowrulesep}{1.5pt}
\setlength{\tabcolsep}{3.5pt}
\begin{tabularx}{\linewidth}{l*{5}{>{\centering\arraybackslash}X}}
\toprule
Variant & \multicolumn{1}{c}{ATE (m)$\downarrow$} & \multicolumn{1}{c}{Rot ($^\circ$)$\downarrow$} & \multicolumn{1}{c}{AbsRel$\downarrow$} & \multicolumn{1}{c}{RMSE (m)$\downarrow$} & \multicolumn{1}{c}{$\delta_1\uparrow$} \\
\midrule
0.05B & 0.0086 & 1.86 & \bestv{0.117} & \secondv{0.232} & \bestv{0.857} \\
0.1B & 0.0074 & 1.83 & \secondv{0.122} & 0.256 & 0.825 \\
MANO & 0.0084 & 1.60 & 0.132 & 0.256 & 0.806 \\
w/o $\Omega$ & 0.0069 & 1.15 & 0.178 & 0.242 & 0.728 \\
1-way & 0.0069 & 1.15 & 0.178 & 0.242 & 0.728 \\
S1 & \bestv{0.0043} & \bestv{0.82} & 0.136 & 0.284 & 0.805 \\
S2 & 0.0057 & 1.08 & 0.168 & 0.291 & 0.763 \\
S3 & 0.0057 & 1.07 & 0.169 & 0.292 & 0.759 \\
S4 & 0.0056 & 1.00 & 0.166 & 0.290 & 0.768 \\
\rowcolor{black!4}
S5 & \secondv{0.0055} & \secondv{0.99} & 0.132 & \bestv{0.202} & \secondv{0.847} \\
\bottomrule
\end{tabularx}
\end{table}

\clearpage
\section{Camera-Space Qualitative Results}
\label{app:qualitative}
\begin{figure}[H]
  \centering
  \includegraphics[width=\linewidth]{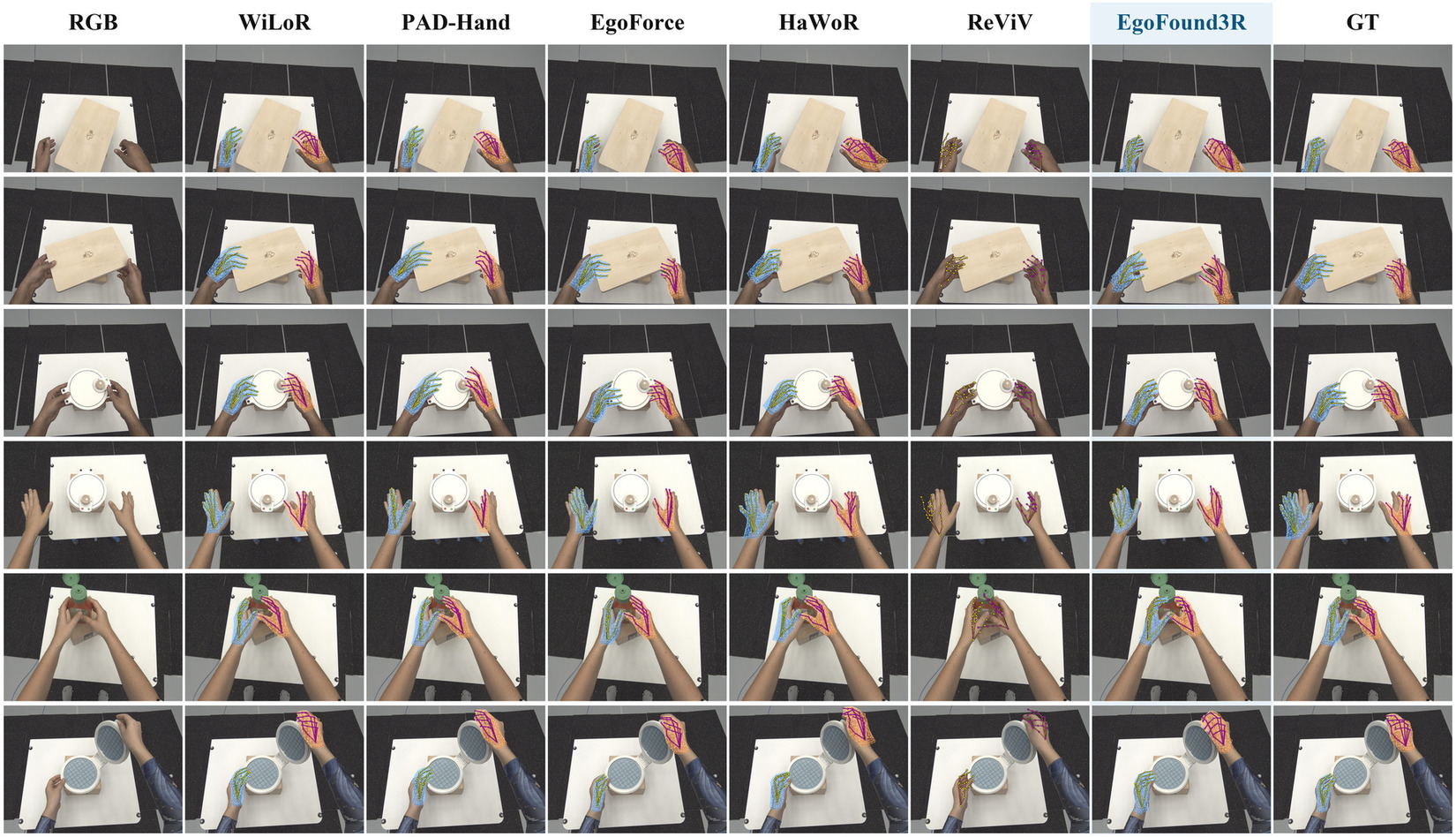}\\[3pt]
  \includegraphics[width=\linewidth]{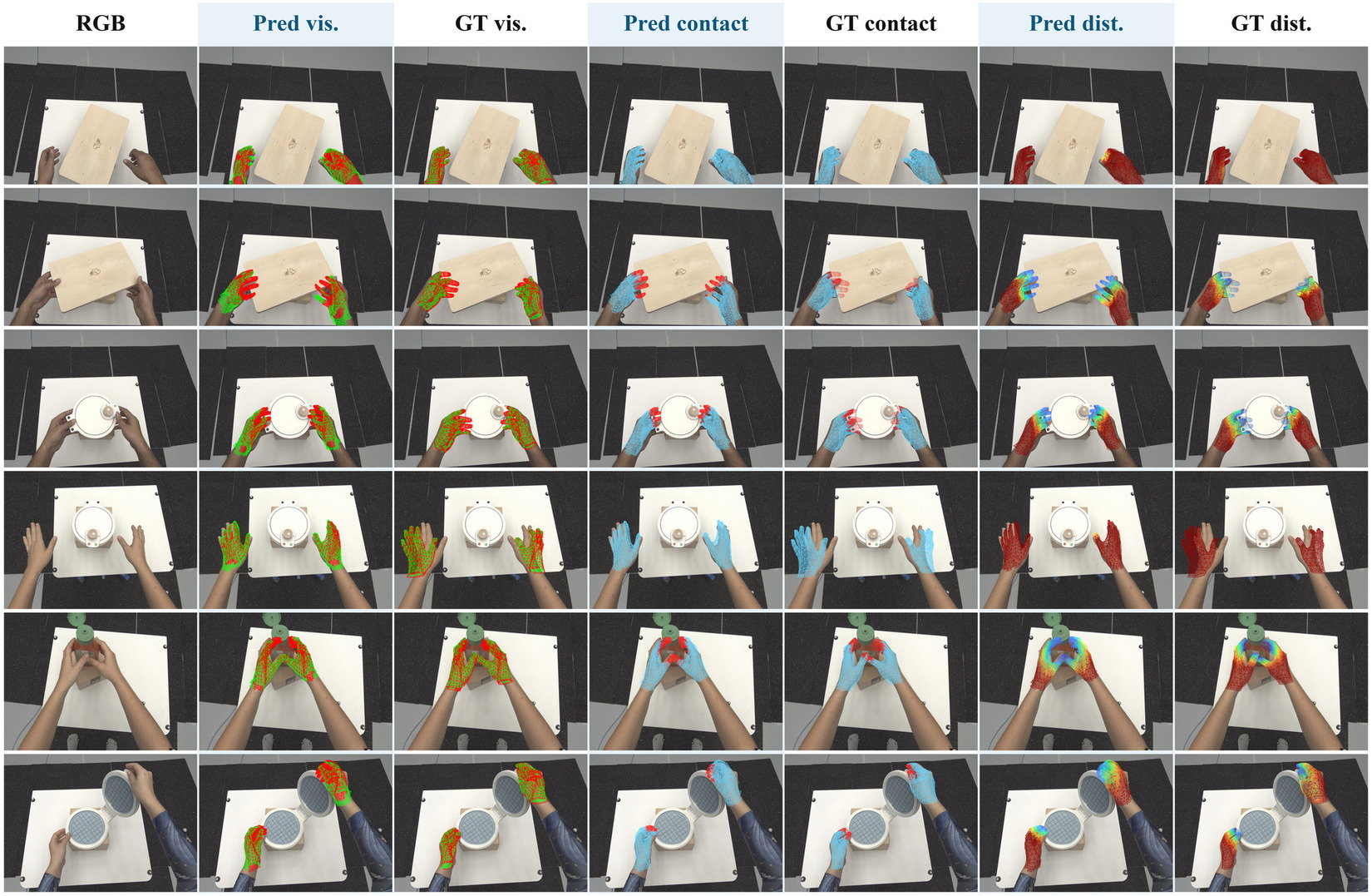}
  \caption{\textbf{Qualitative comparison on ARCTIC.} Top: hand geometry. Bottom: per-point
  visibility, contact, and distance for \method{} (\emph{Pred}) and the ground
  truth (\emph{GT}).}
  \label{fig:qual_arctic}
\end{figure}

\begin{figure}[H]
  \centering
  \includegraphics[width=\linewidth]{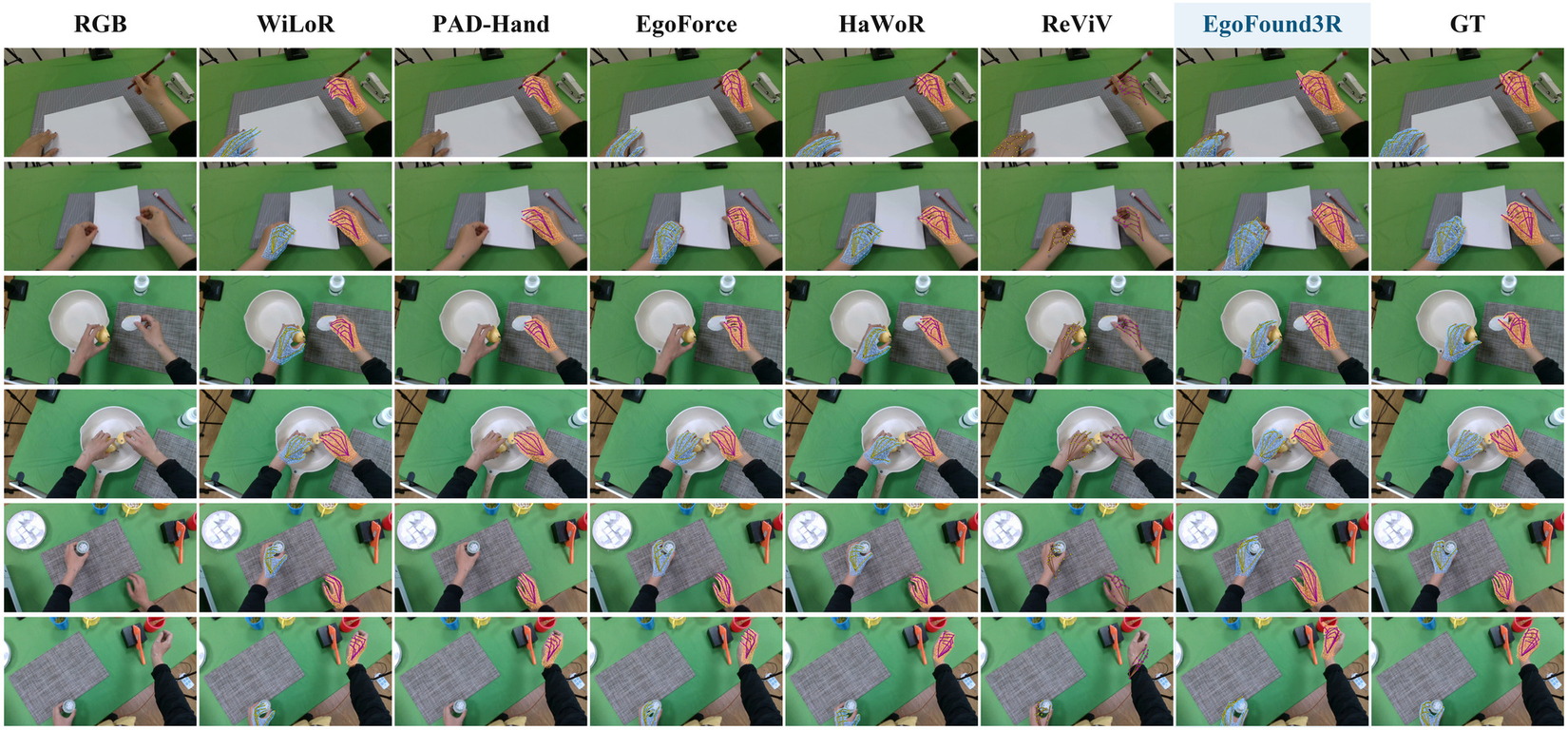}\\[3pt]
  \includegraphics[width=\linewidth]{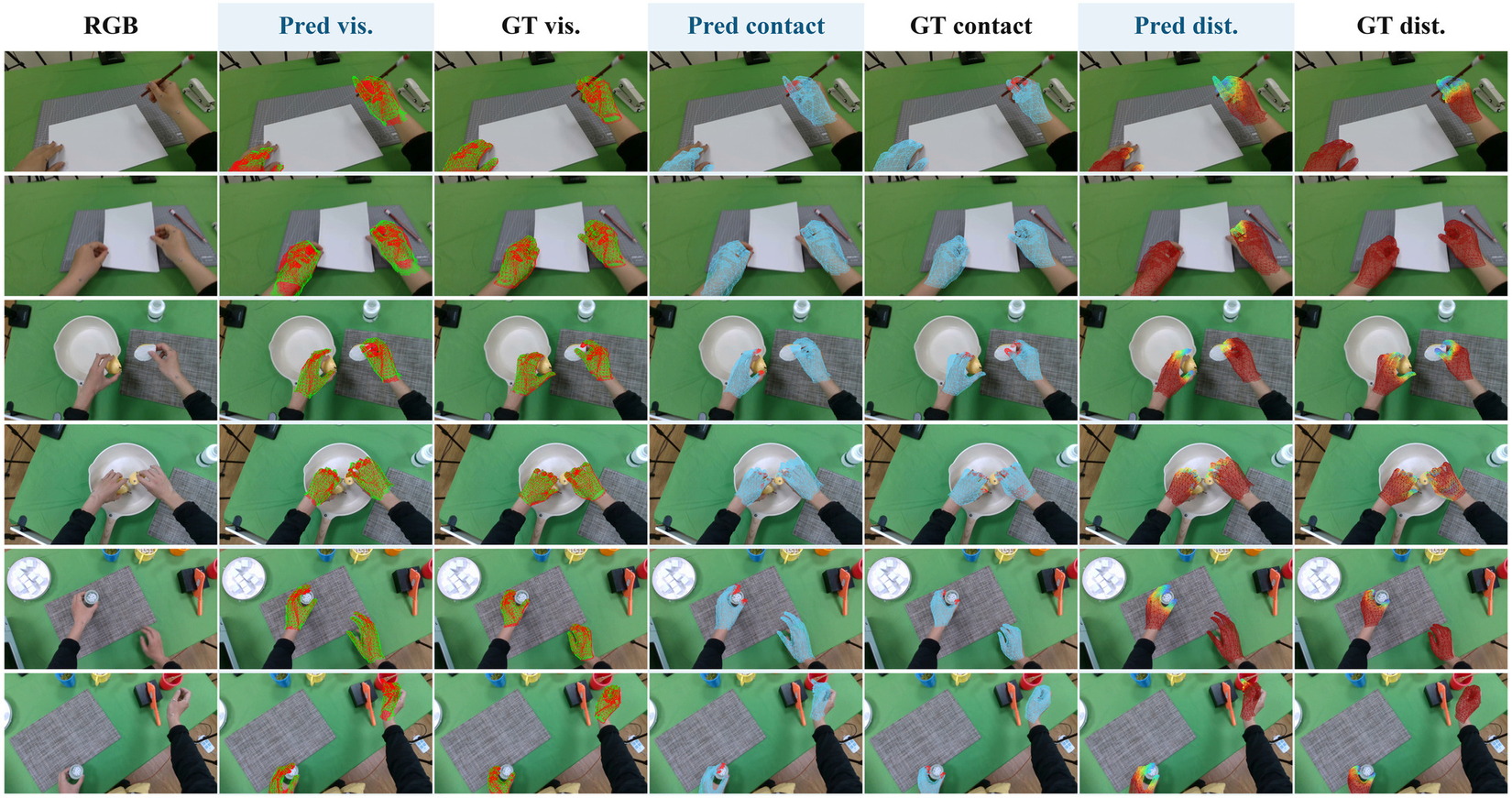}
  \caption{\textbf{Qualitative comparison on OakInk-v2.} Top: hand geometry. Bottom: per-point
  visibility, contact, and distance for \method{} (\emph{Pred}) and the ground
  truth (\emph{GT}).}
  \label{fig:qual_oakink_v2}
\end{figure}

\begin{figure}[H]
  \centering
  \includegraphics[width=0.83\linewidth]{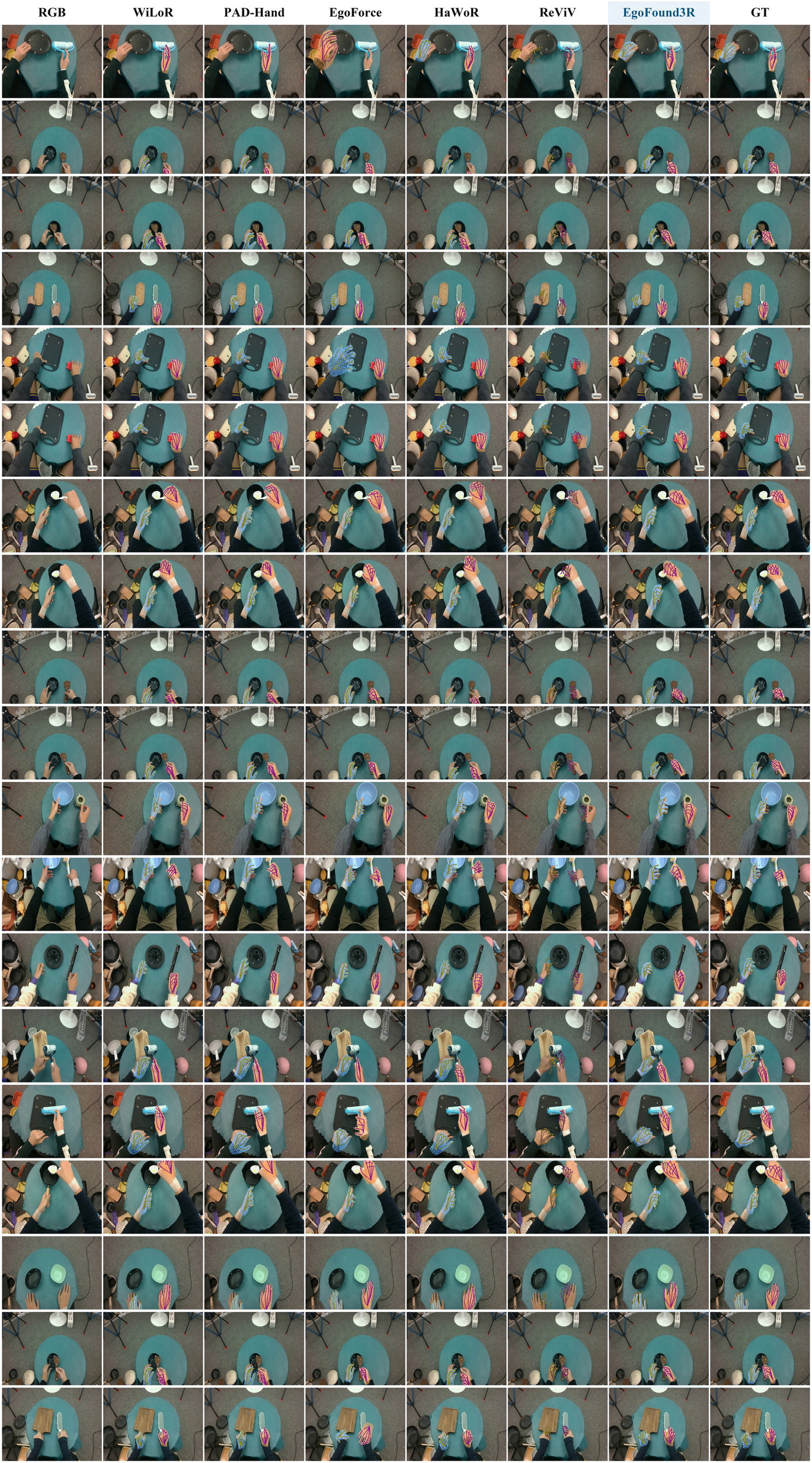}
  \caption{\textbf{Qualitative hand geometry on TACO.} Columns as in \cref{fig:qual_arctic}; the matching attribute panel of the same windows is shown separately below.}
  \label{fig:qual_geom_taco}
\end{figure}

\begin{figure}[H]
  \centering
  \includegraphics[width=0.73\linewidth]{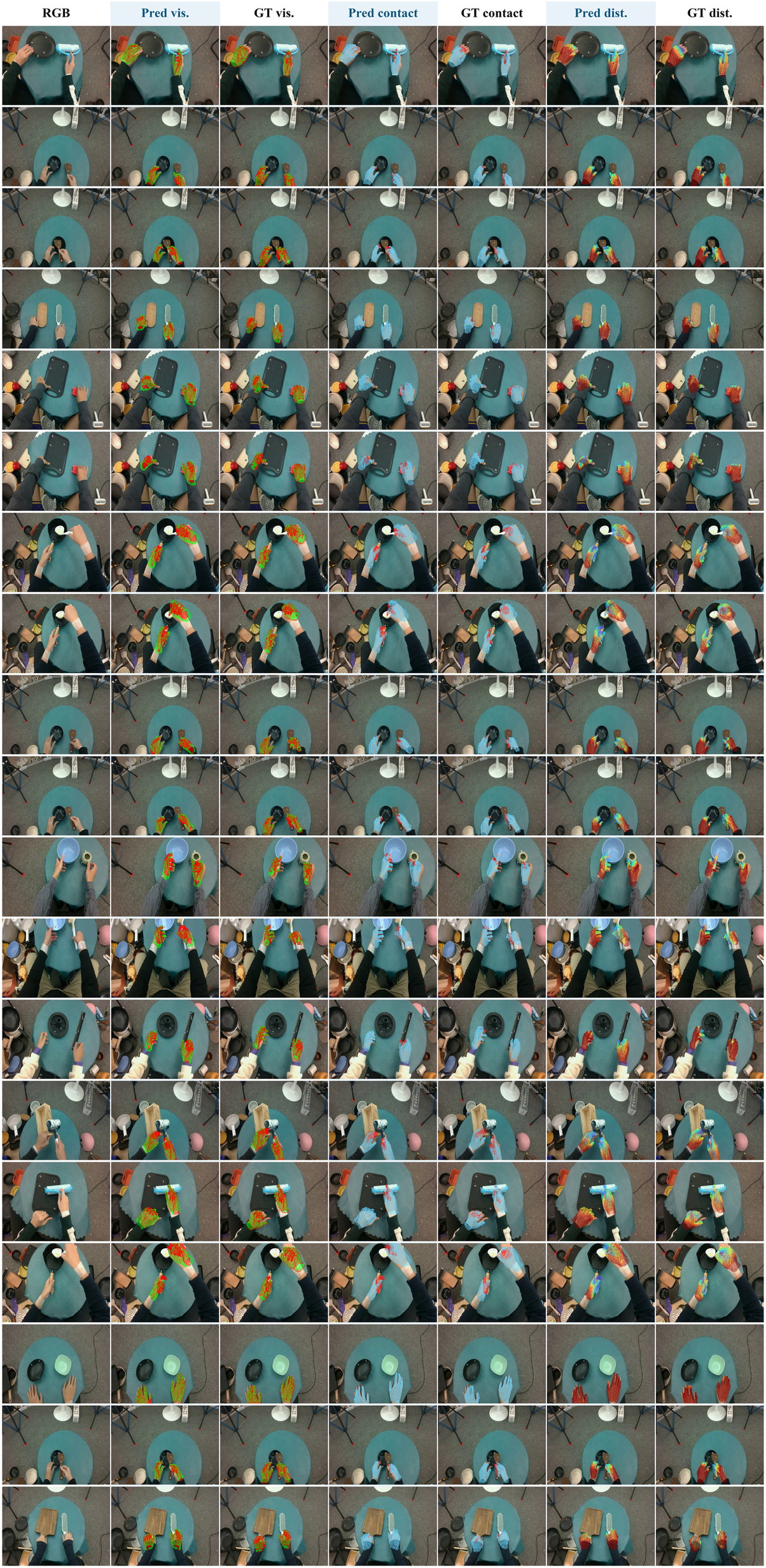}
  \caption{\textbf{Per-point attributes on TACO.} Rows follow the same windows and order as \cref{fig:qual_geom_taco}; columns show the RGB input and the per-point visibility, contact, and distance for \method{} (\emph{Pred}) and the ground truth (\emph{GT}).}
  \label{fig:qual_signal_taco}
\end{figure}

\paragraph{Layout.}
\cref{fig:qual_arctic,fig:qual_oakink_v2} give one dataset per figure, with the hand geometry on top and the attributes of the same windows, in the same row order, below; the two panels of the 19-window TACO block are kept apart in \cref{fig:qual_geom_taco,fig:qual_signal_taco} because the pair does not fit on one page.
Columns follow the tables: the first column is the unmodified RGB input, the next three are the camera-space estimators WiLoR, PAD-Hand, and EgoForce, then the world-space pipelines HaWoR and ReViV, then \method{}, and the last column is the ground-truth annotation.

\paragraph{Reading the panels.}
Each panel overlays the prediction on the pixels of the input frame without additional 2D refinement or fitting, so the comparison does not depend on camera extrinsics.
The attribute panels show visibility, contact, and point-to-surface distance for \method{} (\emph{Pred}) and for the ground truth (\emph{GT}) on the same points as the quantitative tables.

\clearpage
\section{World-Space Qualitative Results}
\label{app:world3d}
\begin{figure}[H]
  \centering
  \includegraphics[width=0.94\linewidth]{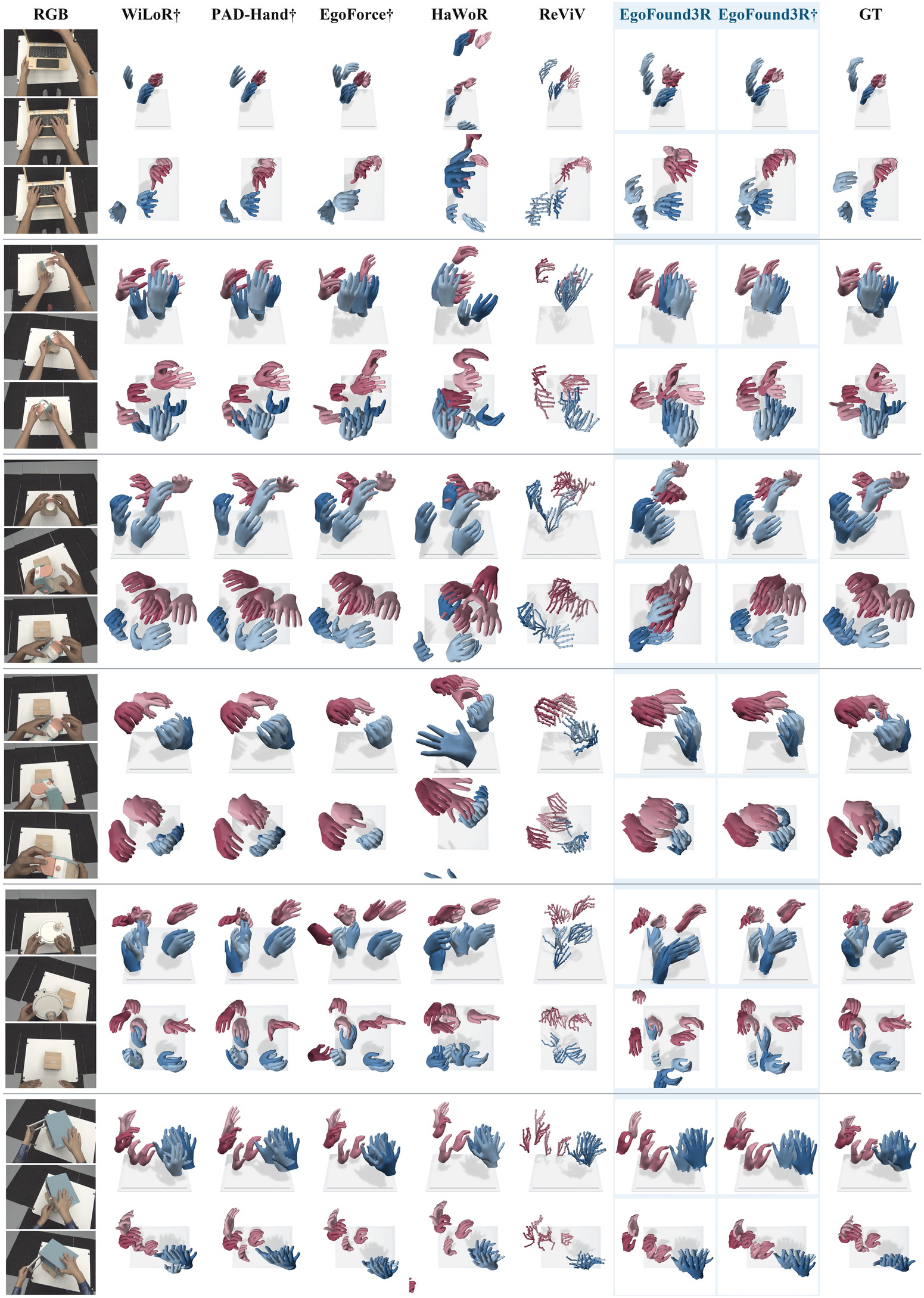}
  \caption{\textbf{World-space qualitative comparison, part 1.}}
  \label{fig:world3d_appendix_01}
\end{figure}

\begin{figure}[H]
  \centering
  \includegraphics[width=0.94\linewidth]{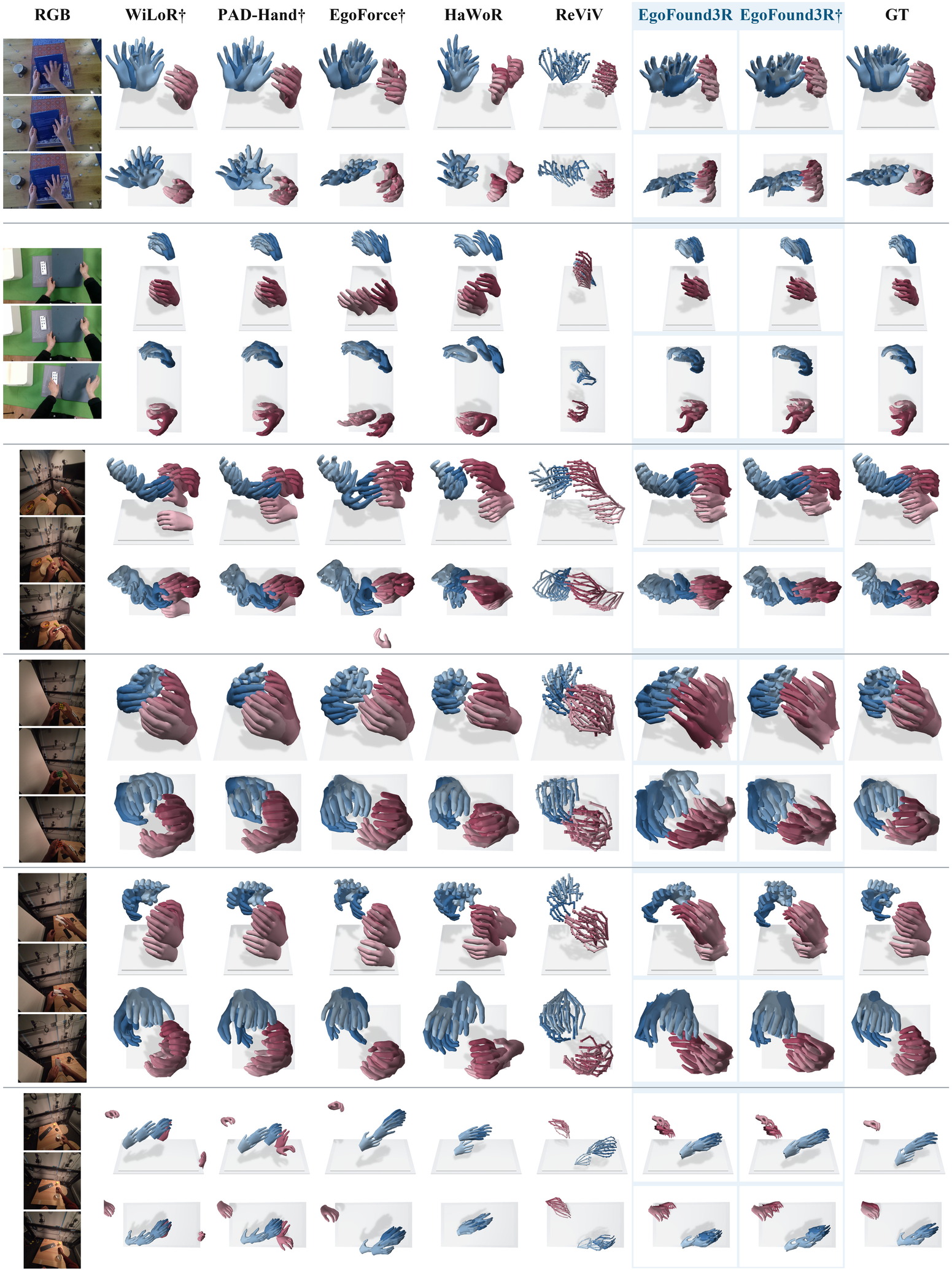}
  \caption{\textbf{World-space qualitative comparison, part 2.}}
  \label{fig:world3d_appendix_02}
\end{figure}

\cref{fig:world3d_appendix_01,fig:world3d_appendix_02} show additional results for the world-space comparison in \cref{fig:world3d_main}.
Both figures keep the layout, columns, hand colors, time shading, and extrinsics convention of \cref{fig:world3d_main}: two fixed viewpoints per clip, one column per method, and the sampled RGB frames on the left.

\end{document}